\documentclass[10pt,twocolumn,letterpaper]{article}

\usepackage[pagenumbers]{wacv} % To force page numbers, e.g. for an arXiv version

\usepackage{graphicx}
\usepackage{amsmath}
\usepackage{amssymb}
\usepackage{booktabs}

\newenvironment{myenum}
{ \begin{enumerate}
    \setlength{\itemsep}{0pt}
    \setlength{\parskip}{0pt}
    \setlength{\parsep}{0pt}     }
{ \end{enumerate}                  }

\usepackage[colorlinks]{hyperref}

\usepackage[capitalize]{cleveref}
\crefname{section}{Sec.}{Secs.}
\Crefname{section}{Section}{Sections}
\Crefname{table}{Table}{Tables}
\crefname{table}{Tab.}{Tabs.}

\def\wacvPaperID{1810} % *** Enter the WACV Paper ID here
\def\confName{WACV}
\def\confYear{2024}

\newcount\Comments  % 1 suppresses notes to selves in text
\usepackage{color}
\definecolor{darkgreen}{rgb}{0,0.5,0}
\newcommand{\kibitz}[2]{\ifnum\Comments=0\textcolor{#1}{#2}\fi}

\begin{document}
\title{So you think you can track?}

\author{Derek Gloudemans$\dagger$ \and Gergely Zachár$\dagger$ \and Yanbing Wang$\dagger$ \and Junyi Ji$\dagger$ \and Matt Nice$\dagger$ \and Matt Bunting$\dagger$ \and William W. Barbour$\dagger$ \and Jonathan Sprinkle$\dagger$ \and Benedetto Piccoli$\ddagger$ \and Maria Laura Delle Monache$\mathsection$ \and Alexandre Bayen$\mathsection$ \and Benjamin Seibold* \and \ \hspace{16em} Daniel B. Work$\dagger$\\ 
Vanderbilt University $\dagger$\\
1625 16th Ave S, Nashville, TN 37212\\
{\tt\small derek.gloudemans@vanderbilt.edu} 
\and \\ 
Rutgers University-Camden $\ddagger$ \\
303 Cooper St, Camden, NJ 08102 
\and  UC Berkeley $\mathsection$\\
University Ave and Oxford St, Berkeley, CA 94720
\and  Temple University * \\
1801 N Broad St, Philadelphia, PA 19122 \\
}
\maketitle

%%%%%%%%% ABSTRACT
\begin{abstract}
  \textit{This work introduces a multi-camera tracking dataset consisting of 234 hours of video data recorded concurrently from 234 overlapping HD cameras covering a 4.2 mile stretch of 8-10 lane interstate highway near Nashville, TN. The video is recorded during a period of high traffic density with 500+ objects typically visible within the scene and typical object longevities of 3-15 minutes. GPS trajectories from 270 vehicle passes through the scene are manually corrected in the video data to provide a set of ground-truth trajectories for recall-oriented tracking metrics, and object detections are provided for each camera in the scene (159 million total before cross-camera fusion). Initial benchmarking of tracking-by-detection algorithms is performed against the GPS trajectories, and a best HOTA of only 9.5\% is obtained (best recall 75.9\% at IOU 0.1, 47.9 average IDs per ground truth object), indicating the benchmarked trackers do not perform sufficiently well at the long temporal and spatial durations required for traffic scene understanding.}
\end{abstract}

%%%%%%%%% BODY TEXT
\section{Introduction}
\label{sec:intro}

\begin{figure*}[htb]
    \centering
    \includegraphics[width = \textwidth]{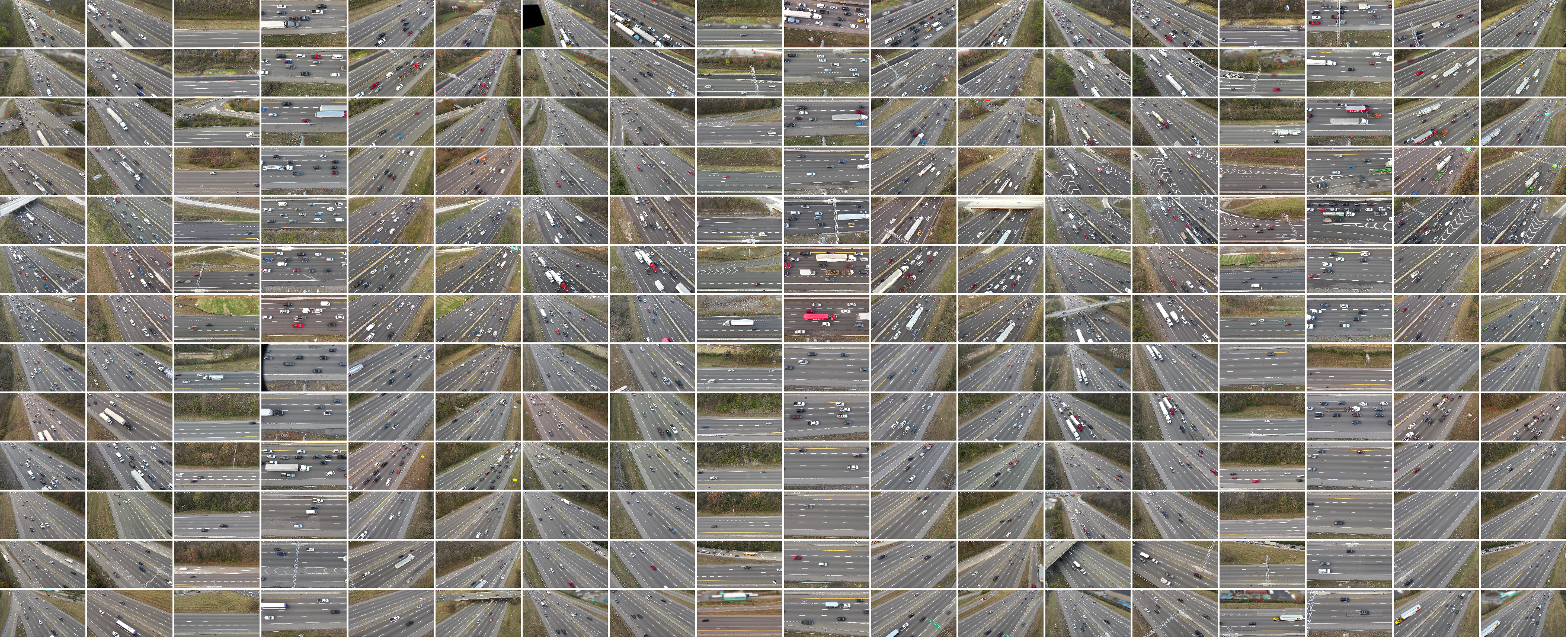}
    \vspace{-0.3in}
    \caption{Example fields of view from each of the 234 cameras included in the I24V dataset. Each camera is recorded in 1920 $\times$ 1080 resolution and at 30 frames per second. Scene information is provided for each roadway direction of travel in each camera.}
    \label{fig:enter-label} 
    \vspace{-0.15in}
\end{figure*}

Much concerted work has been spent on multiple object tracking benchmarks in recent years, primarily from the perspective of pedestrian tracking in crowds \cite{dendorfer2020mot20,milan2016mot16} or vehicle tracking from an AV perspective \cite{geiger2013vision,caesar2020nuscenes}. These datasets generally have high object density, short scenes (1-2 minutes), and short object longevity ($\sim$10 seconds), focusing on high localization accuracy, precision and recall. As a result they do not emphasize challenging aspects of long-term tracking: appearance changes, long-term occlusions, and increasing chance of fragmentation or ID swaps with increasing track length.

Crucially, there is no existing multiple object tracking dataset with a high object density (over 250), long moving object durations (over 5 minutes), and more than 25 overlapping cameras covering a single scene or scenario at the same time. As a result, researchers cannot answer whether existing tracking algorithms are suitable for tracking objects through dense scenes over tens of thousands of frames, because there is no dataset to perform this evaluation on. Such tracking is crucial in the context of traffic science, where origin-destination information for individual vehicles and long-term vehicle behavior are paramount for designing well-fitting models of human driver behavior \cite{li2020trajectory,jones2001keeping}. It is our goal to provide a video dataset of a different spatial and temporal scale than previous works to enable object tracking research in this vein.

To this end, we present the \textit{Interstate 24 Video} (I24V) dataset.  The dataset consists of a single scene, 1 hour in duration, of  4.2 miles of interstate roadway, covered by 234 cameras with overlapping fields of view. Given the scale of this dataset (over 2000 times the video duration of MOTChallenge \cite{dendorfer2020mot20}, 500x the duration of KITTI \cite{geiger2013vision} and 80x the scale of CityFlow \cite{tang2019cityflow}) traditional manual annotation of objects is infeasible. To combat this difficulty, we provide a set of 270 manually-corrected GPS trajectories from over 100 instrumented vehicles on the roadway during the recording duration. Objects persist for an average of 6.6 minutes (11880 frames average at 30 \textit{frames per second} (FPS)) and a high object density ($>$ 500 across the scene) is typically observable. This annotation set is suitable for assessing object tracking algorithms along recall-oriented metrics. Initial experiments show that existing high-performing trackers fall well short of acceptable tracking performance on data of this scale, and further work is needed to develop suitable algorithms for long-term tracking tasks. We take considerable care to make the data useful for computer vision applications, developing new techniques for keeping camera homographies more accurately aligned than existing stabilization methods allow. Succinctly, the contributions of this work are: 
\vspace{-0.05in}
\begin{myenum}
    \item The largest multi-camera video dataset (234 cameras and 234 hours of video covering a scene with high object density and long object durations).
    \item A sparse set of 270 GPS-produced annotations correponding to 1782 minutes of labeled vehicle trajectory.
    \item Preliminary benchmarking of existing object tracking algorithms on this dataset.
    \item Precise scene information and a unified curvilinear coordinate system for the entire scene, useful for filter-based tracking and downstream traffic science.
    \item Methods for precisely re-aligning camera homographies to account for drift outperforming existing image stabilization techniques in over 99\% of cases.
\end{myenum}

The rest of this paper is organized as follows: Section \ref{sec:lit} situates this work within existing literature. Section \ref{sec:dataset} introduces the dataset, its attributes, and methods used to ensure its fidelity. Section \ref{sec:experiments} describes the numerical experiments and Section \ref{sec:results} the  results for homography re-estimation methods and for object tracking algorithms benchmarked on the dataset.  Much additional explanation and analysis omitted for brevity is available in the supplement.

\section{Related Work}
\label{sec:lit}
This section provides a brief overview of existing multiple object tracking and multiple camera datasets, and briefly explores existing multi-camera tracking approaches. 

\noindent \textbf{Multiple Object Tracking Datasets}: 
The task of \textit{multiple object tracking} (MOT) has been well studied, thanks to a number of MOT datasets in different contexts including traffic monitoring \cite{wen2020ua}, drone footage \cite{zhu2018visdrone}, crowded pedestrian scenes \cite{milan2016mot16,dendorfer2020mot20}, and autonomous vehicle scenarios \cite{geiger2013vision,sun2020scalability,caesar2020nuscenes}. Objects are annotated either as 2D rectangular bounding boxes \cite{zhu2018visdrone,wen2020ua,milan2016mot16,dendorfer2020mot20} or 3D rectangular prisms projected from a from a ground-plane into the image \cite{geiger2013vision,sun2020scalability,caesar2020nuscenes}. 3D annotations are primarily seen in the AV context as the collocation of cameras with rich sensor suites such as LIDAR and depth sensors makes the semi-automated production of annotations possible \cite{caesar2020nuscenes}. As a result of this abundance of datasets, a huge variety of accurate MOT methods have been developed in these contexts \cite{luo2021multiple}. A variety of simple yet successful approaches to this task work in an online \textit{tracking-by-detection} paradigm \cite{bochinski2017high,bewley2016simple,wojke2017simple,kiout}. Many modern approaches to the MOT problem solve tracking and object detection \textit{jointly}, sharing information such as object priors, multiple frames, or scene hidden states to aid in detection \cite{zhou2020tracking,bergmann2019tracking,peng2020chained,meinhardt2022trackformer}.

\noindent \textbf{Multi-camera Multiple Object Tracking Datasets}: Relatively less work has been devoted to the task of \textit{multiple-camera, multiple object/target tracking} (MCMT), perhaps due to the fact that such datasets are harder to produce and until recently relatively few have been available. The PETS dataset  \cite{ferryman2009pets2009}, CamNeT \cite{zhang2015camera} and EFPL \cite{fleuret2007multicamera} datasets each track a few pedestrians across up to 8 cameras for relatively short durations in each of a few scenes \cite{fleuret2007multicamera,chavdarova2018wildtrack}.  The Wildtrack Dataset \cite{chavdarova2018wildtrack} and DukeMCMT dataset \cite{gou2017dukemtmc4reid} track pedestrians across 7 and 8 cameras for much longer scene durations (up to 85 minutes). Modern AV datasets include a variety of cameras onboard so can be utilized as multi-camera datasets \cite{caesar2020nuscenes}, but have short scene durations and object longevity. More recently, the CityFlow dataset \cite{tang2019cityflow} contains 40 total cameras (up to 25 for a single scene, with some non-overlapping) totaling over 4 hours of video data in a traffic monitoring context. The pNEUMA Vision dataset \cite{kim2023visual} provides up to 10 drone-mounted camera views and scenes of up to 13 minutes in duration, though has known annotation shortcomings. Synthehicle \cite{herzog2023synthehicle} contains synthetic 3-minute scenes with up to 7 cameras in a traffic monitoring context, totalling over 17 hours of video footage, and the I24-MC3D dataset has scenes with up to 17 cameras and 1.5 minutes in length \cite{gloudemans2023dataset} Table \ref{tab:data} summarizes existing works. Crucially, there is no multi-camera dataset with a high object density (over 100), long object durations (5+ minutes), and more than 25 overlapping cameras. 

\noindent \textbf{Multi-camera Tracking Approaches} generally handle multiple inputs by one of 3 methods: i.) \textit{Detector input fusion} performs object detection utilizing  frames from all cameras simultaneously \cite{zhang2022mutr3d,wu2019accurate,kaygusuz2021multi,singh2023transformer}. ii.)\textit{ Detection fusion} combines all object detections in a shared space via non-maximal suppression \cite{caesar2020nuscenes}, hierarchical clustering of detections \cite{luna2022online,tang2013development}, Gaussian mixture models  \cite{liem2014joint,strigel2013vehicle}, or other methods \cite{dockstader2001multiple} before performing object tracking via traditional approaches. iii.) \textit{Tracklet fusion} methods combine single-camera MOT tracklets with graph-based formulations \cite{zhang2015camera,wu2019multiview,specker2021occlusion}, trajectory to tracklet matching \cite{he2020multi}, or greedy methods based on trajectory similarity measures \cite{tang2018single,hsu2021multi}, often also incorporating camera-link based models \cite{hsu2021multi,yang2021tracklet,he2020multi} or performing smoothing to output more feasible trajectories \cite{wang2022automatic}.

\begin{table}[]
\begingroup
\setlength{\tabcolsep}{3pt} % Default value: 6pt
\renewcommand{\arraystretch}{0.8} % Default value: 1

\begin{tabular}{@{}lccc@{}}
\toprule
\textbf{Dataset}   & \textbf{Cameras} & \textbf{Video} (hr) & \textbf{Scene} (min)  \\ \midrule
DukeMTMC \cite{gou2017dukemtmc4reid}         & 8    & 11.3 & 85   \\
Wildtrack \cite{chavdarova2018wildtrack}     & 7    & 1.0  & 8.6  \\
CityFlow \cite{tang2019cityflow}             & 25   & 3.3  & 6.5  \\
Synthehicle \cite{herzog2023synthehicle}     & 7    & 17   & 3    \\
EPFL-Terrace \cite{fleuret2007multicamera}   & 4    & 14   & 3.5  \\
PETS \cite{ferryman2009pets2009}             & 8    & 0.2  & 0.3 \\
pNEUMA Vision \cite{kim2023visual}           & 10   & 3.9  & 13   \\ 
I24-3D \cite{gloudemans2023dataset}          & 17   & 1.0  & 1.5  \\ \midrule
I24-Video (proposed)                         & 234  & 234  & 60   \\ \bottomrule
\end{tabular} 
\endgroup
\vspace{-0.1in}
\caption{This table summarizes the most comparable existing multi-camera datasets according to \textit{Cameras}, the total number of camera fields of view covering a single scene, \textit{Video}, the total length of all included video, and typical \textit{Scene} duration as estimated from available information for each work.}
\label{tab:data}
\vspace{-0.2in}
\end{table}

\section{Dataset}
\label{sec:dataset}

This section describes the data released in this work. This dataset includes: i.) 234 hours of video concurrently recorded from 234 cameras. ii.) Scene information for each roadway direction of travel in each camera. iii.) A unified curvilinear coordinate system aligned with the primary roadway direction of travel. iv.) Ground truth GPS trajectories for 270 vehicle runs through the camera fields of view v.) Object detections produced at 30Hz on the video scene. Each is described in more detail in the following sections.

\subsection{Video Data}

\subsubsection{Location}
\label{sec:location}
\begin{figure}[htb]
    \centering
    \includegraphics[width = \columnwidth]{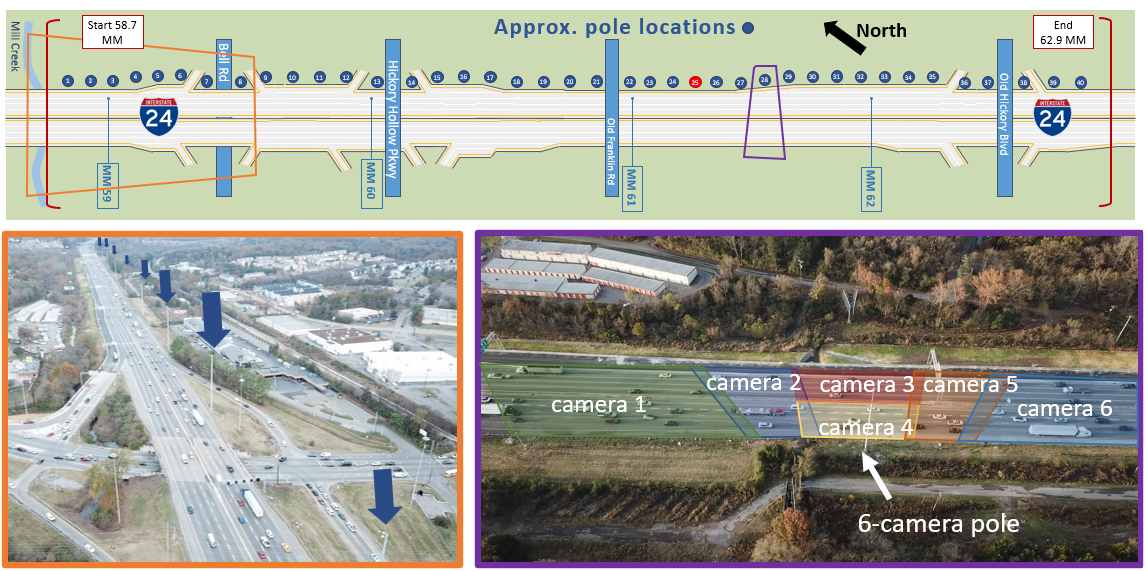}
    \caption{\textbf{(top)} Graphical overview of the I-24 MOTION system. Each blue dot represents a camera pole with 6 cameras. Red dot indicates a camera pole outage (Pole 25). \textbf{(orange)} drone image showing 8 of the 40 system camera poles. \textbf{(purple)} Typical 6 camera per pole coverage layout. Best viewed zoomed-in.} 
    \label{fig:MOTION}
    \vspace{-0.2in}
\end{figure}

Video of a single complex traffic scene was recorded using the I-24 MOTION traffic testbed \cite{gloudemans202324}. Briefly, this system is comprised of 294 IP pan-tilt-zoom cameras densely covering a 4.2 mile stretch of 8-10 lane interstate roadway near Nashville, Tennessee. The main system features 40 $\sim$110-foot tall traffic poles, each with six cameras mounted to provide seamless coverage of roughly 500 feet of the interstate. The primary goal of this camera system is to provide accurate, anonymized vehicle trajectory and dimension information to enable traffic science. See \cite{gloudemans202324} for more details. Figure \ref{fig:MOTION} provides an overview of system features and a typical camera coverage layout for a single camera pole. Due to the layout of the cameras, any object passing through the whole system is visible in a minimum of 185 cameras, and roughly 1-3 cameras at any point in time with a few exceptions for overpasses and camera pole outages.

\begin{figure*}[htb]
    \centering
    \begin{subfigure}[b]{0.32\textwidth}
        \centering
        \includegraphics[width=\textwidth]{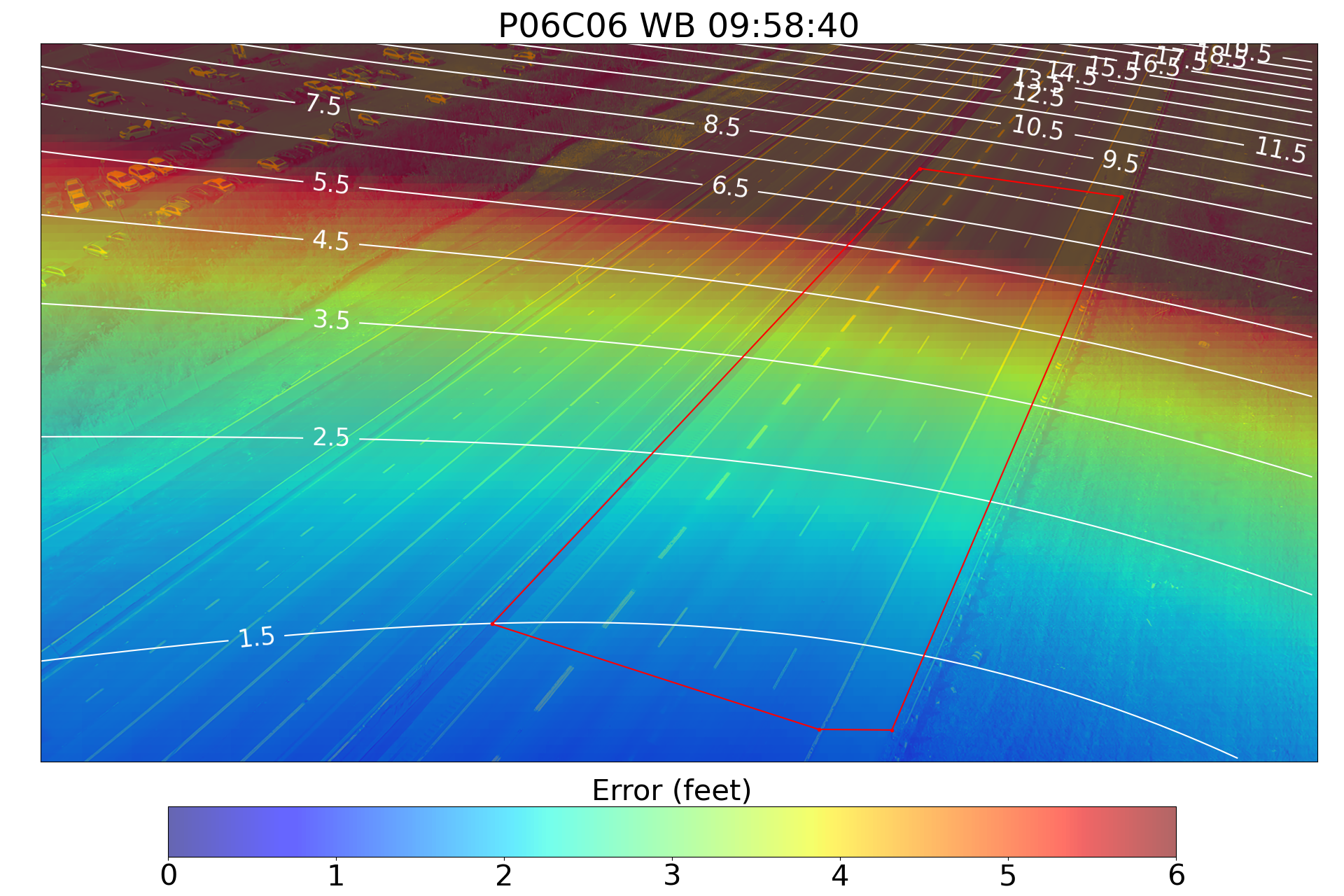}
        \caption{}
        \label{fig:homo_error_cam1}
    \end{subfigure}
    \hfill    
    \begin{subfigure}[b]{0.32\textwidth}
        \centering
        \includegraphics[width=\textwidth]{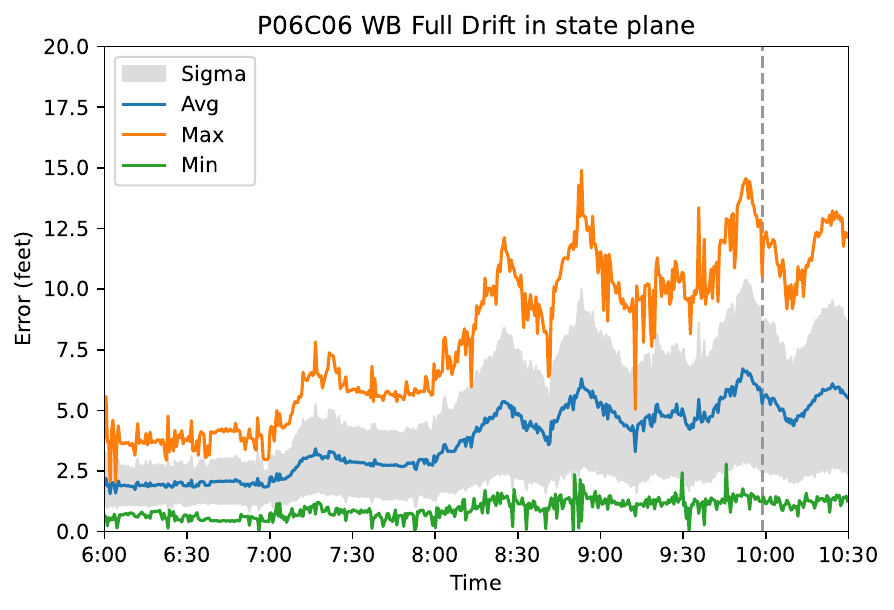}
        \caption{}
        \label{fig:homo_error_cam2}
    \end{subfigure}
    \hfill
    \begin{subfigure}[b]{0.32\textwidth}
        \centering
        \includegraphics[width=\textwidth]{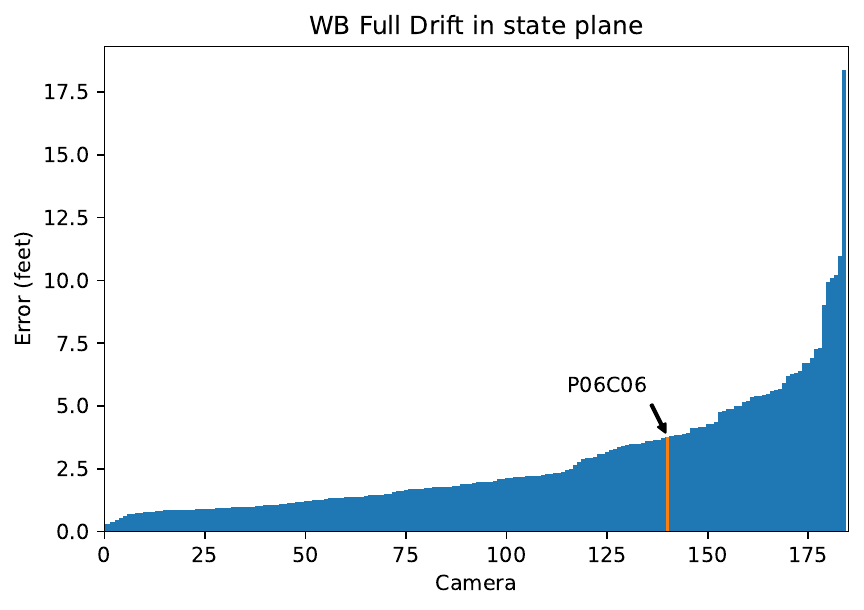}
        \caption{}
        \label{fig:homo_error_all}
    \end{subfigure}
    \vspace{-0.1in}
    \caption{Typical homography error dynamic and the representation of the Sunflower Effect: \textbf{(a.)} uncorrected displacement of image points  showing the magnitude of error (in feet) that using the original homography without accounting for drift would cause. The red polygon area represents the camera FOV, \textbf{(b.)} The displacement error of a typical camera over the day. Gray vertical line indicates the time instant shown in (a).  \textbf{(c.)} Mean average displacement of all the cameras for the westbound roadway side, sorted by magnitude of error.}
    \label{fig:homo_error}
    \vspace{-0.2in}
\end{figure*}

\subsubsection{Recording Details}
On a morning in November 2022, video data was recorded from 234 of the 296 cameras simultaneously from 6:00AM to 10:30AM, roughly covering the morning rush hours. The 7:00-8:00AM hour is published here. HD video (1920 $\times$ 1080 pixels)  was recorded at 30 frames per second from each of the cameras and stored in H.264 compressed format, totaling $\sim$1 TB. Example videos can be found in the supplement. As in \cite{tang2019cityflow}, any visible license plates are redacted using \cite{silva2021flexible}. Each video is then manually inspected, to remove any pedestrians, private property, or other personally identifiable information (see supplement). The one-hour scene has notable features, including i.) several anomalous events, including at least 10 stopped vehicles, ii.) high object density ($>$500 objects present at most times during the recording), and iii.) significant occlusion of vehicles by taller vehicles with moderate frequency. 

\subsection{Scene Homography}
Particular care with scene information is taken in this work as an accurate transformation from image pixel space to a unified coordinate system is a vital pre-requisite for precise multi-camera tracking. The standard approach \cite{hartley2003multiple} utilizes a \textit{homography}, which relates two planar surfaces via a linear transformation, in this case the road surface visible within camera frames and a suitable world coordinate system (We use Tennessee State Plane coordinates (EPSG:2274), which are preferred to other systems such as WSG84 (standard GPS convention) in that they utilize a globally orthonormal basis.) The road surface is treated as a planar surface (for each direction of travel) within a limited \textit{field of view} (FOV) for each camera. Intrinsic-extrinsic camera calibration as used in the AV context \cite{geiger2013vision,caesar2020nuscenes,sun2020scalability} is infeasible here as cameras were not accessible prior to installation, can be replaced or moved, the focus is not fixed, and in-situ intrinsic camera calibration is not possible.

To compute each homography, lane marking corners are utilized as well-defined, semantically meaningful features. World coordinate system points are obtained by manually labeling aerial survey footage ($\sim$1 inch/pixel), while the corresponding image coordinates are produced with semi-automatic labeling on the recorded images. Manual aid was required because the lane markings are identical and repetitive, so additional visual clues were required to uniquely label each lane marking. The homography matrix is fit to these correspondence points via a least-squares formulation as implemented in OpenCV \cite{bradski2000opencv}. See supplement for details.

\subsubsection{Homography Re-estimation} 
\label{sec:homography-re}
Ideally, homographies ensure that multiple views of the same point map to a single unique point on the state plane. In reality, camera fields-of-view are constantly changing due to inaccuracies in the pan-tilt mechanism during homing, settlement of the foundation, and most significantly the \textit{sunflower effect} (the tilting of metal infrastructure poles away from the sun due to differential heating of the sun and shade-facing sides of the pole) \cite{sunflower2020}. Uncorrected, these factors produce significant homography errors sometimes greater than 10 feet. Figure \ref{fig:homo_error_cam1} shows the magnitude of these shifts at one time for a typical camera homography. Figure \ref{fig:homo_error_cam2} illustrates how the average shift for a camera changes over time, due to both the initial error (due to long term phenomena since the initial camera calibration) and the fluctuations in error over a single morning due to the rising morning temperature and changing cloud cover (peaks and valleys).

Repeated manual correction is not feasible, and proper correction of the camera movement is challenging because traditional video stabilization methods (utilizing feature-matching techniques, based on e.g. SIFT \cite{lowe2004distinctive} or SURF \cite{bay2008speeded}) are ill suited for our scenes; a.) a large portion of the image corresponds to ``noise'' (e.g. trees, grass), producing hard-to-match feature points, b.) feature points are usually not semantically meaningful and potentially do not lie on the plane of the road surface, thus are unsuitable for homography estimation, c.) the relevant features on the ground plane in the region of interest are frequently occluded by vehicles, d.) a large number of co-moving vehicles can skew the calculation of optical flow along the direction of vehicle travel. To circumvent these issues, we propose the following homography re-estimation procedure:

\begin{myenum}
    \item Average frames for a suitable time (\(\sim\) 1 min) to remove vehicles from the scene.
    \item Find an initial, rough alignment based on a SIFT and a FLANN-based matcher \cite{muja2009fast} (as in OpenCV \cite{bradski2000opencv}).
    \item Shift original correspondence points using rough alignment. Use to seed re-detection of lane markers.
    \item Filter and refine the detected lane marker corner points.
    \item Calculate the homography matrix using successfully re-identified corner points.
\end{myenum}

For a specific time instance this automatic re-detection and homography re-estimation often fails, either due to i.) lane marking occlusion in heavy traffic or ii.) failure of FLANN matcher. To provide a robust homography in spite of these failures, two methods are proposed and implemented: i.) calculation of a single, \textit{static homography} for an extended period (e.g. all-day) by filtering and averaging homographies over the period, and ii.)  a \textit{dynamic, time-varying homography}. The later a time-varying kernel-based filter of the homography parameters, with a variable window size. Each method is computed offline (utilizing all information for the whole day). Additional details are given in the supplement. The effectiveness of each solution is compared to the existing approach (FLANN-based matcher) in Section \ref{exp:homography}.

\subsection{Roadway Coordinate System}
\begin{figure}
    \centering
    \includegraphics[width = \columnwidth]{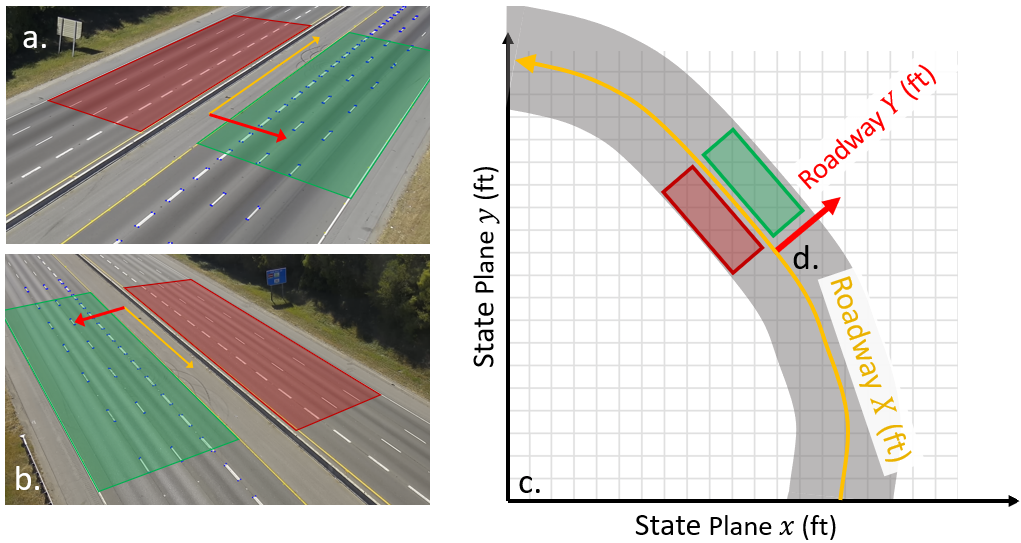}
    \vspace{-0.27in}
    \caption{Camera fields of view (a and b) are related to (c) state plane coordinates, a rectilinear coordinate system, via perspective transforms. State plane coordinates are related to curvilinear roadway coordinates (d) via straightforward mathematical equations.} 
    \label{fig:coordinate-systems}
    \vspace{-0.15in}
\end{figure}

We define an additional roadway coordinate system with the primary (X) axis aligned with the roadway direction of travel, and the secondary(Y) axis always perpendicular to the roadway direction of travel. Since the roadway is not perfectly straight, a \textit{curvilinear coordinate system} is required to achieve the desired attributes, resulting in a locally orthonormal coordinate basis (see Figure \ref{fig:coordinate-systems} for a comparison). Such a coordinate system enables strongly domain-informed filter-based trackers \cite{bewley2016simple} to be implemented trivially (e.g assume that the primary direction of motion for objects is along the primary axis and enforce reasonable vehicle physics). This coordinate formulation is also  preferred for traffic science because quantities such as lane position and inter-vehicle spacing within a lane can be easily computed.  A full description is given in the supplement.

\begin{figure*}[htb]
    \centering
    \includegraphics[width = \textwidth]{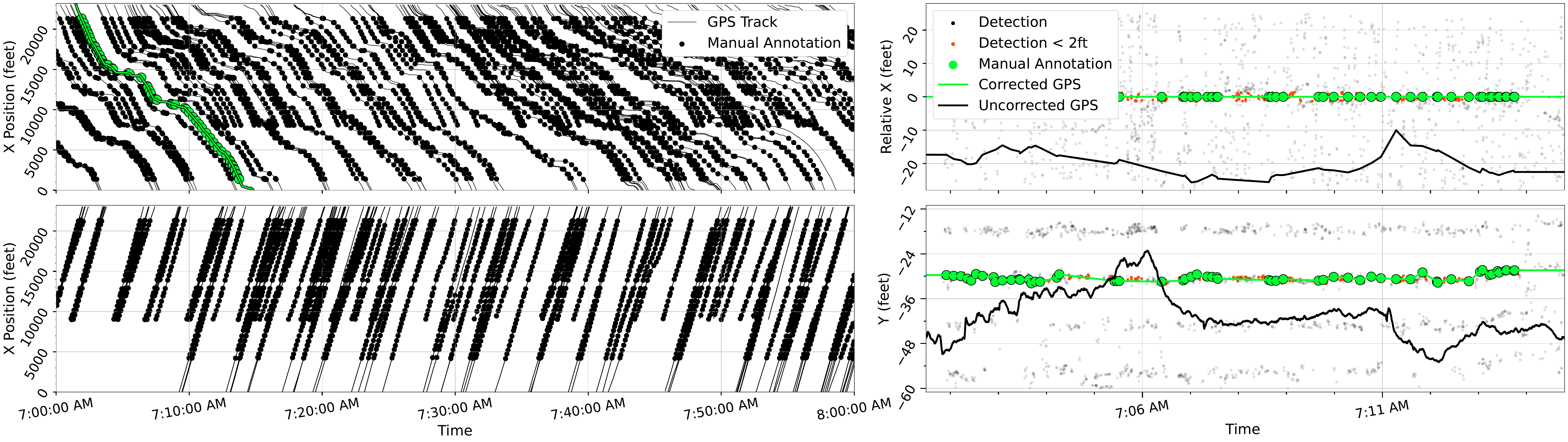}
    \vspace{-0.25in}
    \caption{\textbf{(left)} GPS tracks (lines) and corresponding manual annotations (circles) for westbound (top) and eastbound (bottom) roadway directions of travel. One GPS trajectory is highlighted in green. \textbf{(right)} Detail for highlighted trajectory, showing relative x-position (top) and y-position (bottom) of nearby object detections (black dots), manual annotations (green circles), and the uncorrected corresponding GPS track. Deviations of over 20ft x / 12ft y position can be seen. Detections closely matching corrected GPS track shown in red. (Detections for every 30th frame are plotted for clarity.)}
    \label{fig:GPS-correction}
    \vspace{-0.15in}
\end{figure*}

\subsection{GPS Tracks and Correction}
Concurrent with video recording, a fleet of 103 GPS instrumented vehicles was driven through the portion of roadway observed by the I-24 MOTION testbed. Details on vehicle instrumentation can be found in \cite{bunting2021libpanda}. On these vehicles, positional data was recorded at 0.1s intervals. A total of 270 vehicle passes through the roadway were made during the recording period, providing the same number of vehicle trajectories for comparison. 

\subsubsection{GPS Track Refinement}
Initial attempts to compare GPS track data against known, ground truth object positions (manually annotated) revealed that GPS data contained positional errors (mainly bias along primary direction of travel, and mainly high variance perpendicular to direction of travel), consistent with the GPS sensor's reported error metric of 2.5m \textit{circular error probable} (CEP) (see Figure \ref{fig:GPS-correction}). Additionally, a small time discrepancy between some GPS track data and the camera network is observable. The following protocol was utilized to make GPS trajectories suitable for direct comparison against object tracking outputs from camera data:

\begin{myenum}
    \item Manually annotate a `perfect' position for each GPS track, once per camera pole (e.g. 37+ annotations for a  GPS track that travels the full length of the camera system). See Figure \ref{fig:GPS-correction}.
    \item Correct GPS bias in the roadway coordinate system primary (longitudinal) axis direction by finding the mean offset between GPS positions and manually annotated object positions.
    \item Determine the time offset in the range [-2s,2s] that minimizes the variance in GPS positional offsets relative to manually annotated object positions.
    \item Correct residual error in the longitudinal direction by linearly interpolating the required offset between consecutively labeled offsets between GPS and manually annotated object positions.
    \item Linearly interpolate lateral coordinate between manually annotated object positions for each GPS track.
\end{myenum}

\begin{figure}[b]
    \vspace{-0.2in}
    \centering
    \includegraphics[width = \columnwidth]{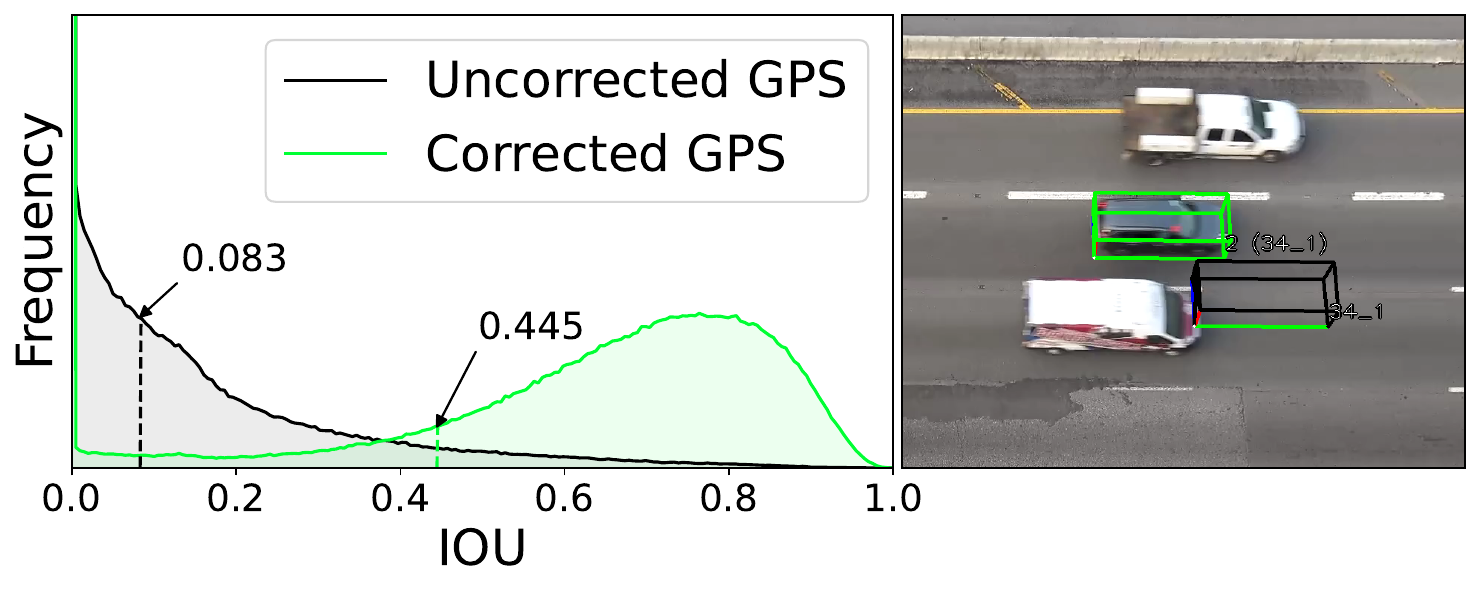}
    \vspace{-0.3in}
    \caption{\textbf{(left)} Intersection-over-union histogram between GPS and closest automatically detected object position, before (black, mean 0.083) and after (green, mean 0.445) manual correction. \textbf{(right)} Examples of corrected (green) and uncorrected (black) GPS positions in a camera field of view.} 
    \label{fig:histogram}
\end{figure}

Figure \ref{fig:GPS-correction} shows the alignment between manually annotated object positions (circles) and GPS positions (lines) for a single typical GPS track. In total 7885 manual annotations are made. Figure \ref{fig:histogram} shows a histogram of GPS intersection-over-union alignment with object detections (see Section \ref{sec:detection}) before and after correction. Corrected GPS tracks align more closely with CNN-produced object detections than original GPS tracks (IOU of 45\% vs 8\%). After correction, 270 vehicle trajectories were produced with an average length of 6.6 minutes and 17560 feet. Each object is virtually always visible in at least one camera, corresponding to a minimum of 3207600 roughly annotated bounding boxes. 

\subsection{Detections}
\label{sec:detection}
To allow preliminary analysis of existing object tracking methods, a baseline set of object detections was produced. Because the cameras in this dataset have widely varying fields of view, a viewpoint agnostic monocular object detector was utilized (i.e. an object detector that does not explicitly or implicitly code scene information into its structure or parameter weights). This allows a single set of network parameters to be utilized for all camera fields of view (rather than training a separate model for each camera field of view, which was infeasible based on storage, implementation, and training time constraints). This work utilized a Retinanet ResNet50-FPN backbone object detector \cite{lin2017focal} to provide detections. The network outputs were parameterized to produce rectangular prism representations for 3D bounding boxes in addition to 2D bounding box outputs for predicted objects (see supplement). Detections are nominally produced at 30 Hz with some frames skipped to provide $\pm$ 1/60s synchronization across all cameras. The resulting dataset contains 158,976,915 detections, each including a 3D bounding box defined in the roadway coordinate system, a 2D bounding box in image coordinates, vehicle class (\textit{sedan}, \textit{midsize}, \textit{van}, \textit{pickup}, \textit{semi truck} or \textit{other truck}), timestamp, camera, and detection confidence.

% \begin{figure}
%     \centering
%     \begin{subfigure}[b]{0.7\columnwidth}
%         \centering
%         \includegraphics[width=\textwidth]{im/histograms.pdf}
%         \caption{}
%         \label{fig:histogram}
%     \end{subfigure}
%     \hfill    
%     \begin{subfigure}[b]{0.29\columnwidth}
%         \centering
%         \includegraphics[width=\textwidth]{im/gps_box.png}
%         \caption{}
%         \label{fig:box}
%     \end{subfigure}
    
%     \caption{(a.) Typical homography goodness-of-fit (mean L2-norm) for a single camera, (b.) error dynamics for a single camera over time with each homography re-estimation methods, (c.) Remaining error for each camera after (black) SIFT-FLANN feature-matching, (orange) one-day best fit homography re-estimation, and (red) dynamic homography re-estimation methods relative to orignal reference homography baseline (blue). Cameras are grouped by position on pole (see Figure \ref{fig:MOTION}) and by side of roadway (westbound homographies on top, eastbound on bottom).)}
%     \label{fig:homo_compare}
% \end{figure}

\section{Experiments}
\label{sec:experiments}
This section first describes experiments used to assess the accuracy of the homography re-estimation method proposed in this work, then describes initial MOT algorithm benchmarking performed using baseline object detections. 

\subsection{Homography Re-estimation}
\label{exp:homography}

\begin{figure*}[ht]
    \centering
    \begin{subfigure}[b]{0.32\textwidth}
        \centering
        \includegraphics[width=\textwidth]{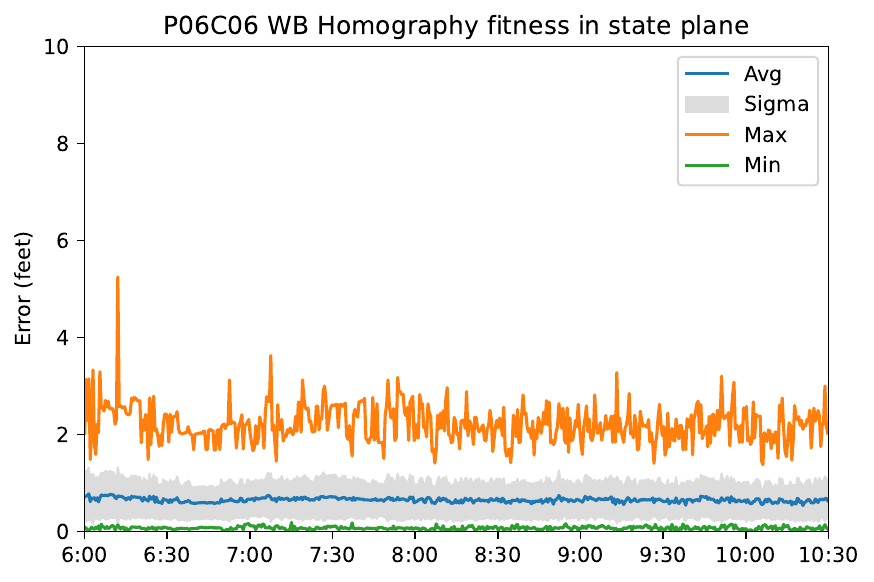}
        \caption{}
        \label{fig:homo_compare_fittness}
    \end{subfigure}
    \hfill    
    \begin{subfigure}[b]{0.32\textwidth}
        \centering
        \includegraphics[width=\textwidth]{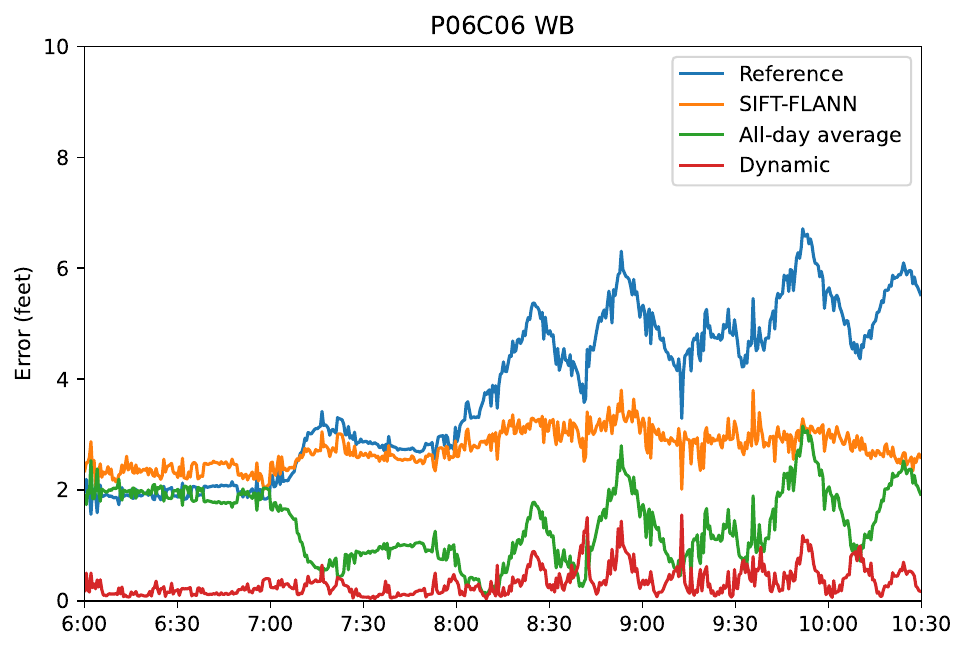}
        \caption{}
        \label{fig:homo_compare_cam}
    \end{subfigure}
    \hfill
    \begin{subfigure}[b]{0.32\textwidth}
        \centering
        \includegraphics[width=\textwidth]{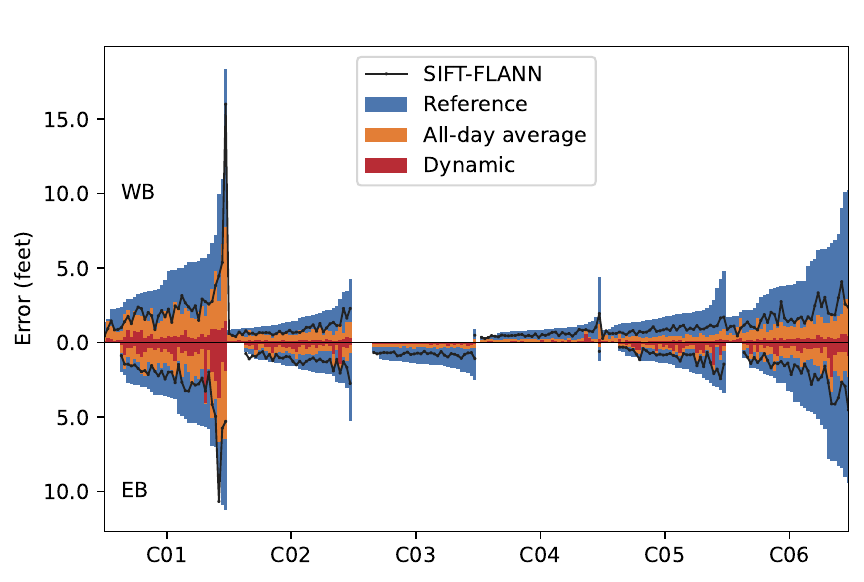}
        \caption{}
        \label{fig:homo_compare_all}
    \end{subfigure}
    \vspace{-0.1in}
    \caption{\textbf{(a.)} Typical homography \textit{fitness} for a single camera, \textbf{(b.)} error dynamics for a single camera over time with each homography re-estimation methods, \textbf{(c.)} Remaining error for each camera after (black) SIFT-FLANN feature-matching, (orange) one-day best fit homography re-estimation, and (red) dynamic homography re-estimation methods relative to orignal reference homography baseline (blue). Cameras are grouped by position on pole (see Figure \ref{fig:MOTION}) and by side of roadway (westbound homographies on top, eastbound on bottom).}
    \label{fig:homo_compare}
    \vspace{-0.2in}
\end{figure*}

To assess the effectiveness of homography re-estimation methods proposed in Section \ref{sec:homography-re}, we utilize the homography goodness-of-fit (equation \ref{eq:fitness}) which indicates how well the homography maps between the image plane and state plane, and the error metric defined in equation \ref{eq:error} which indicates average the positional error in points translated between the image plane and state plane via the computed homography. For each method, homographies are computed at 1 minute intervals overlapping by 50\%. The computed homography's fitness is assessed according to:

\begin{equation}
    fitness(t) =  || \mathcal{A}_t , \mathcal{I}_t' \xrightarrow{H_t} ||_2
    \label{eq:fitness}
\end{equation}
where $\mathcal{I}_t'$ is the subset of correspondence points successfully rediscovered in the image at time $t$,  $A_t$ is the corresponding subset of points in state plane coordinates, $\xrightarrow{H_t}$ indicates a linear transform between coordinate spaces using $H_t$, the homography matrix fit directly to the rediscovered points at time $t$. Error is computed as:

\begin{equation}
    error(t) =  ||\mathcal{I} \xrightarrow{H_t} , \mathcal{I} \xrightarrow{H_t^*}||_2
    \label{eq:error}
\end{equation}
where $\mathcal{I}$ is the full set of correspondence points labeled in the original reference image, $H_t$ is the homography fit directly to time $t$ between the rediscovered points $I'_t$ and the corresponding state plane points $A_t$, and $H_t^*$ is the homography for time $t$ produced by the selected method. Because $H_t$ is prone to error, any reported error may come either from the instantaneous homography $H_t$ or the method-fit homography $H^*_t$ (i.e. $H_t$ is a good baseline when sufficiently many correspondence points are rediscovered.) We report other metrics independent from $H_t$ in supplement.

\subsection{MOT Algorithm Benchmarking}
A limited set of detection-fusion tracking algorithms (SORT \cite{bewley2016simple}, IOUT \cite{bochinski2017high}, KIOU \cite{kiout}, and ByteTrack with both Euclidean distance and IOU as similarity metric \cite{zhang2022bytetrack}) is implemented based on the criteria that i.) algorithms must not require retraining on the tracking data as no training data for object detection is made available, ii.) must not require additional inputs (e.g. appearance embeddings), and iii.) must be tracking by detection-based (not joint detection and tracking-based) methods. These criteria are necessary because, on a dataset of this size, generating auxiliary information or conducting one-off algorithm runs on all videos is prohibitively time-intensive.  For comparison, an \textit{oracle} tracker is implemented which selects all detections close to a GPS trajectory and linearly interpolates tracklet positions between these selected positions. The oracle represents performance theoretically obtainable using the existing set of object detections with a perfect motion model.  This evaluation is merely a first step at gauging the difficulty of this dataset; we make annotations and evaluation protocols public so that researchers may evaluate their own algorithms and report state of the art performance. 

Tracking methods are evaluated using recall, assigned IDs per ground truth trajectory, and \textit{Multiple Object Tracking Precision} (MOTP) in terms of both IOU and Euclidean distance from \cite{bernardin2008evaluating}, \textit{Longest Consecutive Subsequence} (LCSS) by distance and time from \cite{toohey2015trajectory}, and DetA, AssA, and HOTA from \cite{luiten2021hota}. Because the dataset does not densely label objects, a false positive count cannot be obtained. Thus, the DetA metric from \cite{luiten2021hota} is modified:
\begin{equation}
    DetA^*_\alpha = \frac{TP}{TP + FN} 
\end{equation}
where $TP$ represents the number of object positions that are matched to a ground truth position with at least $\alpha$ IOU overlap, and $FN$ represents the number of ground truth object positions with no such match. We follow the rest of the protocol from \cite{luiten2021hota} for calculating AssA and HOTA. 

\subsubsection{Evaluation Protocol}
Each object tracker is run using the detection set from Section \ref{sec:detection}. GPS trajectories and detections from each camera are obtained at slightly different times. To account for this, tracking evaluation is performed at fixed 0.1 second intervals, and each GPS trajectory and object tracklet position is linearly interpolated at each evaluation time. Evaluation is performed as in \cite{bernardin2008evaluating}. For all metrics other than HOTA metrics, a lax IOU of 0.1 is required for an object tracklet and GPS trajectory to be matched.

\section{Results}
\label{sec:results}
\subsection{Homography Re-estimation Performance}
\label{results:homography}

\begin{table*}[ht]
\centering
\begingroup
\setlength{\tabcolsep}{2.5pt} % Default value: 6pt
\renewcommand{\arraystretch}{0.8} % Default value: 1

\begin{tabular}{@{}lcccccccccc@{}}
\toprule
\textbf{Tracker} & \textbf{HOTA}  & \textbf{DetA} & \textbf{AssA} & \textbf{Recall}& \textbf{IDs/GT} $\downarrow$ & \textbf{LCSS$_t$} (s) & \textbf{LCSS$_d$ }(ft)  & \textbf{MOTP$_{i}$ }&\textbf{ MOTP$_{e}$} (ft) $\downarrow$ &\textbf{ TD} (s)\\ \midrule
SORT \cite{bewley2016simple}                          & \textbf{9.5} & 51.3 & \textbf{1.8} & 73.6 & 53.1 & \textbf{51.9} & 2609 & 68.0 & \textbf{2.70} & 12.3 \\                                                         
IOU\cite{bochinski2017high}                           & 1.1 & 7.4 & 0.2 & 20.4  & 60.0 & 16.8 & 53.2 & 36.7 & 7.31 & 8.4  \\
KIOU \cite{bochinski2017high,kiout}                   & 8.5 & 51.2 & 1.4 & 73.9 & \textbf{47.9} & 40.6 & 2181 & 66.9 & 2.72 & \textbf{15.1} \\
ByteTrack (L2) \cite{zhang2022bytetrack}              & \textbf{9.5} & 51.5 & \textbf{1.8} & 73.6 & 53.3 & 51.5 & \textbf{2575} & \textbf{70.0} & 2.71 & 12.4 \\ 
ByteTrack (IOU) \cite{zhang2022bytetrack}             & 8.5 & \textbf{53.1} & 1.4 & \textbf{75.9} & 50.3 & 44.1 & 2390 & 67.1 & 2.72 & 14.9 \\ \midrule
\textit{Oracle}                                       & 53.1 & 55.1 & 51.0 & 86.4 & 1.2 & 636 & 14699 & 75.3 & 2.53 & 690 \\ \bottomrule
\end{tabular}
\endgroup
\vspace{-0.15in}
\caption{Tracking results for limited benchmark algorithm set. For each metric, a higher score is better unless indicated with a $\downarrow$. DetA and AssA indicate the detection and association components of HOTA, respectively. LCSS denotes the average longest consecutive subsequence (in seconds or feet) averaged across all trajectories. MOTP indicates the average precision (by IOU of object footprint or Euclidean distance) for all matched object bounding boxes, averaged over all trajectories. TD indicates mean tracklet duration.}
\label{tab:results-tracking}
\vspace{-0.05in}
\end{table*}

\begin{figure*}[htb]
    \centering
    \includegraphics[width = \textwidth]{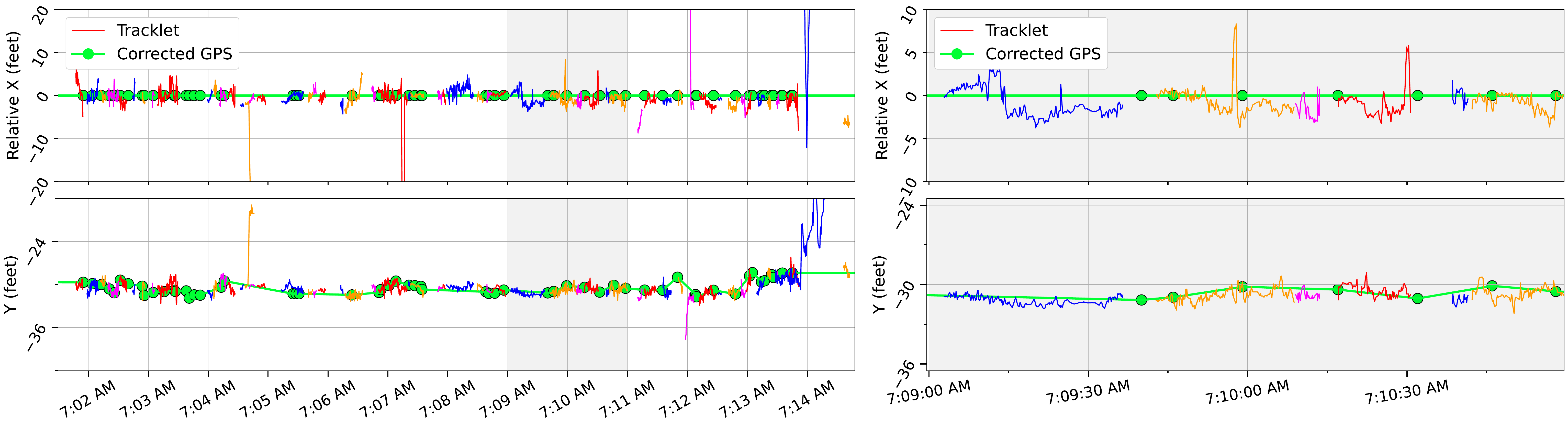}
    \vspace{-0.3in}
    \caption{\textbf{(left)} A single trajectory (green) and all SORT \cite{bewley2016simple} matched to this trajectory at least once (other colors). Manual annotations shown as green circles. \textbf{(right)} A close-up showing the LCSS matched to this trajectory (blue line), lasting $\sim$32 seconds.}
    \label{fig:results-tbd}
    \vspace{-0.15in}
\end{figure*}

Figure \ref{fig:homo_compare_fittness} reports the homography ($H_t$) goodness-of-fit metric (equation \ref{eq:fitness} over time for a typical camera using the homography re-estimation process defined in Section \ref{sec:homography-re}. This $\sim$2ft tightness is guaranteed by outlier removal processes during homography fitting; remaining error is due primarily to camera lens distortions and errors in the flat-plane assumption. The fitness of $H_t$ represents an ``error floor'' for a homography based on the same assumptions.

Figure \ref{fig:homo_compare_cam} show the additional error above the error floor for different homography re-estimation methods utilizing the error metric from equation \ref{eq:error}. The reference (blue) indicates the resulting error without any mitigation, showing both long term (high mean) and short term (high variance) error (3.78 feet whole-day average). The SIFT-FLANN method (the existing optical flow-based "camera stabilization" \cite{bradski2000opencv}), is inferior (2.74 feet whole-day average) in almost all cases to the proposed methods utilizing semantically meaningful lane markers. The static, all-day average homography removes the long term error, although it is mostly unable to remove the error caused by the \textit{sunflower effect} especially in highly fluctuating cases (1.39 feet whole-day average). Lastly, the dynamic homography utilizes nearby (temporally) homography estimations for a given time instance, and can cope with short-term fluctuations caused by camera pole movement, substantially reducing (0.33 feet whole-day average) the residual error caused by the static homography.

Figure \ref{fig:homo_compare_all} compares the whole-day average error, per camera, for each homography re-estimation method. The SIFT-FLANN based method (black line) improves on the reference homography (baseline) for 98.6\% of cameras. The static all-day reestimated homography (green) improves on the baseline for 100\% of cameras and outperforms the SIFT-FLANN method for 88.1\% of cameras. The dynamic homography method (red) improves upon the baseline in 100\% of cases and on the SIFT-FLANN method in 99.7\% of cases. The mean average error over all cameras is 2.78 feet for the reference homography , 1.42ft for the SIFT-FLANN method (49\% reduction), 1.03ft for the all-day average method (63\% reduction), and  0.33 for the dynamic method (88\% reduction). 

\subsection{Multiple Object Tracking Performance}
\label{results:tracking}

Table \ref{tab:results-tracking} shows multiple object tracking performance for the implemented trackers. First, note that HOTA is quite low for all trackers; driven primarily by low AssA scores. This indicates that object tracklets are not strongly persistent (this is also supported by the relatively low LCSS and mean tracklet durations compared to the 6.6 minute mean trajectory length, and high average IDs per ground truth). Such high fragmentation means the tracking outputs are not useful for traffic science applications requiring long and accurate object tracklets. All trackers with a motion model (all but IOU) achieve higher mean recall than raw object detections  of $44.5\%$ (see Figure \ref{fig:histogram}), which indicates that the motion model is crucial for filling in object positional information when detections are missing. Figure \ref{fig:results-tbd} shows an example of all tracklets produced by SORT \cite{bewley2016simple} matched to a single trajectory, and demonstrates the large number of tracklets associated with the ground truth trajectory. 

The purpose of this initial benchmarking is not to claim that no existing tracker can perform well on the I24V dataset, but rather to show that popular off-the-shelf methods are not suitable without substantial enhancement such as more strongly physics and scene-informed models. For instance ByteTrack \cite{zhang2022bytetrack} achieves high performance on datasets such as MOTChallenge, where ID switches and fragmentations play a relatively smaller role in overall scores, but performs poorly on this dataset where object persistence plays a more outsized role in overall tracking performance, especially in the AssA component of HOTA.

\section{Conclusion}
\label{sec:conclusion}

This work introduced the I24-Video Dataset, with concurrent video from 234 cameras recorded for one continuous hour capturing rush-hour traffic along 4.2 miles of interstate roadway, scene information for each camera, and 270 manually corrected GPS trajectories within the video data. These GPS trajectories were used to perform a preliminary benchmarking of object tracking algorithms, indicating that trackers utilizing stronger motion and appearance models are crucial for high performance on this dataset. The work also introduced new methods for keeping traffic camera homographies more precisely synchronized over time than existing methods allow. In the future, we plan to use this dataset to explore and design additional tracking algorithms that prioritize long term (10 minute, 18000 frame) object persistence, necessary for many traffic science applications. Several additional hours of GPS data are also recorded for future public benchmark competitions.

\section*{Acknowledgements}
The authors would like to thank Brad Freeze, Mohamed Osman, Michelle Nickerson the Tennessee Department of Transportation, Matt D'Angelo and Gresham Smith for their efforts on conceptualizing, designing, and implementing the system, Craig Philip and Janos Sztipanovits for their assistance conceptualizing I-24 MOTION. The authors would like to thank Eric Hall for his support on network, hardware, and software integration. This work is supported by the National Science Foundation (NSF) under Grant No. 2135579, the NSF Graduate Research Fellowship Grant No. DGE-1937963 and the USDOT Dwight D. Eisenhower Fellowship program under Grant No. 693JJ32245006 (Gloudemans) and No. 693JJ322NF5201 (Wang). This material is based upon work supported by the U.S. Department of Energy’s Office of Energy Efficiency and Renewable Energy (EERE) award number CID DE-EE0008872. This material is based upon work supported by the CMAQ award number TN20210003. The views expressed herein do not necessarily represent the views of the Tennessee Department of Transportation, U.S. Department of Energy, or the United States Government.

%%%%%%%%% REFERENCES
{\small
\bibliographystyle{ieee_fullname}
\bibliography{sources}

\begin{thebibliography}{10}\itemsep=-1pt

\bibitem{bay2008speeded}
Herbert Bay, Andreas Ess, Tinne Tuytelaars, and Luc Van~Gool.
\newblock Speeded-up robust features (surf).
\newblock {\em Computer vision and image understanding}, 110(3):346--359, 2008.

\bibitem{bergmann2019tracking}
Philipp Bergmann, Tim Meinhardt, and Laura Leal-Taixe.
\newblock Tracking without bells and whistles.
\newblock In {\em Proceedings of the IEEE/CVF International Conference on
  Computer Vision}, pages 941--951, 2019.

\bibitem{bernardin2008evaluating}
Keni Bernardin and Rainer Stiefelhagen.
\newblock Evaluating multiple object tracking performance: the clear mot
  metrics.
\newblock {\em EURASIP Journal on Image and Video Processing}, 2008:1--10,
  2008.

\bibitem{bewley2016simple}
Alex Bewley, Zongyuan Ge, Lionel Ott, Fabio Ramos, and Ben Upcroft.
\newblock Simple online and realtime tracking.
\newblock In {\em 2016 IEEE international conference on image processing
  (ICIP)}, pages 3464--3468. IEEE, 2016.

\bibitem{bochinski2017high}
Erik Bochinski, Volker Eiselein, and Thomas Sikora.
\newblock High-speed tracking-by-detection without using image information.
\newblock In {\em 2017 14th IEEE international conference on advanced video and
  signal based surveillance (AVSS)}, pages 1--6. IEEE, 2017.

\bibitem{bradski2000opencv}
Gary Bradski.
\newblock The opencv library.
\newblock {\em Dr. Dobb's Journal: Software Tools for the Professional
  Programmer}, 25(11):120--123, 2000.

\bibitem{bunting2021libpanda}
Matthew Bunting, Rahul Bhadani, and Jonathan Sprinkle.
\newblock Libpanda: A high performance library for vehicle data collection.
\newblock In {\em Proceedings of the Workshop on Data-Driven and Intelligent
  Cyber-Physical Systems}, DI-CPS'21, page 32–40, New York, NY, USA, 2021.
  Association for Computing Machinery.

\bibitem{caesar2020nuscenes}
Holger Caesar, Varun Bankiti, Alex~H Lang, Sourabh Vora, Venice~Erin Liong,
  Qiang Xu, Anush Krishnan, Yu Pan, Giancarlo Baldan, and Oscar Beijbom.
\newblock nuscenes: A multimodal dataset for autonomous driving.
\newblock In {\em Proceedings of the IEEE/CVF conference on computer vision and
  pattern recognition}, pages 11621--11631, 2020.

\bibitem{chavdarova2018wildtrack}
Tatjana Chavdarova, Pierre Baqu{\'e}, St{\'e}phane Bouquet, Andrii Maksai, Cijo
  Jose, Timur Bagautdinov, Louis Lettry, Pascal Fua, Luc Van~Gool, and
  Fran{\c{c}}ois Fleuret.
\newblock Wildtrack: A multi-camera hd dataset for dense unscripted pedestrian
  detection.
\newblock In {\em Proceedings of the IEEE Conference on Computer Vision and
  Pattern Recognition}, pages 5030--5039, 2018.

\bibitem{kiout}
S. Chen and C. Shao.
\newblock Python implementation of the kalman-iou tracker., 2017.
\newblock \url{https://github.com/siyuanc2/kiout}.

\bibitem{dendorfer2020mot20}
Patrick Dendorfer, Hamid Rezatofighi, Anton Milan, Javen Shi, Daniel Cremers,
  Ian Reid, Stefan Roth, Konrad Schindler, and Laura Leal-Taix{\'e}.
\newblock Mot20: A benchmark for multi object tracking in crowded scenes.
\newblock {\em arXiv preprint arXiv:2003.09003}, 2020.

\bibitem{dockstader2001multiple}
Shiloh~L Dockstader and A~Murat Tekalp.
\newblock Multiple camera tracking of interacting and occluded human motion.
\newblock {\em Proceedings of the IEEE}, 89(10):1441--1455, 2001.

\bibitem{sunflower2020}
Wireless Estimator.
\newblock A safe out of plumb monopole is most likely caused by the thermal
  ‘sunflower effect', 2020.
\newblock
  \url{https://wirelessestimator.com/articles/2020/a-safe-out-of-plumb-monopole-is-most-likely-caused-by-the-thermal-sunflower-effect/}.

\bibitem{ferryman2009pets2009}
James Ferryman and Ali Shahrokni.
\newblock Pets2009: Dataset and challenge.
\newblock In {\em 2009 Twelfth IEEE international workshop on performance
  evaluation of tracking and surveillance}, pages 1--6. IEEE, 2009.

\bibitem{fleuret2007multicamera}
Francois Fleuret, Jerome Berclaz, Richard Lengagne, and Pascal Fua.
\newblock Multicamera people tracking with a probabilistic occupancy map.
\newblock {\em IEEE transactions on pattern analysis and machine intelligence},
  30(2):267--282, 2007.

\bibitem{geiger2013vision}
Andreas Geiger, Philip Lenz, Christoph Stiller, and Raquel Urtasun.
\newblock Vision meets robotics: The kitti dataset.
\newblock {\em The International Journal of Robotics Research},
  32(11):1231--1237, 2013.

\bibitem{gloudemans2023dataset}
Derek Gloudemans, Gracie Gumm, Yanbing Wang, William Barbour, and Daniel~B.
  Work.
\newblock The interstate-24 3d dataset: a new benchmark for 3d multi-camera
  vehicle tracking.
\newblock {\em arXiv preprint arXiv:2308.14833}, 2023.

\bibitem{gloudemans202324}
Derek Gloudemans, Yanbing Wang, Junyi Ji, Gergely Zachar, Will Barbour, and
  Daniel~B Work.
\newblock I-24 motion: An instrument for freeway traffic science.
\newblock {\em arXiv preprint arXiv:2301.11198}, 2023.

\bibitem{gou2017dukemtmc4reid}
Mengran Gou, Srikrishna Karanam, Wenqian Liu, Octavia Camps, and Richard~J
  Radke.
\newblock Dukemtmc4reid: A large-scale multi-camera person re-identification
  dataset.
\newblock In {\em Proceedings of the IEEE Conference on Computer Vision and
  Pattern Recognition Workshops}, pages 10--19, 2017.

\bibitem{hartley2003multiple}
Richard Hartley and Andrew Zisserman.
\newblock {\em Multiple view geometry in computer vision}.
\newblock Cambridge university press, 2003.

\bibitem{he2020multi}
Yuhang He, Xing Wei, Xiaopeng Hong, Weiwei Shi, and Yihong Gong.
\newblock Multi-target multi-camera tracking by tracklet-to-target assignment.
\newblock {\em IEEE Transactions on Image Processing}, 29:5191--5205, 2020.

\bibitem{herzog2023synthehicle}
Fabian Herzog, Junpeng Chen, Torben Teepe, Johannes Gilg, Stefan H{\"o}rmann,
  and Gerhard Rigoll.
\newblock Synthehicle: Multi-vehicle multi-camera tracking in virtual cities.
\newblock In {\em Proceedings of the IEEE/CVF Winter Conference on Applications
  of Computer Vision}, pages 1--11, 2023.

\bibitem{hsu2021multi}
Hung-Min Hsu, Jiarui Cai, Yizhou Wang, Jenq-Neng Hwang, and Kwang-Ju Kim.
\newblock Multi-target multi-camera tracking of vehicles using metadata-aided
  re-id and trajectory-based camera link model.
\newblock {\em IEEE Transactions on Image Processing}, 30:5198--5210, 2021.

\bibitem{jones2001keeping}
Willie~D Jones.
\newblock Keeping cars from crashing.
\newblock {\em IEEE spectrum}, 38(9):40--45, 2001.

\bibitem{kaygusuz2021multi}
Nimet Kaygusuz, Oscar Mendez, and Richard Bowden.
\newblock Multi-camera sensor fusion for visual odometry using deep uncertainty
  estimation.
\newblock In {\em 2021 IEEE International Intelligent Transportation Systems
  Conference (ITSC)}, pages 2944--2949. IEEE, 2021.

\bibitem{kim2023visual}
Sohyeong Kim, Georg Anagnostopoulos, Emmanouil Barmpounakis, and Nikolas
  Geroliminis.
\newblock Visual extensions and anomaly detection in the pneuma experiment with
  a swarm of drones.
\newblock {\em Transportation Research Part C: Emerging Technologies},
  147:103966, 2023.

\bibitem{li2020trajectory}
Li Li, Rui Jiang, Zhengbing He, Xiqun~Michael Chen, and Xuesong Zhou.
\newblock Trajectory data-based traffic flow studies: A revisit.
\newblock {\em Transportation Research Part C: Emerging Technologies},
  114:225--240, 2020.

\bibitem{liem2014joint}
Martijn~C Liem and Dariu~M Gavrila.
\newblock Joint multi-person detection and tracking from overlapping cameras.
\newblock {\em Computer Vision and Image Understanding}, 128:36--50, 2014.

\bibitem{lin2017focal}
Tsung-Yi Lin, Priya Goyal, Ross Girshick, Kaiming He, and Piotr Doll{\'a}r.
\newblock Focal loss for dense object detection.
\newblock In {\em Proceedings of the IEEE international conference on computer
  vision}, pages 2980--2988, 2017.

\bibitem{lowe2004distinctive}
David~G Lowe.
\newblock Distinctive image features from scale-invariant keypoints.
\newblock {\em International journal of computer vision}, 60:91--110, 2004.

\bibitem{luiten2021hota}
Jonathon Luiten, Aljosa Osep, Patrick Dendorfer, Philip Torr, Andreas Geiger,
  Laura Leal-Taix{\'e}, and Bastian Leibe.
\newblock Hota: A higher order metric for evaluating multi-object tracking.
\newblock {\em International journal of computer vision}, 129:548--578, 2021.

\bibitem{luna2022online}
Elena Luna, Juan~C SanMiguel, Jos{\'e}~M Mart{\'\i}nez, and Marcos
  Escudero-Vi{\~n}olo.
\newblock Online clustering-based multi-camera vehicle tracking in scenarios
  with overlapping fovs.
\newblock {\em Multimedia Tools and Applications}, pages 1--21, 2022.

\bibitem{luo2021multiple}
Wenhan Luo, Junliang Xing, Anton Milan, Xiaoqin Zhang, Wei Liu, and Tae-Kyun
  Kim.
\newblock Multiple object tracking: A literature review.
\newblock {\em Artificial intelligence}, 293:103448, 2021.

\bibitem{meinhardt2022trackformer}
Tim Meinhardt, Alexander Kirillov, Laura Leal-Taixe, and Christoph
  Feichtenhofer.
\newblock Trackformer: Multi-object tracking with transformers.
\newblock In {\em Proceedings of the IEEE/CVF conference on computer vision and
  pattern recognition}, pages 8844--8854, 2022.

\bibitem{milan2016mot16}
Anton Milan, Laura Leal-Taix{\'e}, Ian Reid, Stefan Roth, and Konrad Schindler.
\newblock Mot16: A benchmark for multi-object tracking.
\newblock {\em arXiv preprint arXiv:1603.00831}, 2016.

\bibitem{muja2009fast}
Marius Muja and David~G Lowe.
\newblock Fast approximate nearest neighbors with automatic algorithm
  configuration.
\newblock {\em VISAPP (1)}, 2(331-340):2, 2009.

\bibitem{peng2020chained}
Jinlong Peng, Changan Wang, Fangbin Wan, Yang Wu, Yabiao Wang, Ying Tai,
  Chengjie Wang, Jilin Li, Feiyue Huang, and Yanwei Fu.
\newblock Chained-tracker: Chaining paired attentive regression results for
  end-to-end joint multiple-object detection and tracking.
\newblock In {\em Computer Vision--ECCV 2020: 16th European Conference,
  Glasgow, UK, August 23--28, 2020, Proceedings, Part IV 16}, pages 145--161.
  Springer, 2020.

\bibitem{silva2021flexible}
Sergio~M Silva and Cl{\'a}udio~Rosito Jung.
\newblock A flexible approach for automatic license plate recognition in
  unconstrained scenarios.
\newblock {\em IEEE Transactions on Intelligent Transportation Systems},
  23(6):5693--5703, 2021.

\bibitem{singh2023transformer}
Apoorv Singh.
\newblock Transformer-based sensor fusion for autonomous driving: A survey.
\newblock {\em arXiv preprint arXiv:2302.11481}, 2023.

\bibitem{specker2021occlusion}
Andreas Specker, Daniel Stadler, Lucas Florin, and Jurgen Beyerer.
\newblock An occlusion-aware multi-target multi-camera tracking system.
\newblock In {\em Proceedings of the IEEE/CVF Conference on Computer Vision and
  Pattern Recognition}, pages 4173--4182, 2021.

\bibitem{strigel2013vehicle}
Elias Strigel, Daniel Meissner, and Klaus Dietmayer.
\newblock Vehicle detection and tracking at intersections by fusing multiple
  camera views.
\newblock In {\em 2013 IEEE Intelligent Vehicles Symposium (IV)}, pages
  882--887. IEEE, 2013.

\bibitem{sun2020scalability}
Pei Sun, Henrik Kretzschmar, Xerxes Dotiwalla, Aurelien Chouard, Vijaysai
  Patnaik, Paul Tsui, James Guo, Yin Zhou, Yuning Chai, Benjamin Caine, et~al.
\newblock Scalability in perception for autonomous driving: Waymo open dataset.
\newblock In {\em Proceedings of the IEEE/CVF conference on computer vision and
  pattern recognition}, pages 2446--2454, 2020.

\bibitem{tang2013development}
Hua Tang.
\newblock Development of a multiple-camera tracking system for accurate traffic
  performance measurements at intersections, 2013.

\bibitem{tang2019cityflow}
Zheng Tang, Milind Naphade, Ming-Yu Liu, Xiaodong Yang, Stan Birchfield, Shuo
  Wang, Ratnesh Kumar, David Anastasiu, and Jenq-Neng Hwang.
\newblock Cityflow: A city-scale benchmark for multi-target multi-camera
  vehicle tracking and re-identification.
\newblock In {\em Proceedings of the IEEE/CVF Conference on Computer Vision and
  Pattern Recognition}, pages 8797--8806, 2019.

\bibitem{tang2018single}
Zheng Tang, Gaoang Wang, Hao Xiao, Aotian Zheng, and Jenq-Neng Hwang.
\newblock Single-camera and inter-camera vehicle tracking and 3d speed
  estimation based on fusion of visual and semantic features.
\newblock In {\em Proceedings of the IEEE conference on computer vision and
  pattern recognition workshops}, pages 108--115, 2018.

\bibitem{toohey2015trajectory}
Kevin Toohey and Matt Duckham.
\newblock Trajectory similarity measures.
\newblock {\em Sigspatial Special}, 7(1):43--50, 2015.

\bibitem{wang2022automatic}
Yanbing Wang, Derek Gloudemans, Zi~Nean Teoh, Lisa Liu, Gergely Zach{\'a}r,
  William Barbour, and Daniel Work.
\newblock Automatic vehicle trajectory data reconstruction at scale.
\newblock {\em arXiv preprint arXiv:2212.07907}, 2022.

\bibitem{wen2020ua}
Longyin Wen, Dawei Du, Zhaowei Cai, Zhen Lei, Ming-Ching Chang, Honggang Qi,
  Jongwoo Lim, Ming-Hsuan Yang, and Siwei Lyu.
\newblock Ua-detrac: A new benchmark and protocol for multi-object detection
  and tracking.
\newblock {\em Computer Vision and Image Understanding}, 193:102907, 2020.

\bibitem{wojke2017simple}
Nicolai Wojke, Alex Bewley, and Dietrich Paulus.
\newblock Simple online and realtime tracking with a deep association metric.
\newblock In {\em 2017 IEEE international conference on image processing
  (ICIP)}, pages 3645--3649. IEEE, 2017.

\bibitem{wu2019accurate}
Hao Wu, Xinxiang Zhang, Brett Story, and Dinesh Rajan.
\newblock Accurate vehicle detection using multi-camera data fusion and machine
  learning.
\newblock In {\em ICASSP 2019-2019 IEEE International Conference on Acoustics,
  Speech and Signal Processing (ICASSP)}, pages 3767--3771. IEEE, 2019.

\bibitem{wu2019multiview}
Minye Wu, Guli Zhang, Ning Bi, Ling Xie, Yuanquan Hu, and Zhiru Shi.
\newblock Multiview vehicle tracking by graph matching model.
\newblock In {\em CVPR Workshops}, pages 29--36, 2019.

\bibitem{yang2021tracklet}
Kai-Siang Yang, Yu-Kai Chen, Tsai-Shien Chen, Chih-Ting Liu, and Shao-Yi Chien.
\newblock Tracklet-refined multi-camera tracking based on balanced cross-domain
  re-identification for vehicles.
\newblock In {\em Proceedings of the IEEE/CVF Conference on Computer Vision and
  Pattern Recognition}, pages 3983--3992, 2021.

\bibitem{zhang2015camera}
Shu Zhang, Elliot Staudt, Tim Faltemier, and Amit~K Roy-Chowdhury.
\newblock A camera network tracking (camnet) dataset and performance baseline.
\newblock In {\em 2015 IEEE Winter Conference on Applications of Computer
  Vision}, pages 365--372. IEEE, 2015.

\bibitem{zhang2022mutr3d}
Tianyuan Zhang, Xuanyao Chen, Yue Wang, Yilun Wang, and Hang Zhao.
\newblock Mutr3d: A multi-camera tracking framework via 3d-to-2d queries.
\newblock In {\em Proceedings of the IEEE/CVF Conference on Computer Vision and
  Pattern Recognition}, pages 4537--4546, 2022.

\bibitem{zhang2022bytetrack}
Yifu Zhang, Peize Sun, Yi Jiang, Dongdong Yu, Fucheng Weng, Zehuan Yuan, Ping
  Luo, Wenyu Liu, and Xinggang Wang.
\newblock Bytetrack: Multi-object tracking by associating every detection box.
\newblock In {\em Computer Vision--ECCV 2022: 17th European Conference, Tel
  Aviv, Israel, October 23--27, 2022, Proceedings, Part XXII}, pages 1--21.
  Springer, 2022.

\bibitem{zhou2020tracking}
Xingyi Zhou, Vladlen Koltun, and Philipp Kr{\"a}henb{\"u}hl.
\newblock Tracking objects as points.
\newblock In {\em Computer Vision--ECCV 2020: 16th European Conference,
  Glasgow, UK, August 23--28, 2020, Proceedings, Part IV}, pages 474--490.
  Springer, 2020.

\bibitem{zhu2018visdrone}
Pengfei Zhu, Longyin Wen, Dawei Du, Xiao Bian, Haibin Ling, Qinghua Hu, Haotian
  Wu, Qinqin Nie, Hao Cheng, Chenfeng Liu, et~al.
\newblock Visdrone-vdt2018: The vision meets drone video detection and tracking
  challenge results.
\newblock In {\em Proceedings of the European Conference on Computer Vision
  (ECCV) Workshops}, pages 0--0, 2018.

\end{thebibliography}
}

\end{document}

\title{So you think you can track? Supplementary material}

% \author{Derek Gloudemans \and Gergely Zachár \and Yanbing Wang \and Junyi Ji \and Matt Nice \and Matt Bunting \and William W. Barbour \and Jonathan Sprinkle  \and Daniel B. Work \\
% Vanderbilt University \\
% 1625 16th Ave S, Nashville, TN 37212\\
% {\tt\small derek.gloudemans@vanderbilt.edu}
% \and
\maketitle

The following appendices are included in this file to supplement the main paper.
\begin{myenum}   
    % \item Dataset File Structure
    \item Dataset Coordinate Systems
    \item Homography Re-estimation Method
    \item Homography Error Metrics
    \item Additional GPS Trajectory Plots
    \item Object Detector
    \item Experimental Details
    \item Treatment of Personally Identifiable Information 
    \item Known Data Artifacts and Anomalies
\end{myenum}

\pagebreak
\section{Dataset Coordinate Systems}
\label{sec:coordinates}

The Interstate-24 Video dataset utilizes 3 sets of coordinates utilized throughout I-24 MOTION system \cite{gloudemans202324} datasets:
\begin{itemize}
    \item \textbf{Image Coordinates:} are given in pixels. $(y^{(im)},x^{(im)})$ denotes the row and column of the specified pixel. By convention the top left pixel is (0,0).     
    \item \textbf{State Plane Coordinates:} specify a rectilinear and orthogonal coordinate system. The EPSG 2274 state plane coordinate system for Tennessee is specified in feet relative to a known survey point. $(x^{(st)},y^{(st)})$ indicates the coordinate (in feet) along the first (roughly east-west) and second (roughly north-south) coordinate axis defined by the state plane coordinate system. (Note that a common conversion from state plane coordinates to latitude/longitude coordinates (e.g. WSG84 or NAD83) can be utilized if desired.) A third orthogonal coordinate axis (z-axis) is defined and corresponds to distance off the roadway, such that $z^{(st)} = 0$ for all points on the roadway plane. 
    \item \textbf{Roadway Coordinates:} are defined such that the primary (x) axis lies along the median (or more precisely, midway between the two interior yellow lines for the interstate) at all points within the instrument extents, and the secondary (y) axis is defined locally to perpendicular to the primary axis at all points along the roadway. All coordinates with a distance from the primary axis less than the local radius of curvature (including all points on the roadway) have a unique $(x^{(r)},y^{(r)})$ coordinate. By left-hand rule convention, we  define the positive y-axis to be in the direction of the eastbound roadway lanes at all points along the roadway.
\end{itemize}

\subsection{Notation}
Throughout the rest of this appendix to disambiguate the various coordinate systems, the following notation is used: 
\begin{itemize}
    \item $x$,$y$, and $z$ refer to coordinate axes. A superscript $(im)$, $(st)$, or $(r)$  specifies all variables corresponding to a specific coordinate system (e.g. $x^{(st)}$).
    \item Vectors and matrices in a specified coordinate system are denoted in bold (e.g. $\textbf{O}^{(st)}$).
    \item Homography matrices are also listed in bold script, without superscript (e.g $\textbf{H}$). 
    \item A subscript indexes a specific point (e.g. $x^{(st)}_{bbl}$), and subscript $i$ indicates an arbitrary element index from a set of elements (e.g $a_i$). 
    \item An $x$,$y$, or $z$ without a subscript indicates a generic variable along the specified axis within the specified coordinate system. 
\end{itemize} 

\begin{table}[ht]
    \centering
    \begin{tabular}{ll}
         \toprule
         \textbf{Symbol} & \textbf{Definition} \\
         \midrule
         
         $\textbf{H}$          & 3$\times$3 matrix of homography parameters $h_{ij}$ \\ 
         $\textbf{P}$          & 3$\times$4 matrix of homography parameters $p_{ij}$ \\ 
         $s$                    & homography scale parameter \\
         $x^{(im)}$, $y^{(im)}$ & image coordinates (y indicates pixel row and x indicates pixel column)\\
         $x^{(st)}$, $y^{(st)}$, $z^{(st)}$ & state plane coordinates \\
         $x^{(r)}$, $y^{(r)}$    & roadway coordinates \\
         $\mathbf{O}^{(st)}$     & state plane coordinates for object, equal to [$\mathbf{o}^{(st)}_{bbl}$,$\mathbf{o}^{(st)}_{bbr}$,$\mathbf{o}^{(st)}_{btl}$ ,$\mathbf{o}^{(st)}_{btr}$ ,$\mathbf{o}^{(st)}_{fbl}$ ,$\mathbf{o}^{(st)}_{fbr}$ ,$\mathbf{o}^{(st)}_{ftl}$ ,$\mathbf{o}^{(st)}_{ftr}$] \\ 
         
         $\mathbf{o}^{(st)}_{bbl}$ & back bottom left state plane coordinate of object, equal to [$x^{(st)}_{bbl},y^{(st)}_{bbl},z^{(st)}_{bbl}$] \\ 
         $\mathbf{o}^{(st)}_{c}$   & back bottom center state plane coordinate, primary reference coordinate for the object \\
         $\mathbf{o}^{(st)}_{spl}$ & state plane coordinates of point on center-line spline ($y^{(r)} = 0$) with the same $x^{(r)}$ coordinate as $\mathbf{o}^{(st)}_{c}$ \\

         $\mathbf{O}^{(r)}$ & roadway coordinates for object, $[x^{(r)_o},y^{(r)_o},l,w,h]$ \\
         $x^{(r)}$          & generic longitudinal roadway coordinate along curvilinear spline axis\\
         $y^{(r)}$          & generic lateral roadway coordinate along axis locally perpendicular to longitudinal roadway coordinate axis\\
         $x^{(r)}_o$          & object longitudinal roadway coordinate along curvilinear spline axis\\
         $y^{(r)}_o$          & object lateral roadway coordinate along axis locally perpendicular to longitudinal roadway coordinate axis\\
         $l$,$w$,$h$        & rectangular prism dimensions (length, width and height) \\
         $F(x^{(r)})$          & spline defining state plane coordinate roadway center-line spline parameterized by $x^{(r)}$ \\
         $\tilde{G}(x^{(st)})$ & spline approximating the  center-line spline in roadway coordinates $x^{(r)}$ parameterized by $x^{(st)}$ \\
         \bottomrule 
    \end{tabular}
    \caption{Summary of symbols used in this section.}
    \label{tab:symbols-rcs}
\end{table}

A list of all variables along with their descriptions is given in Table \ref{tab:symbols-rcs}. Transformations between image and state plane coordinates, and transformations between state plane and roadway coordinates are detailed in the next two sections.

\subsection{Image $\leftrightarrow$ State Plane Conversion}

(parts of this subsection rely on similar definitions and descriptions to the supplement in \cite{gloudemans2023dataset}.) A \textit{homography} relates two views of a planar surface. For each camera, we provide homography information such that the 8-corner coordinates of the stored 3D bounding-box annotation can be projected into any camera view for which the vehicle is visible, creating a monocular 3D bounding box within that camera field of view.  For each direction of travel in each camera view, for each scene, a homography relating the image pixel coordinates to the state plane coordinate system is defined. (Though the same cameras are used for different scenes, the positions of the cameras changes slightly over time due). A local flat plane assumption is used (the state plane coordinate system is assumed to be piece-wise flat) \cite{hartley2003multiple}. A series of correspondence points $a_i = [x^{(im)},y^{(im)},x^{(st)},y^{(st)},z^{(st)}]$ are used to define this relation, where $(y^{(im)},x^{(im)})$ is the coordinate of selected correspondence point $a$ in pixel coordinates (row, column) and $(x^{(st)},y^{(st)},z^{(st)})$ is the selected correspondence point in state plane coordinates. 

All selected points are assumed to lie on the state plane, so $z^{(st)} = 0$ for all selected correspondence points. Visible lane marking lines and other easily recognizable landmarks on the roadway are used as correspondence points in each camera field of view. Each correspondence point is also labeled in \textit{global information system} (GIS) software, giving the precise GPS / state-plane coordinate system coordinates for each labeled corresponding point. The corresponding pixel coordinates are manually selected in each camera field of view, for each direction of travel on the roadway (see Appendix \ref{sec:correspondence}).

A \textit{perspective transform} (Equation \ref{eq:H}) is fit to these correspondence points. We first define a 2D perspective transform which defines a linear mapping (Equation \ref{eq:perspective}) of points from one plane to another that preserves straight lines. The correspondence points are then used to solve for the best perspective transform $\textbf{H}$ as defined in equation \ref{eq:H}, where $s$ is a scale factor. 

\begin{equation}
\label{eq:perspective}
    s \begin{bmatrix}
                x^{(st)}_i \\
                y^{(st)}_i \\
                1
        \end{bmatrix} 
        \sim
        \textbf{H} 
            \begin{bmatrix}
                x^{(im)}_i \\
                y^{(im)}_i \\
                1
            \end{bmatrix}
\end{equation}

where $\textbf{H}$ is a $3\times 3$ matrix of parameters:
\begin{equation}
\label{eq:H}
\textbf{H} = 
            \begin{bmatrix}
            h_{11} & h_{12} & h_{13} \\
            h_{21} & h_{22} & h_{23} \\
            h_{31} & h_{32} & h_{33}
            \end{bmatrix} 
\end{equation}

 For each camera field of view and each direction of travel, the best perspective transform $\textbf{H}^*$ is determined by minimizing the sum of squared re-projection errors according to equation \ref{eq:H_mse} as implemented in OpenCV's $find\mathunderscore homography$ function~\cite{bradski2000opencv}:
\begin{equation}
\label{eq:H_mse}
   \textbf{H}^* = \argmin_\textbf{H} \sum_i \left(x^{(st)}_i - \frac{h_{11}x^{(im)}_i + h_{12}y^{(im)}_i + h_{13}} {h_{31}x^{(im)}_i + h_{32}y^{(im)}_i + h_{33}}\right)^2 + \left(y^{(st)}_i - \frac{h_{21}x^{(im)}_i + h_{22}y^{(im)}_i + h_{23}} {h_{31}x^{(im)}_i + h_{32}y^{(im)}_i + h_{33}}\right)^2
\end{equation} 

The resulting matrix $\textbf{H}^*$ allows any point lying on the plane within the camera field of view to be converted into state plane coordinates. The corresponding matrix $\textbf{H}_{inv}$ can easily be obtained to project state plane coordinates on the $z=0$ plane into image coordinates.  However, since each vehicle is represented by a 3D bounding box, the top corner coordinates of the box do not lie on the ground plane. A 3D perspective transform $\textbf{P}$ is needed to linearly map coordinates from 3D state plane coordinate space to 2D image coordinate space, where $\textbf{P}$ is a $3\times 4$ matrix of parameters:
\begin{equation}
\label{eq:P}
\textbf{P} = 
            \begin{bmatrix}
            p_{11} & p_{12} & p_{13} & p_{14}\\
            p_{21} & p_{22} & p_{23} & p_{24}\\
            p_{31} & p_{32} & p_{33} & p_{34}
            \end{bmatrix} 
\end{equation}

and $\textbf{P}$ projects a point in 3D space $(x',y',z')$ into the corresponding image point $(x,y)$ according to:

\begin{equation}
\label{eq:P_transform}
    \textbf{P}  \begin{bmatrix}
                x^{(st)} \\
                y^{(st)} \\
                z^{(st)} \\
                1
        \end{bmatrix} 
        \sim
            s'
            \begin{bmatrix}
                x^{(im)} \\
                y^{(im)} \\
                1
            \end{bmatrix}
\end{equation}

where $s'$ is a new scaling parameter. By observing the case where $z^{(st)} = 0$, it is evident columns 1,2, and 4 of $\textbf{P}$ are equivalent to the columns of $\textbf{H}_{inv}$ and can be fit in the same way. Thus, we need only solve for column 3 of $\textbf{P}$. Next, we note as in \cite{hartley2003multiple} that $(\frac{p_{11}}{p_{31}},\frac{p_{21}}{p_{31}})$ is the vanishing point (in image coordinates) of perspective lines drawn in the same direction as the state plane coordinate x-axis. The same is true for the 2nd column and the state plane coordinate y-axis, the 3rd column and the state plane coordinate z-axis, and the 4th column and the state plane coordinate origin.

Thus, to fully determine $\textbf{P}$ it is sufficient to locate the vanishing point of the z-axis in state plane coordinates and to estimate the scaling parameter $p_{33}$. The vanishing point is located in image coordinates by finding the intersection point between lines drawn in the z-direction. Such lines are obtained by manually annotating vertical lines in each camera field of view. The scale parameter is estimated by minimizing the sum of squared reprojection errors defined in equation \ref{eq:P_mse} for a sufficiently large set of state plane coordinates and corresponding, manually annotated coordinates in image space.

\begin{align}
\label{eq:P_mse}
   \textbf{P}^* = \argmin_{p_{33}} \sum_i &\left(x^{(im)}_i - \frac{p_{11}x^{(st)}_i + p_{12}y^{(st)}_i + p_{13}z^{(st)}_i + p_{14}} {p_{31}x^{(st)}_i + p_{32}y^{(st)}_i + p_{33}z^{(st)}_i + p_{34}}\right)^2 + \nonumber \\ &\left(y^{(im)}_i - \frac{p_{21}x^{(st)}_i + p_{22}y^{(st)}_i + p_{23}z^{(st)}_i + h_{24}} {p_{31}x^{(st)}_i + p_{32}y^{(st)}_i + p_{33}z^{(st)}_i + h_{34}}\right)^2
\end{align} 
 
 The resulting 3D perspective transform $\textbf{P}^*$ allows for the lossless conversion of points in roadway coordinates to the corresponding points in image coordinates. Observing that a lossless conversion from image coordinates to state plane coordinates is available provided that the converted point lies on the $z^{(st)} = 0$ plane, it is possible to precisely convert a rectangular prism from image space to state plane coordinates by i.) converting the footprint of the prism near-losslessly into state plane coordinates (the only source of error comes from a set of 4 image coordinates that cannot be perfectly converted into a rectangle in state plane coordinates), ii.) shifting the footprint in state plane coordinates along the z-axis, iii.) re-projecting the resulting points back into the image, iv.) comparing the reprojected ``top points" to the original top of the rectangular prism in image coordinates, and v.) adjusting the height iteratively to minimize the re-projection error until convergence.

\subsection{State Plane $\rightarrow$ Roadway Coordinate System Conversion}
Next, we consider the conversion of points in state plane coordinates to roadway coordinates. In most cases, we care to convert a set of state plane coordinate points roughly in a rectangular prism (i.e. vehicle 3D bounding box) into roadway coordinates; thus, we define this conversion for a rectangular prism. A single point can be converted between state plane coordinates and roadway coordinates by treating it as a rectangular prism with zero length, width and height. 

Let $\mathbf{O}^{(st)}$ be a 3D bounding box representation in state plane coordinates, an 8$\times$3 matrix of x,y, and z coordinates for each corner of the box. (Note that these corners need not exactly correspond to an orthogonal rectangular prism, but the roadway coordinate equivalent will be exactly orthogonal so some truncation will occur.) We reference, for example, the back bottom right (from the perspective of the rear of the vehicle) of object $\mathbf{O}^{(st)}$ as $\mathbf{o}_{bbr}^{(st)} = [x^{(st)}_{bbr},y^{(st)}_{bbr},z^{(st)}_{bbr}]$, such that  = $\mathbf{O}^{(st)} =[\mathbf{o}^{(st)}_{bbl},\mathbf{o}^{(st)}_{bbr},\mathbf{o}^{(st)}_{btl},\mathbf{o}^{(st)}_{btr},\mathbf{o}^{(st)}_{fbl},\mathbf{o}^{(st)}_{fbr},\mathbf{o}^{(st)}_{ftl},\mathbf{o}^{(st)}_{ftr}]$. (For the single-point case described above, all 8 corner coordinates are identical).

Next, Let $\mathbf{O}^{(r)} = [x^{(r)}_o,y^{(r)}_o,l,w,h]$ be the corresponding object representation of $\mathbf{O}^{(st)}$ in roadway coordinates. $x^{(r)}$ and $y^{(r)}$ are the roadway coordinate longitudinal and lateral coordinates (in feet), and $l$, $w$,and $h$ are the length, width, and height of the object respectively (in feet).

Let $\mathbf{o}^{(st)_c}$ denote the back bottom center coordinate of object $\mathbf{O}^{(st)}$. By convention, this point is referenced as the primary position of object $\mathbf{O}^{(st)}$. Let $\mathbf{o}^{(st)_{spl}}$ denote the point on the center-line spline (i.e. $y^{(r)} = 0$) with the same $x^{(r)}$ coordinate as $\mathbf{o}^{(st)}_c$.

Let $F$ be the  second-order spline parameterizing the roadway center-line in state plane coordinates. In other words, $F$ defines the longitudinal curvilinear axis $y^{(r)} =0$ along this spline. $F$ is fit by manually labeling a sufficiently large number of points along the interior yellow line for both directions of travel (in state plane coordinates). A spline is fit to each yellow line, and a third spline is fit to lie precisely halfway between these two splines. Spline control points are selected at suitably sparse intervals (200 foot minimum spacing) such that the spline is relatively smooth while still capturing the roadway curvature (see Appendix \ref{sec:spline}).

Given $\mathbf{O}^{(st)}$, we first obtain $l$,$w$ and $h$ by computing the average distance between points on the front and back, left and right, or top and bottom of the vehicle respectively. Next, we obtain $\mathbf{o}^{(st)}_{c}$ by computing the average $x^{(st)}$ and $y^{(st)}$ state plane coordinates of the 4 back rectangular prism corners. 

Next, we solve for $x^{(r)}_o$ by solving the following optimization:

\begin{equation}
   x^{(r)}_o =  \argmin_{x^{(r)}} ||(F(x^{(r)}),\mathbf{o}^{(st)}_{c}||_2
\end{equation}

using L2-norm (Euclidean distance) between the two points in state plane coordinate space. In other words, determine the point on the roadway spline closest to the back center of the rectangular prism $\mathbf{o}^{(st)}_{c}$. This minimizing point is the corresponding roadway longitudinal coordinate $x^{(r)}_o$, and the distance from the minimum distance point is roadway lateral coordinate $y^{(r)}_o$.

\begin{equation}
   y^{(r)}_o =  \min_{x^{(r)}} ||(F(x^{(r)}),\mathbf{o}^{(st)}_{c}||_2
\end{equation}

Noting that the I-24 MOTION roadway segment has monotonically increasing $x^{(st)}$ coordinate, a secondary spline $\tilde{G}(x^{(st)})$ is defined to parameterize $x^{(r)}$ as a function of $x^{(st)}$, which yields a good initial guess for the closest roadway longitudinal coordinate for a given point in state plane coordinates. This optimization can then be solved to arbitrary precision, yielding the complete roadway coordinate for the object $\mathbf{O}^{(r)} = [x^{(r)}_o,y^{(r)}_o,l,w,h]$. \\

\noindent\textbf{Constant Yellow Line Constraint:} It is observed that points along the yellow line for each roadway direction of travel have non-constant y-position due to the varying width of the median. It is desirable that the yellow line (and by extension each set of lane-dividing markings) has constant y position. We finally apply the following shift:

\begin{equation}
    y^{(r)}_o \assign{+}{=}  (C - \gamma(x^{(r)}_o))
\end{equation} 

where $C$ is the desired constant yellow line y-coordinate (in this case -12 for westbound-side coordinates and +12 for eastbound-side coordinates) and $\gamma(x^{(r)}_o)$ represents the uncorrected yellow-line coordinate at the given x-position per roadway side. Note that this creates a discontinuity in the coordinate system near $y=0$ on each side, but these portions of the coordinate system are not used for objects on the roadway.

\subsection{Roadway $\rightarrow$ State Plane Conversion}

\noindent\textbf{Reverse Yellow Line Constraint:} First, the inverse yellow-line shift must be applied to the y-coordinate:
\begin{equation}
    y^{(r)}_o \assign{-}{=}  (C - \gamma(x^{(r)}_o))
\end{equation} 

Next, given roadway coordinates for an object $\mathbf{O}^{(r)}$, first find the corresponding point on the roadway center-line spline in state plane coordinates $\mathbf{o}^{(st)}_{c}$ according to:

\begin{equation}
     F(x^{(r)}_o)  = \mathbf{o}^{(st)}_{spl}
\end{equation}

To obtain the back center coordinate $\mathbf{o}^{(st)}_{c}$, we must offset $\mathbf{o}^{(st)}_{spl}$ by length $y^{(r)}$ in the direction perpendicular to the roadway centerline spline at $\mathbf{o}^{(st)}_{c}$. Let $\overrightarrow{\mathbf{u}}_{F}$ be the unit vector in the same direction as the derivative spline $F'$, and let $\overrightarrow{\mathbf{u}}_{1/F}$ be the unit vector in the perpendicular direction (along the state plane, i.e. $z^{(st)} = 0$. Note that care should be given to ensure that the positive direction of $\overrightarrow{\mathbf{u}}_{1/F}$ points towards the eastbound side of the roadway with positive $y^{(r)}$.) Then, $\mathbf{o}^{(st)}_{c}$ is given by: 

\begin{equation}
     \mathbf{o}^{(st)}_{c} = \mathbf{o}^{(st)}_{spl} + y^{(r)} \cdot \overrightarrow{\mathbf{u}}_{1/F}
\end{equation}

From here, the corner state plane coordinates for the right and left coordinates of the rectangular prism can be obtained by offsetting $\mathbf{o}^{(st)}_{c}$ by $ \pm \frac{1}{2}$ times $w$ in the direction of $\overrightarrow{\mathbf{u}}_{1/F}$, and the front coordinates of the rectangular prism can similarly be obtained by offsetting by $l$ in the direction of $\overrightarrow{\mathbf{u}}_{F}$ or in the opposite direction for  objects on the westbound or negative $y^{(r)}$ side of the roadway. Similarly, the top coordinates can be obtained by offsetting by a factor of $h$ in the $z^{(st)}$ direction. The direction of travel for an object can be obtained as the sign of the $y^{(r)}$ coordinate (negative for WB, positive for EB).For example, for an eastbound object the front top left coordinate can be obtained as:

\begin{equation}
    \mathbf{o}^{(st)}_{ftl}  = \mathbf{o}^{(st)}_{c} - \frac{1}{2} \cdot w \cdot \overrightarrow{\mathbf{u}}_{1/F} + l \cdot \overrightarrow{\mathbf{u}}_{F} + h \cdot \mathbf{[0,0,1]}
\end{equation}

\subsection{Correspondence Point Labeling}
\label{sec:correspondence}
A brief description and a few pictures of overhead and camera fields of view.

\begin{figure}
    \centering
    \includegraphics[width=\textwidth]{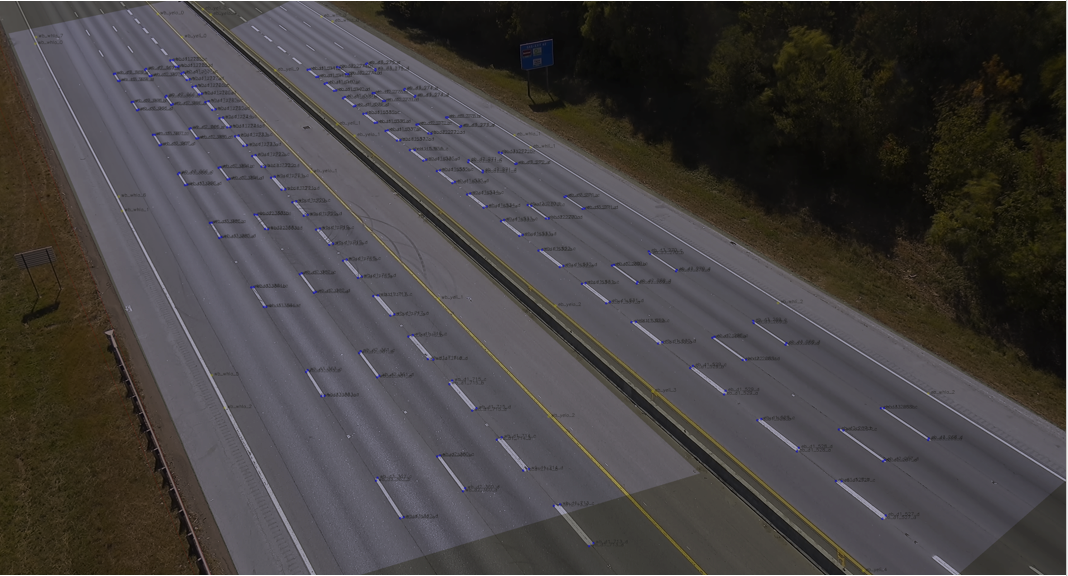}
    \caption{Correspondence points (blue) and fields of view (shaded polygons) for each roadway direction of travel, labeled in a single camera field of view (P27C01). Around 100 correspondence points are visible for each side of the roadway.}
    \label{fig:gui-corr}
\end{figure}

Correspondence points are obtained for visible lane marking lines and other easily recognizable landmarks on the roadway are used as correspondence points in each camera field of view. Labeling is carried out in a custom OpenCV GUI, with semi-automated point selection to speed this process over the 230+ cameras utilized in this work. An average frame from each camera is used to minimize the presence of occluding vehicles during labeling. Figure \ref{fig:gui-corr} shows an example of labeled points and the labeled relevant field of view for a single camera. The corresponding pixel coordinates for each manually selected point in each camera field of view, are stored, for each direction of travel on the roadway, each with a unique identifier. Typically, at least 10 roadway markings or 40 correspondence points are labeled per roadway direction of travel. Additionally, note that this initial correspondence point selection was performed for a day a few weeks prior to the day on which video is recorded in this work.

\begin{figure}
    \centering
    \includegraphics[width=\textwidth]{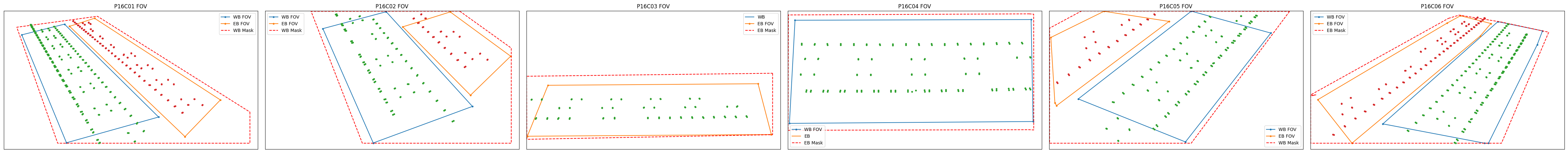}
    \caption{Correspondence points (green for WB red for EB) labeled in camera imagery for all cameras on one pole (P17). Fields of view (blue for WB and orange for EB) and a mask denoting relevant portions of the image for tracking (dotted red polygon) are also shown. Similar plots are included in a separate file for all cameras.}
    \label{fig:FOV_P16Cxx}
\end{figure}

% \begin{figure}
%     \centering
%     \includegraphics[width=\textwidth]{im/FOV_all_P16.png}
%     \caption{\Gergely{FIX THIS Pole17 Correspondence points labeled in aerial imagery using the GIS tool for a very small subset ($\sim$500 feet) of roadway, for one direction of travel only. }}
%     \label{fig:FOV_P17Cxx}
% \end{figure}

\begin{figure}
    \centering
    \includegraphics[width=\textwidth]{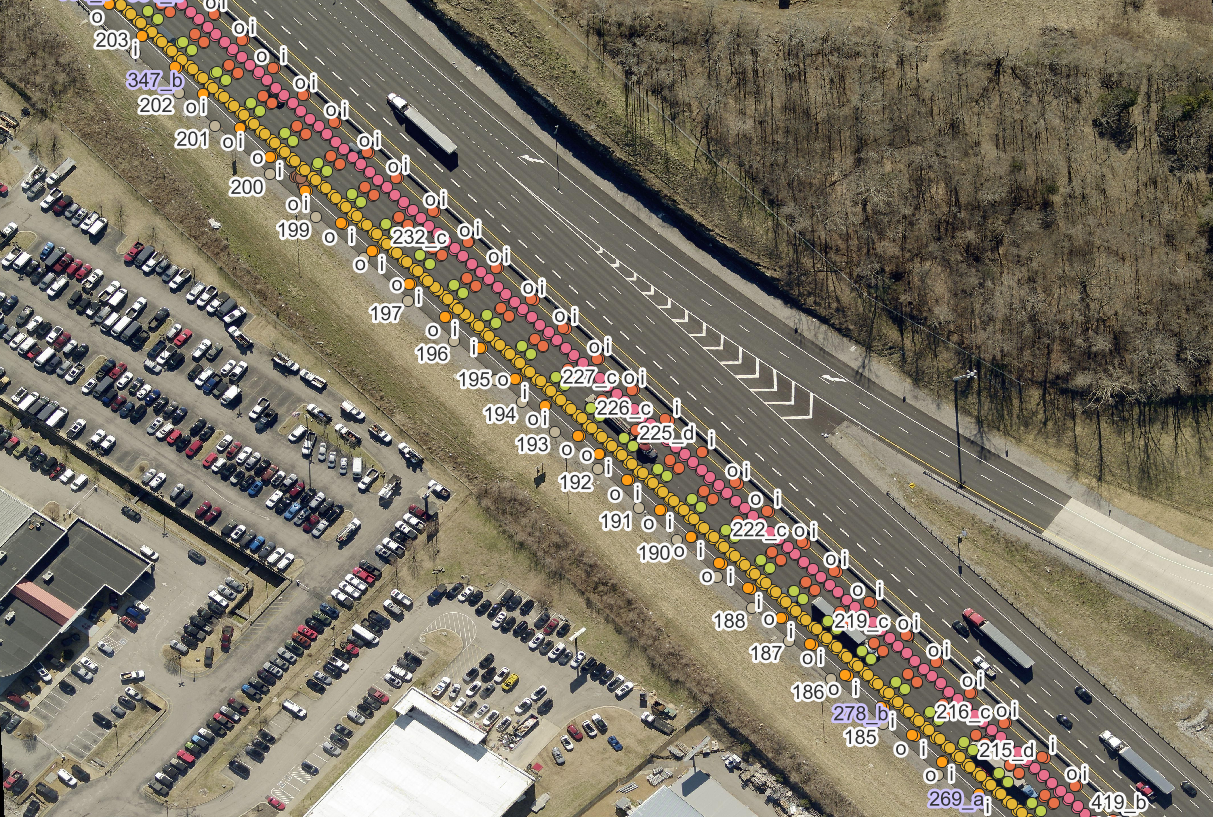}
    \caption{Correspondence points labeled in aerial imagery using the GIS tool for a very small subset ($\sim$500 feet) of roadway, for one direction of travel only.}
    \label{fig:gui-gis}
\end{figure}

Each correspondence point is also labeled in \textit{ global information system }(GIS) software, giving the precise GPS / state-plane coordinate system coordinates for each labeled corresponding point. The same unique identifier is assigned to each corresponding point previously labeled within a camera image. Additionally, points along each yellow line were labeled periodically (every $\sim$50 feet). Figure \ref{fig:gui-gis} shows an example of labeled points in the state plane aerial imagery. 

\subsection{Centerline Spline Fitting}
\label{sec:spline}

The following procedure was used to fit the center-line spline for the roadway coordinate system  ($y^{(r)} = 0$):
\begin{enumerate}
    \item For each roadway direction of travel, fit a spline parameterizing the $(x^{(st)},y^{(st)})$ points as a function of $u$. Let $F_{(WB)}(u)$ and $F_{(EB)}(u)$ denote these splines. All splines implemented in this work are based on SciPy's spline package \cite{2020SciPy-NMeth} and are constrained to have a minimum control point spacing of 200 feet to ensure smoothness.
    \item Moving along the EB yellow line spline $F_{(EB)}(u)$, sample points at fine (1 foot) intervals. For each sampled point $u$, find the closest point on $F_{(WB)}(u)$ $u'$. Store the midpoint $u*$.
    \item Fit another spline $F_{med}(u)$ to the set of midpoints from the previous step.
    \item Sample $F_{med}(u)$ at fine (0.1 foot intervals). Compute the distance between each consecutive pair, and compute the cumulative distance along the spline to each point via finite difference approximation.
    \item Re-parameterize $F_{med}(u)$ such that each point on the midpoint corresponds to the integrated distance along the spline to that point. This yields the final spline $F(x^{(r)})$ used above.
\end{enumerate}

Lastly, $\gamma(x^{(r)}))$ must be computed for each roadway direction of travel. This is done by sampling $F(x^{(r)})$ for each direction at regular (5-foot) intervals and recording the distance to the closest point on each of $F_{(WB)}$ and $F_{(EB)}$. Since these offsets change very little ($\sim$1 foot per mile), the offsets are stored in a lookup table rather than fitting a true offset function. Thus, $\gamma(x^{(r)}))$ returns the recorded offset from $F(x^{(r')})$ to $F_{(WB)}$ or $F_{(EB)}$ for the closest sampled value $x^{(r')}$.

\begin{figure*}[htb]
    \centering
    \begin{subfigure}[b]{0.49\textwidth}
        \centering
        \includegraphics[width=\textwidth]{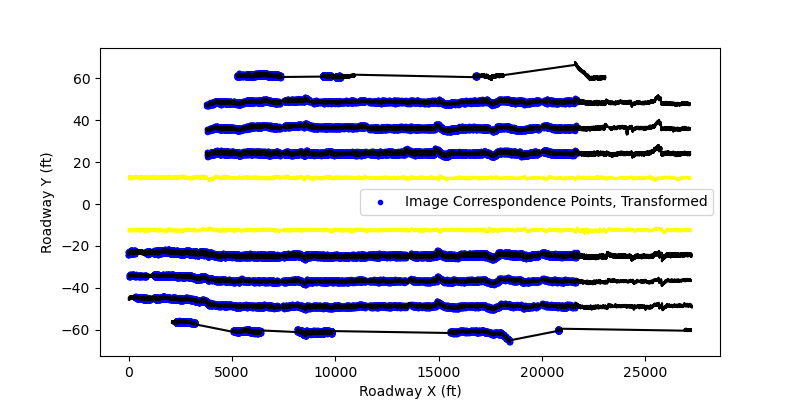}
        \caption{}
        \label{fig:splinesa}

    \end{subfigure}
    \hfill    
    \begin{subfigure}[b]{0.5\textwidth}
        \centering
        \includegraphics[width=\textwidth]{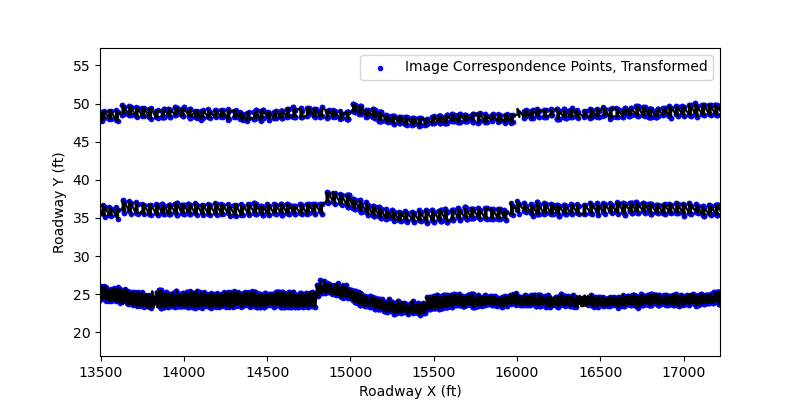}
        \caption{}
        \label{fig:splinesb}

    \end{subfigure}
    \hfill
    
    \vspace{-0.1in}
    \caption{(a) correspondence points labeled in aerial imagery (yellow and black) and in individual camera fields of view (blue) are projected into the roadway coordinate system. (b) A close-up detailing a type of artifact visible in the coordinate system as a result of misalignment in the aerial imagery.}
    \label{fig:splines}
    \vspace{-0.2in}
\end{figure*}

Figure \ref{fig:splinesa} shows the labeled points from aerial imagery (yellow and black lines) and the points labeled in each image (blue dots), transformed into roadway coordinates. Figure \ref{fig:splinesb} shows the largest magnitude shift in labeled correspondence points, when converted to the roadway coordinate system. Such ``discontinuities'' are due to slight misalignments betweeen aerial images taken during different passes of the photographing aircraft, and result in small (less than 2 foot) errors in all cases. While these misalignments could be corrected for by applying a smoothing to the correspondence points labeled within the aerial imagery, this was ultimately decided against because the smoothing would need to be performed with respect to, essentially, the roadway coordinate system spline itself, which is in turn a product of the labeled aerial imagery points. This would produce complex and less well-understood artifacts in the resulting final coordinate system.

\FloatBarrier
\pagebreak

\section{Homography Re-estimation Methods}
\label{sec:re-estimation}
\begin{figure}[h]
    \centering
    \begin{subfigure}[b]{0.32\textwidth}
        \includegraphics[width=\textwidth]{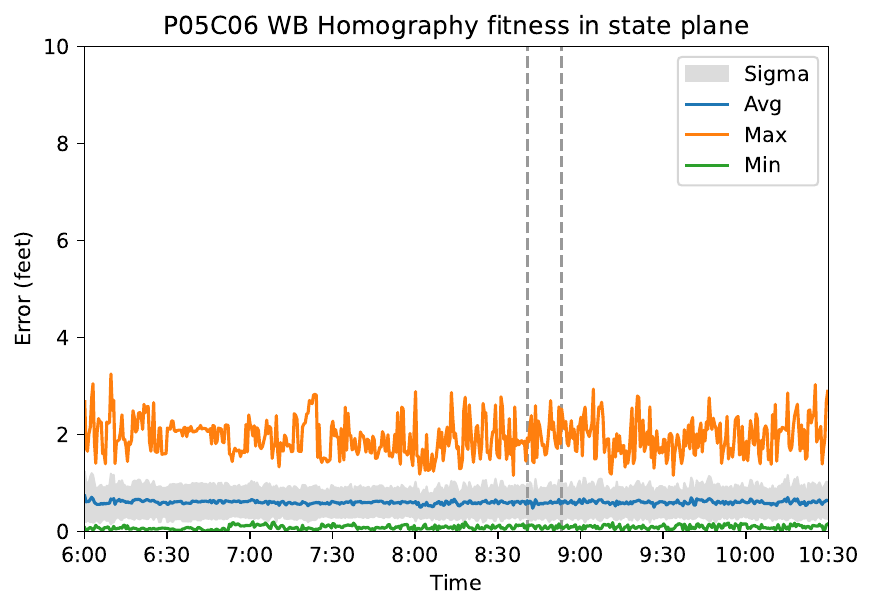}
        \caption{}
        \label{fig:error_fitness_P05C06f}
    \end{subfigure}
    \hfill
    \begin{subfigure}[b]{0.32\textwidth}
        \includegraphics[width=\textwidth]{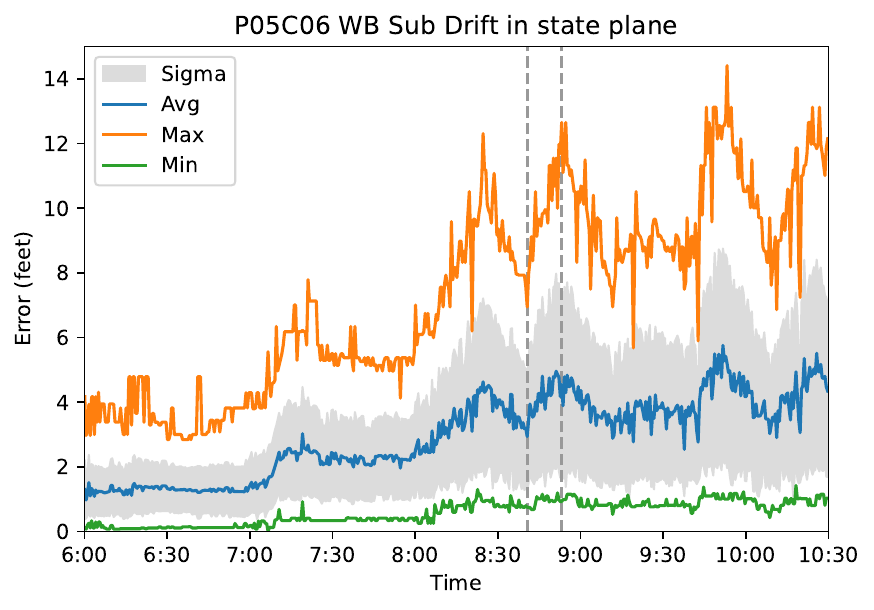}
        \caption{}
        \label{fig:error_subdrift_P05C06f}
    \end{subfigure}
    \hfill
    \begin{subfigure}[b]{0.32\textwidth}
        \includegraphics[width=\textwidth]{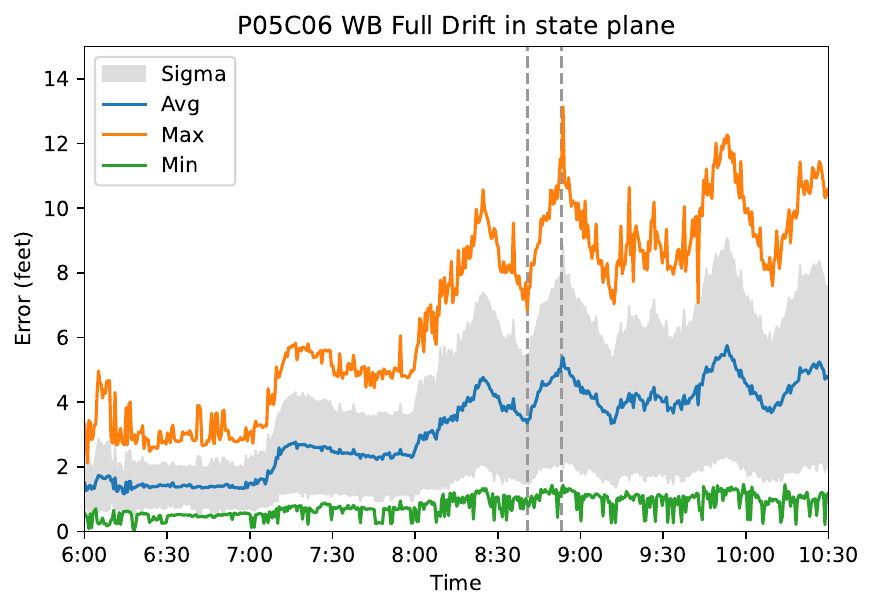}
        \caption{}
        \label{fig:error_fulldrift_P05C06f}
    \end{subfigure}
    \caption{Homography goodness of fit and uncorrected drift between reference homography and true scene homography according to two metrics (Sub Drift and Full Drift) for a single typical camera. Horizontal grey dashes show the two time instances for Figure \ref{fig:homo_error_mapf}.}
    \label{fig:error_P05C06f}

\end{figure}

\begin{figure}[h]
    \centering
    \begin{subfigure}[b]{0.49\textwidth}
        \includegraphics[width=\textwidth]{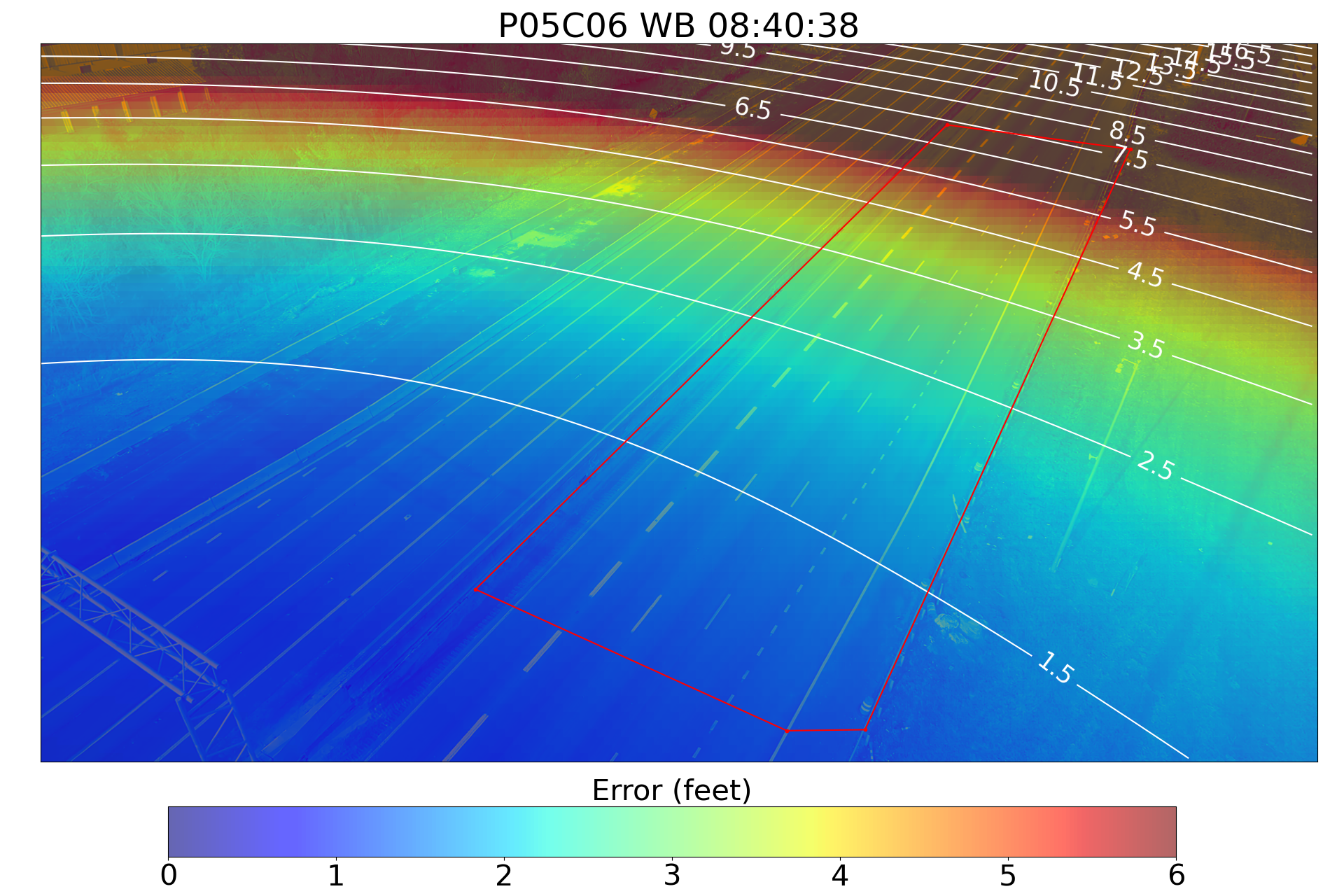}
        \caption{}
        \label{fig:homo_error_map_840f}
    \end{subfigure}
    \hfill
    \begin{subfigure}[b]{0.49\textwidth}
        \includegraphics[width=\textwidth]{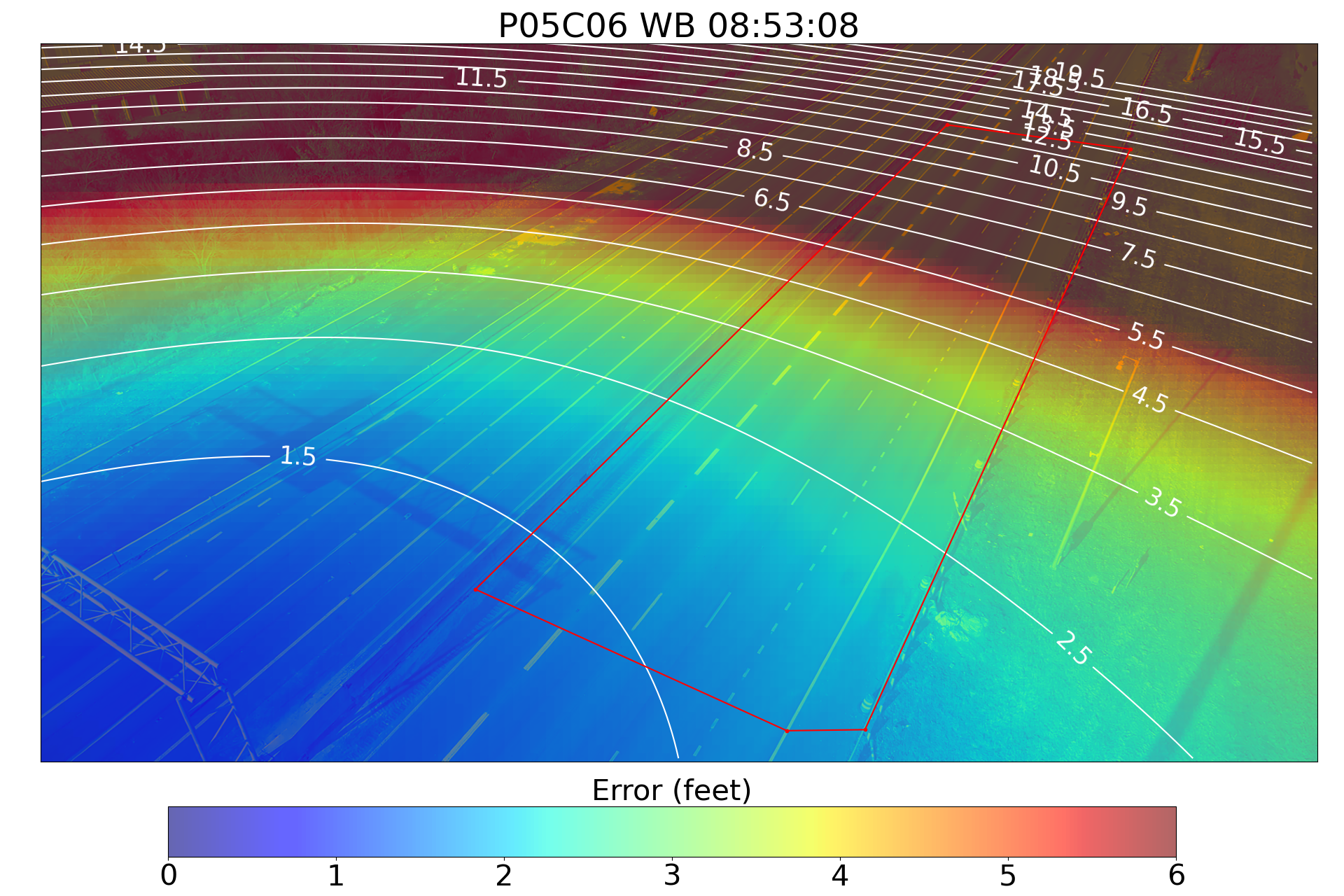}
        \caption{}
        \label{fig:homo_error_map_853f}
    \end{subfigure}

    \caption{Full drift for the same camera at two time instances. For a given error value $X$, this indicates that the original reference homography and the true homography for the scene map the same image coordinate to points in the state plane $X$ feet apart. The westbound field of view (i.e. the most important portion of the image) is shown as a red polygon.}
    \label{fig:homo_error_mapf}

\end{figure}

In ideal circumstances homographies can be estimated once (as discussed in the previous section), and used continuously until some drastic change in the system, (e.g. the roadway section is rebuilt or the lane markings are repainted). Aside from these rare events there are several other factors which significantly degrade accuracy; long term events can be the replacement of the camera, inaccuracies in the pan-tilt mechanism during homing, settlement of the foundation or the seasonal temperature change. Although these can be dealt with occasional re-calibration, the \textit{sunflower effect} requires constant re-estimation of the homographies. This effect, the tilting of the metal infrastructure poles due to the differential heating of the sun and shade-facing sides, can cause significant homography errors (sometimes greater than 10 feet, see Figure \ref{fig:error_P05C06f} for a typical camera example), both in timescales of hours for the daily warm up and cool down cycle, and also in the timescales of minutes, caused by the varying cloud coverage (see Figure \ref{fig:homo_error_mapf}, showing how the positional drift in homographies varies over just 12.5 minutes). A video is included in supplementary material showing the magnitude of drift over the course of a day. The camera movement can easily be seen by viewing the relation between the lane markers and the ROI rectangle (which maintains constant pixel coordinates throughout the video). For reference, a typical dash line is 10 feet long.

Note, that tilting not only degrades single-camera accuracy, but due to the multiple pole architecture and the camera FOVs, homography errors cause significant misalignment in state plane coordinates between poles, resulting in vehicle tracking fragmentations . (Cameras on the same pole are less susceptible because they are rigidly mounted and roughly "moving" together.) Thus, the re-estimation of the homography is necessary on relatively short timescales, or at a minimum on a daily basis.

Since manual re-annotation is not possible for continuous operation, a feature point re-identification method is developed, along with re-estimation methods for a static daily and for a dynamic, time-dependent homography.

All methods and filtering steps are based on the assumption that the image is only slightly distorted from the manually labeled ground truth. This assumption is necessary, because the homography and the re-discovery relies on the lane markings which are non-unique and repetitive, thus a significant shift can cause a mis-identification and misalignment. All presented methods are \textit{offline} (they utilize both past and future information relative to the specified time instance (i.e. use a non-casual filter)).

Automated homography generation is divided into two distinct steps. First the feature points are re-discovered and a new homography is calculated for a single time instance. Second, a refinement step operates on the homographies themselves of which removes outliers, filters, and aggregates over time. This approach provides robustness for the final estimates based on the assumption that the re-estimation provides good results \textit{in general} but are prone to outliers and errors. To comply to this idea the implemented refinement processes are draconian, and tuned to minimize false positives at the expense of occasionally removing some true positives. The "good in general" assumption is empirically tested, by rigorously inspecting the results and fine tuning the filter parameters of each steps. In case of faulty estimates manual inspection always shows problems which could cause confusion even for human observers or significant noise and drop in image quality (e.g.: at dawn).   

\subsection{Notation}
The following notation convention is used throughout the remainder of this Section. Also see Table \ref{tab:symbols} for a list of utilized symbols and their meanings.

\begin{myitem}   
    \item Mathcal script (e.g. $\mathcal{I}$) is used for sets of points. Bold Text (e.g. $\textbf{H}$) is used for vectors and matrices. 
    \item The subscript $t$ denotes a set of points or a homography for a specific time instance (e.g. $\mathcal{I}_t$)
    \item No subscript denotes the initial reference points or homography (e.g. $\mathcal{I}$, $\mathcal{A}$, and $\textbf{H}$)
    \item The prime symbol denotes rediscovered points (e.g. $\mathcal{I}'_t$, which corresponds to subset $\mathcal{A}_t$) , or homographies based on rediscovered points (e.g.: $\textbf{H}_t^{\prime}$)
    \item a superscript $^s$ denotes an image-to-image homography based on SIFT-FLANN matching (e.g. $\textbf{H}_t^s$)
    \item A superscript asterisk symbol $\ast$ denotes a homography \textit{estimate} calculated with methods presented in the previous appendix (e.g.: $\textbf{H}^{\ast}$).
    \item An arrow $\mathcal{I}\xrightarrow{\textbf{H}}$ indicates a linear transformation on the point set $\mathcal{I}$ according to homography $\textbf{H}$. (e.g. $\mathcal{I}_t' \xrightarrow[]{\textbf{H}_t'}$ $\equalhat$ $\mathcal{A}_t$ denotes that the set $\mathcal{I}_t'$ of rediscovered points at time $t$, projected from image coordinates to state plane coordinates by $\textbf{H}_t'$, estimates the set of corresponding points labeled in state plane coordinates $\mathcal{A}_t$).
\end{myitem}

\begin{table}[h]
\renewcommand{\arraystretch}{1.5} % Default value: 1
    \centering
    \begin{tabular}{ll}
         \toprule
         \textbf{Symbol} & \textbf{Definition} \\
         \midrule
$\mathcal{I}$ & labeled image points (in image coordinates, unit: pixels) \\
$\mathcal{I}_t$ & subset of the labeled image points at time $t$, corresponding to the successfully rediscovered points ($ \mathcal{I}_t \subset \mathcal{I} $) \\
$\mathcal{I}_t^{\prime}$ & rediscovered image points ($|\mathcal{I}_t^{\prime}| \leq |\mathcal{I}|$) \\
$\mathcal{I}_t^s$ & labeled image points after perspective transformation, utilizing the SIFT-FLANN matcher, such that $\mathcal{I} \xrightarrow{\textbf{H}_t^s} \mathcal{I}_t^s$ \\
$\mathcal{I}_t^{\ast}$ & "labeled" image points generated such as $A \xrightarrow{\textbf{H}_t^{\ast\neg}} \mathcal{I}_t^{\ast}$ \\
$\mathcal{A}$ & aerial points (in state plane coordinates, unit: feet) \\
$\mathcal{A}_t$ & subset of the aerial points at time $t$, corresponding to the successfuly rediscovered points ($\mathcal{A}_t \subset \mathcal{A}$) \\ \midrule       
$\textbf{H}$ & homography such that $\mathcal{I} \xrightarrow{\textbf{H}} \mathcal{A}$ \\
$\textbf{H}^{\neg}$ & inverse homography such that $\mathcal{A} \xrightarrow{\textbf{H}^{\neg}} \mathcal{I}$ \\
$\textbf{H}_t^{\prime}$ & homography such that $\mathcal{I}_t^{\prime} \xrightarrow{\textbf{H}_t^{\prime}} \mathcal{A}_t$ \\
$\textbf{H}^{\ast}$ & new \textit{static homography estimate} by method from Appendix \ref{homo:static} \\
$\textbf{H}_t^{\ast}$ & new \textit{dynamic homography estimate} for time $t$ by methods from Sections \ref{homo:dynamic}-\ref{homo:SIFT} \\
$\textbf{H}_t^s$ & projection (image to image) generated with the SIFT-FLANN matcher results \\         \bottomrule 
    \end{tabular}
    \caption{Notations for various point sets and homographies utilized in this work.}
    \label{tab:symbols}
\end{table}

\subsection{Homography Generation for a Single Time Instance}
\label{homo:single-time}
This section briefly describes the steps necessary for the homography generation for a single time instance $t$. 

\begin{enumerate}
    \item Generate a background extracted image to remove the vehicles from the scene. This is implemented as a 1 minute long averaging of frames, with 50\% overlap in time. Note that in case of heavy traffic (essentially for stopped vehicles) this time period might not be sufficient for complete removal.
    \item Compute an initial alignment estimate between the frame and the reference frame, on which the original annotations were made. This step utilize the SIFT algorithm to find feature points and a FLANN based matcher (both implemented in OpenCV). Note that the detected features can be anywhere on the image, not necessarily corresponding to semantically meaningful points (e.g. grass, trees, parking cars outside the road), and most importantly generally not lie on the plane of the roadway.
    \item Based on the SIFT-FLANN generated corresponding points generate an image-to-image homography ($\textbf{H}_t^s$). 
    \item Examine $\textbf{H}_t^s$: transformation (e.g.: translation, rotation, scale) should be minimal, based on the slight distortion assumption. Otherwise discard the current time instant, because the calculated alignment is likely from erroneously matched points. No further steps are possible.
    \item Binarize the averaged frame via OpenCV's $threshold$ function \cite{bradski2000opencv} and run a contour detection algorithm. This provides good results because lane markings are high contrast, distinct features.
    \item Transform the originally labeled feature points $\mathcal{I}$, i.e. the four corner points of the lane markers, with the calculated $\textbf{H}_t^s$ (producing $\mathcal{I}^s_t$) and calculate the geometric center for each marker. This provides a seed point for rediscovery.
    \item Select contour areas which include a seed point, thus creating a set of candidate for lane marks.
    \item For each marker, select the 4 corner points on the corresponding contour that produce the largest area, thus creating a quadrilateral.
    \item Filter the candidates by contour area, contour area to quadrilateral area ratio, ratio of area compared to the labeled reference area. This step is important to remove candidates which are merged, partially occluded, or otherwise not rediscovered correctly.
    \item Select the proper labels and store the 4 corner points as feature points of each remaining markers. With this step the rediscovery steps are complete, yielding $\mathcal{I}'_t$.
    \item Run a homography matrix estimator (which utilizes RANSAC) between the rediscovered $\mathcal{I}'_t$ and the corresponding aerial $\mathcal{A}_t$ points. Remove the feature points which are considered outliers by the algorithm. In the current implementation these are points of which are further than \~2 feet on the state plane. The resulting homography for time instance $t$ is written as $\textbf{H}_t^{\prime}$.

\end{enumerate}

\begin{figure*}
    \centering
    \includegraphics[width=\textwidth]{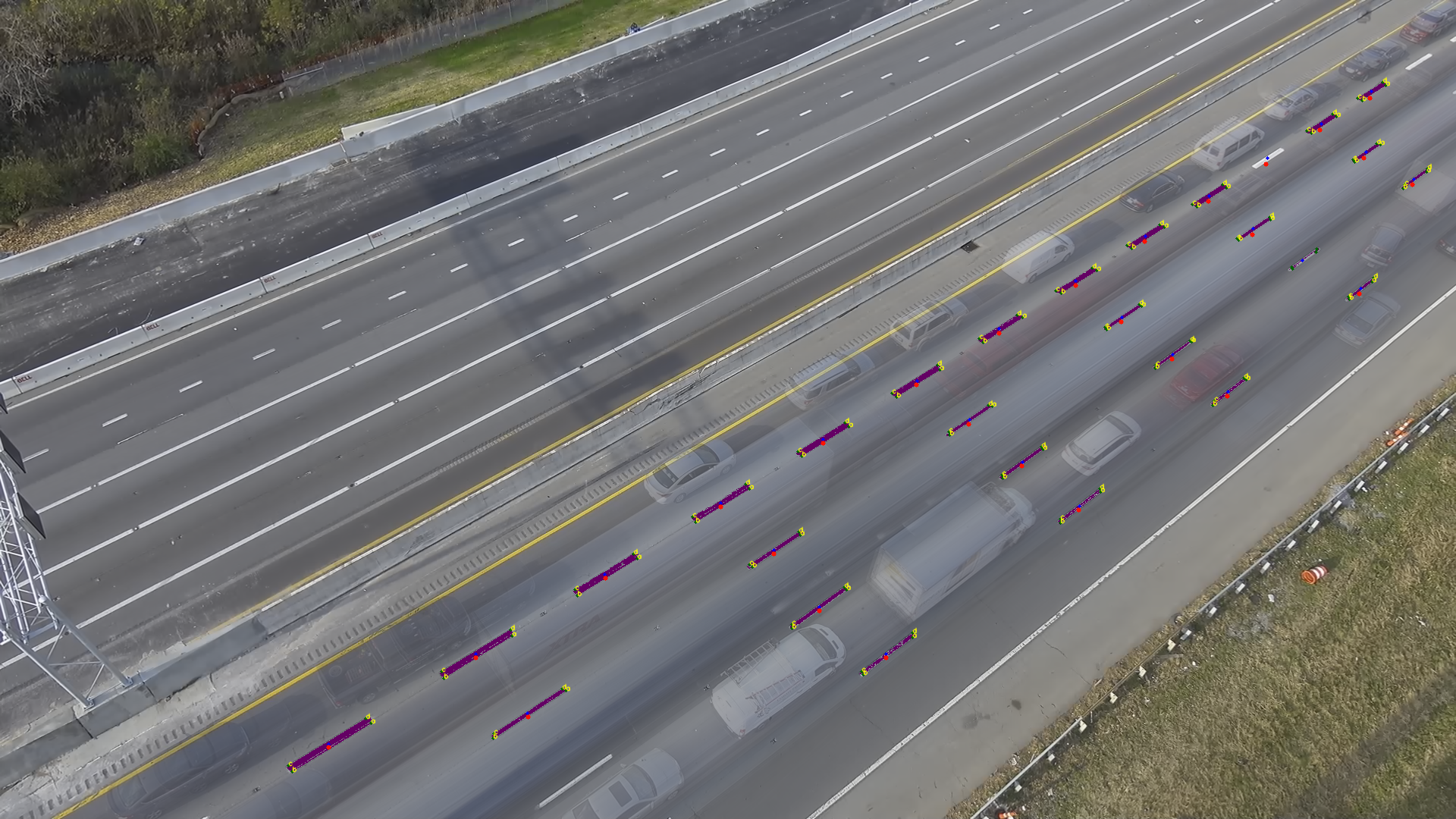}
    \label{}
    \caption{An example of SIFT-FLANN based seed points from step 6 (red dots), detected contours (purple outlines) and re-detected dash corner points (yellow dots) for a 1-minute average frame. Because traffic at this time instance was partially stopped, some vehicle artifacts are visible in the averaged frame. A few dashes are not successfully rediscovered.}
\end{figure*}

The result of the listed process is a set of rediscovered feature points ($\mathcal{I}'_t$)  and a new homography ($\textbf{H}_t^{\prime}$) for a given time instant $t$. Note that although the steps involve heavy filtering and outlier detection the resulting, a standalone homography should not be used without further processing for the following reasons:
\begin{itemize}
\item The number and positions of the rediscovered lane markings are not guaranteed, thus it is possible to only rediscover lane markings which lie on the same line, thus providing poor estimates for the perpendicular axis.
\item Lane markings "far" from the camera are hard to properly detect due to their small area and slight camera movements can cause misidentification.
\item The RANSAC based homography estimator does not guarantee that the selected points are proper. In extreme cases it can select an erroneous subset and fit a good homography to it. As discussed previously, the lane markings are non-unique and repetitive: usually there are multiple subsets of points with nearly identical relative arrangement.
\end{itemize}

\subsection{Static Homography Generation}
\label{homo:static}
To counter for the long term homography errors and somewhat compensate for the sunflower effect a single homography can be generated for a given time interval, e.g.: the length of the recording or for the desired time range. This method has the advantage of time-invariance (simplicity) and can utilize all the instantaneous estimates over the specified window, resulting in ample redundancy for outlier removal.
The method includes two steps:
\begin{enumerate}
\item In the preparation phase the outliers are removed, i.e. the homography estimates of which significantly differ from the others. The current implementations considers a homography an outlier if any of its component (in the 3x3 matrix) deviates more than 30\% from the arithmetic mean of all estimates. Note that this process is iterative.
\item The main step is the averaging, i.e. calculating the arithmetic mean for each matrix component to produce $\textbf{H}^{\ast}$.
\end{enumerate}

\subsection{Dynamic Homography Generation}
\label{homo:dynamic}
To counter for both long and short term homography problems a dynamic, time dependent method is proposed. This solution is capable of modeling short-term camera motion (e.g. from the sunflower effect), but is less trivial implement and use compared to the static version because it requires modifying the homography as a function of time.

The underlying idea of this method is the "smoothing" of the original estimates, based on the temporally nearby values. Note that this is not straightforward because estimates are not necessarily available at all time instances, e.g.: in case of heavy traffic the occlusion prevents homography estimation, similarly at the beginning and at the end of the recording there is no preceding and subsequent information available. (This is especially problematic at dawn and dusk where the even the recorded images are noisy and blurred thus making feature rediscovery hard or even impossible.) Thus a method is necessary of which is capable to adapt to these situations. An important and necessary property of a good dynamic homography is that it provides "smooth" variability over time, preventing discontinuities in positional data (which could in turn cause fragmentations during object tracking.)

The proposed solution includes three steps:
\begin{enumerate}
    \item In the preparation phase the outliers are removed. This step is exactly the same as in the static case.
    \item A "window size" parameter is computed, based on the available number of homography estimates in a specified temporal neighbourhood around the given time instance. This step ensures that in case of missing estimates the window size is larger, thus accommodating more data points.
    \item The final step is the "smoothing" of each homography matrix component, where a Gaussian kernel function is used over time. The shape of the function, or more precisely the variance is determined by the previous step.
\end{enumerate}

Note that this method can provide a homography estimate $\textbf{H}_t^\ast$ at any time instance, not just when instantaneous estimates are available. In our implementation a new estimate is generated at 10 second intervals.

\subsection {SIFT-FLANN-based homography generation (existing)}
\label{homo:SIFT}

% No lane marking rediscovery is performed. Instead, the set of shifted points from Section \ref{homo:single-time} step 6 above ($\mathcal{I}_t^s$) is directly used to fit a homography estimate $\textbf{H}_t^\ast$ for time $t$ such that $\mathcal{I}_t^s \xrightarrow[]{\textbf{H}_t^\ast} \mathcal{A}$.

No lane marking rediscovery is performed. Instead, the set of shifted points ($\mathcal{I}_t^s$) generated from the original labels ($\mathcal{I}$) is direcly used from Appendix \ref{homo:single-time} step 6 and above to fit a homography estimate $\textbf{H}_t^\ast$ for time $t$ i.e. $\mathcal{I} \xrightarrow[]{\textbf{H}_t^s} \mathcal{I}_t^s \xrightarrow[]{\textbf{H}_t^\ast} \mathcal{A}$.

\FloatBarrier
\pagebreak

\section{Homography Error Metrics}
\label{sec:metrics}

Next, we turn to defining metrics for assessing the effectiveness and accuracy of the proposed point rediscovery and homography re-estimation metrics. This appendix provides an overview of the possible metrics and their properties.

\subsection{Notation}
In addition to the notation defined in the previous section, the comparison operator $E$ is defined as a function of two sets of points ($\mathcal{B}, \mathcal{C} \subset \mathbb{R}^2 $). The operator calculates the point-wise L2-norm between the common subset of points shared by $\mathcal{B}$ and $\mathcal{C}$ , (i.e. the distances between two points which correspond to the same feature), and produces a set of values. The resulting set can be used to calculate aggregate metrics, such as mean, maximum, and standard deviation of error. 

% \begin{equation}
%     E(\mathcal{B}, \mathcal{C}) = \{ ||b_i, c_i ||_2, ..., || b_n, c_n ||_2 \}, b_i\in\mathcal{B}, c_i\in \mathcal{C} \text{  s.t.  } b_i \equalhat c_i
% \end{equation}

\begin{equation}
    E(\mathcal{B}, \mathcal{C}) = \{ ||b_i, c_i ||_2, ..., || b_n, c_n ||_2 \} \: \forall i \text{\: s.t. \:} b_i\in\mathcal{B}, c_i\in \mathcal{C},b_i \equalhat c_i
\end{equation}

% \begin{equation}
%     E(\mathcal{B}, \mathcal{C}) = \{ ||b_i, c_i ||_2, ..., || b_n, c_n ||_2 \} \; \forall i \text{\: s.t. \:} i\in (\mathcal{B} \cap \mathcal{C}),b_i \equalhat c_i
% \end{equation}

%Error function use the the intersection of the input points, thus subsets are not indicated

\subsection{Reference Metrics}

Metrics listed in this appendix are computed by comparing the original reference labels ($\mathcal{I}$ and $\mathcal{I}$) to rediscovered points ($\mathcal{I}_t'$). In some cases the same set of points is transformed with the original reference homography ($\textbf{H}$ or $\textbf{H}^\neg$) and the instantaneous homography estimate $\textbf{H}_t'$.   They measure i.) the quality of the rediscovery and instantaneous homography re-estimation process, and ii.) the movement of the cameras. 

Most metrics presented here can be calculated in both image and state plane coordinate systems; thus we present them in pairs.  The resulting error values have units of pixels or distance (feet), respectively. In practical evaluations we utilize the state plane distance (units of feet) as it is invariant to the image resolution and FOV, thus comparable across cameras. As in Appendix \ref{sec:coordinates}, we use a superscript $(im)$ or $(st)$ to denote the coordinate system for each metric.
%\Gergely{This is only necessary if we keep Fitness II for completeness.}

\begin{itemize}

\item \textbf{Sub Drift I:} This metric compares a subset of the originally labeled \textit{image} points to the rediscovered points. The resulting distances can be interpreted as the \textit{drift} caused by the camera movement. Note that in case of perfect alignment the maximum error is zero. % It is true because H doesn't change and I_t exactly equals I.
\begin{equation}
    \text{SubDriftI}^{(im)} = E(\mathcal{I}_t, \mathcal{I}_t^{\prime})
\end{equation}
\begin{equation}
    \boxed{\text{SubDriftI}^{(st)}= E(\mathcal{I}_t \xrightarrow{\textbf{H}}, \mathcal{I}_t^{\prime} \xrightarrow{\textbf{H}})}
\end{equation}

\item \textbf{Sub Drift II (not used):} This metric compares a subset of the originally labeled \textit{aerial} points to the rediscovered points. The resulting distances are a combined error of the drift caused by the camera movement \textit{and} the homography fitness error. Thus even in case of no movement the error is non-zero, as the homography is not a perfectly fit mapping of $\mathcal{I}$ to $\mathcal{A}$.

\begin{equation}
    \text{SubDriftII}^{(im)} = E( \mathcal{A}_t^{\prime} \xrightarrow{\textbf{H}^{\neg}}, \mathcal{I}_t^{\prime} )    
\end{equation}
\begin{equation}
    \text{SubDriftII}^{(st)} = E(\mathcal{I}_t^{\prime} \xrightarrow{\textbf{H}}, \mathcal{A}_t )
\end{equation}

\item \textbf{Full Drift:} This metric compares the full set of the originally labeled image or aerial points transformed by the reference and the re-estimated homography. The resulting distances can be interpreted as \textit{drift} caused by the camera movement. In the case of no camera movement, the error is zero (except for possible stochasticity in the optimal homography $H'_t$ selected by the RANSAC algorithm utilized.)

\begin{equation}
    \text{FullDrift}^{(im)} = E( \mathcal{A} \xrightarrow{\textbf{H}^{\neg}},  \mathcal{A} \xrightarrow{\textbf{H}_t^{\prime\neg}} )    
\end{equation}
\begin{equation}
    \boxed{\text{FullDrift}^{(st)} = E( \mathcal{I} \xrightarrow{\textbf{H}}, \mathcal{I} \xrightarrow{\textbf{H}_t^{\prime}} )}
    \label{eq:full-drift}
\end{equation}

\item \textbf{ Sub Fitness: } This metric shows the fitness of the estimated homography, utilizing the rediscovered points and a matching subset of aerial points. This error is caused by a combination of e.g.: inaccurate feature point rediscovery, lens distortion, non flat roadway.

\begin{equation}
    \text{Fitness}^{(im)} = E( \mathcal{A}_t \xrightarrow{\textbf{H}_t^{\prime\neg}}, \mathcal{I}_t^{\prime} )
\end{equation}
\begin{equation}
    \boxed{\text{Fitness}^{(st)} = E( \mathcal{I}_t^{\prime} \xrightarrow{\textbf{H}_t^{\prime}}, \mathcal{A}_t )}
    \label{eq:sub-fitness}
\end{equation}

\end{itemize}

% \item Sub Fitness? (not used): \Gergely{Probably we should remove.. it is here for completeness.. oh and the image coordinate system error does not exists..}
% Fitness of the homography, using only homographies??

% $[ A \xrightarrow{H^{\neg}}, ? ]$

% $[ I \xrightarrow{H}, I_t^{\prime} \xrightarrow{H_t^{\prime}}]$

\subsection{Homography Re-estimation Metrics}

Metrics listed here are derived from the metrics introduced at the previous section. Substituting the reference labels and homographies with the re-estimated homographies and points sets derived from them. The interpretation and purpose of these metrics are to qualify how good are the new estimations; i.e. a perfect homography provides a perfect alignment to the detections.

\begin{itemize}

\item \textbf{Sub Drift:} This metric compares a subset of points generated with the re-estimated homography (from the originally labeled points) to the rediscovered points. The resulting distances can be interpreted as a combined error of the re-estimation and the homography fitness error. Therefore even in case of a perfectly aligned  homography the error is non-zero. Note that, because the new subset of points are generated from the reference labeled points this metric essentially merge Sub Drift I and II presented in the previous section.

\begin{equation}
    \text{SubDrift}^{(im)} = E(\mathcal{I}_t^{\ast}, \mathcal{I}_t^{\prime}) \Longrightarrow E( \mathcal{A}_t \xrightarrow{\textbf{H}_t^{\ast\neg}}, \mathcal{I}_t^{\prime} )
\end{equation}
\begin{equation}
    \text{SubDrift}^{(st)} = E( \mathcal{I}_t^{\ast} \xrightarrow{\textbf{H}_t^{\ast}}, \mathcal{I}_t^{\prime} \xrightarrow{\textbf{H}_t^{\ast}} ) \Longrightarrow E( \mathcal{A} \xrightarrow{\textbf{H}_t^{\ast\neg}} \xrightarrow{\textbf{H}_t^{\ast}}, \mathcal{I}_t^{\prime} \xrightarrow{\textbf{H}_t^{\ast}} ) \Longrightarrow E( \mathcal{A}_t, \mathcal{I}_t^{\prime} \xrightarrow{\textbf{H}_t^{\ast}} )
\end{equation}

\item \textbf{Full Drift:} This metric compares the full set of the originally labeled image or aerial points transformed by the the re-estimated  and by the instantaneous homography. In effect, these original labeled points are chosen as ``proxy points'' for the real locations of the correspondence points in the image as they are a full set known to lie near the true locations. The resulting distances can be interpreted as the error of the homography re-estimation process. Note that in case of a perfect alignment the error is zero.

\begin{equation}
    \text{FullDrift}^{(im)} = E(  \mathcal{A} \xrightarrow{\textbf{H}_t^{\ast\neg}},  \mathcal{A} \xrightarrow{\textbf{H}_t^{\prime\neg}} )
\end{equation}
\begin{equation}
    \boxed{\text{FullDrift}^{(st)} = E( \mathcal{I} \xrightarrow{\textbf{H}_t^\ast}, \mathcal{I} \xrightarrow{\textbf{H}_t^{\prime}}) }
\end{equation}

\item \textbf{ Fitness: } Since the fitness metrics does not depend on the re-estimated homography, they are equivalent to the ones discussed in the previous section.

\end{itemize}

The homography re-estimation performance can be measured by the Full and Sub Drift metrics, but both have caveats to consider: Sub Drift is a better measure of the \textit{ real magnitude of the error}, because it incorporates inaccuracies of both the homography and the homography fitness, e.g. errors caused by intrinsic camera problems and the non-flat roadway. On the other hand the Sub Drift metric only includes feature points which are re-detected, thus hard to detect points further from the pole are often missing from the set. This is crucial because the perspective those "far" points account for larger state plane errors than the more easily detectable closer ones. The combined effect of these that the Sub Drift is better in the assessment of the expected minimal error, this is not the case for the maximum. Also worth noting is that Sub Drift does not explictly rely on $H'_t$ being well-fit, so produces a reliable estimate of error even when the fitting of $H'_t$ fails (though in practice this usually indicates poor point rediscovery).

The Full Drift metric is useful to assess the performance of the \textit{homography re-estimation} itself, because in case of perfect alignment the error is reduced to zero. In opposite to the Sub Drift this, metric includes all labeled points, some of which are actually outside the processed ROI, thus resulting in an apparently larger maximum error than during vehicle detection.

Based on these considerations, we select Full Drift (Equation \ref{eq:full-drift}) as our primary metric and use it to report the $error$ for each method utilized as equation 2 the main text. We also utilize Sub Fitness (Equation \ref{eq:sub-fitness}) as a baseline measure for the error floor of each fit homography ($fitness$ in equation (1) of the main text).

\subsection{Additional Homography Metric Figures}

\begin{figure}[h]
    \centering
    \begin{subfigure}[b]{0.32\textwidth}
        \includegraphics[width=\textwidth]{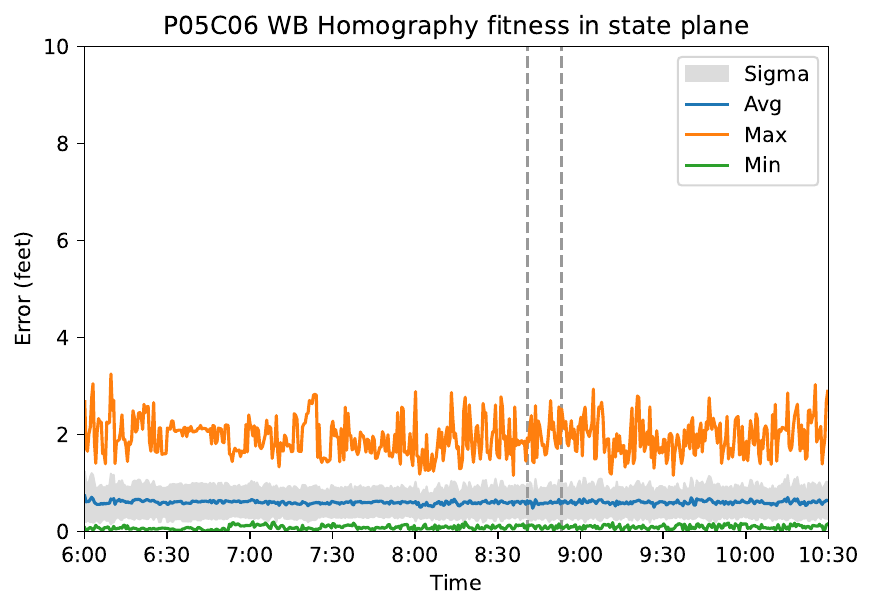}
        \caption{}
        \label{fig:error_fitness_P05C06}
    \end{subfigure}
    \hfill
    \begin{subfigure}[b]{0.32\textwidth}
        \includegraphics[width=\textwidth]{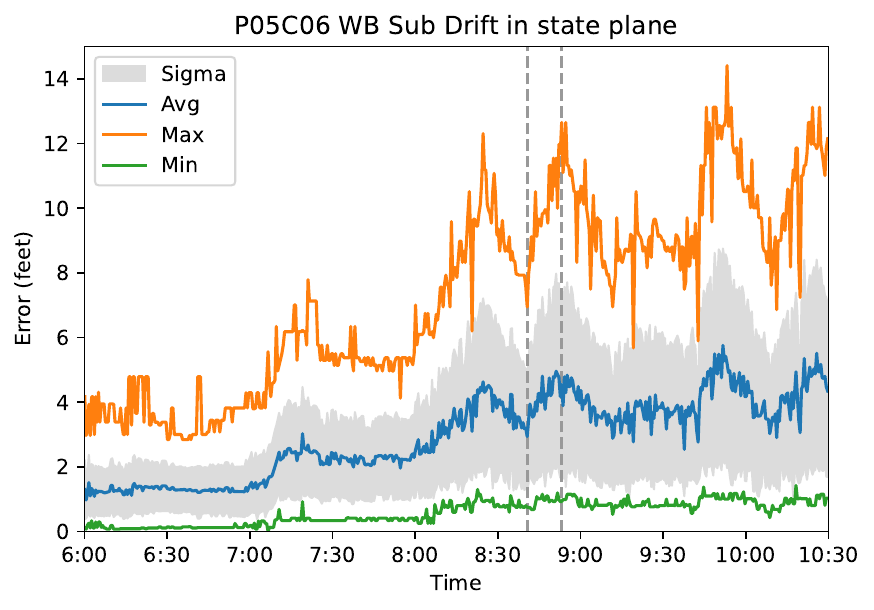}
        \caption{}
        \label{fig:error_subdrift_P05C06}
    \end{subfigure}
    \hfill
    \begin{subfigure}[b]{0.32\textwidth}
        \includegraphics[width=\textwidth]{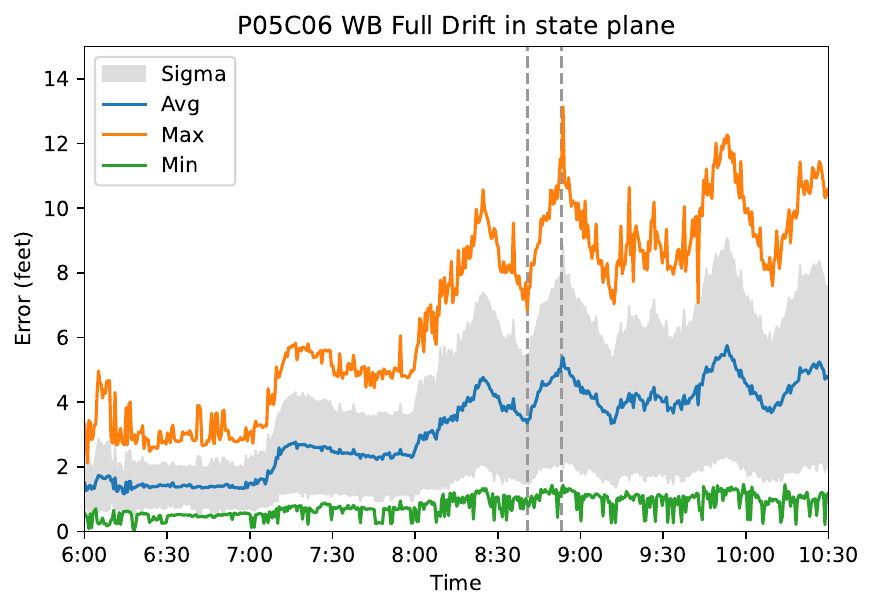}
        \caption{}
        \label{fig:error_fulldrift_P05C06}
    \end{subfigure}
    \caption{(Repeated from Section \ref{sec:re-estimation}.) Goodness of fit and uncorrected drift between reference homography and true scene homography according to two metrics (Sub Drift and Full Drift) for a single typical camera. Horizontal grey dashes show the two time instances for Figure \ref{fig:homo_error_mapf}.}
    \label{fig:error_P05C06}

\end{figure}

Figure \ref{fig:error_P05C06} shows the different state-plane error metrics for a camera with a long field of view. P05C06 is chosen as it clearly illustrates both long and short-term drift in the homography, as well as  the relations between the different metrics: On Figure \ref{fig:error_fitness_P05C06} the homography Fittness is shown, representing the minimum achievable accuracy for the vehicle detection. On Figure \ref{fig:error_subdrift_P05C06} the SubDriftI error is slightly higher than FullDrift shown on Figure \ref{fig:error_fulldrift_P05C06}, because it contains both the fitness and homography error. In this case, the maximum SubDriftI values are higher than the maximum FullDrift values because the rediscovery rate for the feature points is $\sim$80\% for this particular case (relatively high), thus distant points are also likely included. If the drift were more extreme, the most distant correspondence points might not be successfully rediscovered, and the overall reported SubDriftI error may be driven down as a result. This dependence on the number of rediscovered points is one reason why FullDrift is preferred to SubDrift in this work for comparing homography re-estimation performance.

\begin{figure*}[h]
    \centering
    \begin{subfigure}[b]{0.49\textwidth}
        \centering
        \includegraphics[width=\textwidth]{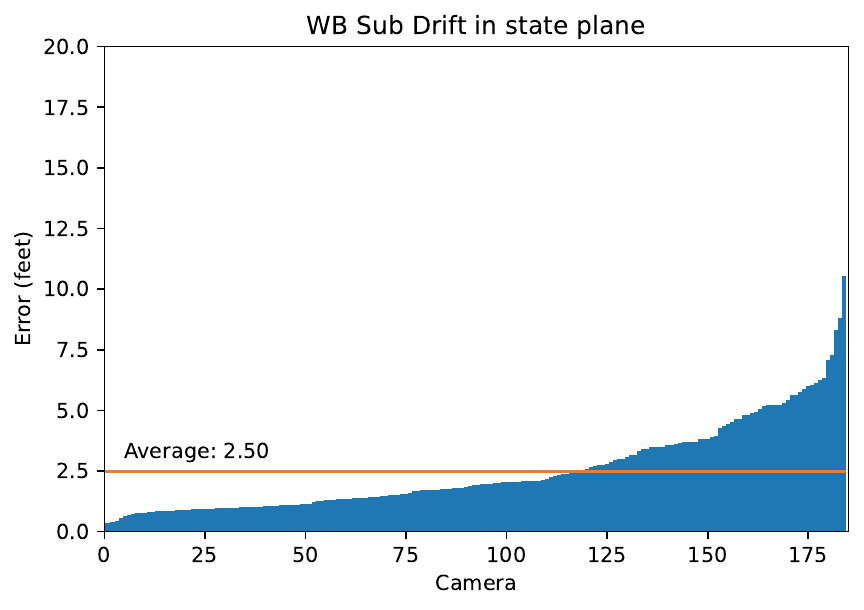}
    \end{subfigure}
    \hfill 
    \begin{subfigure}[b]{0.49\textwidth}
        \centering
        \includegraphics[width=\textwidth]{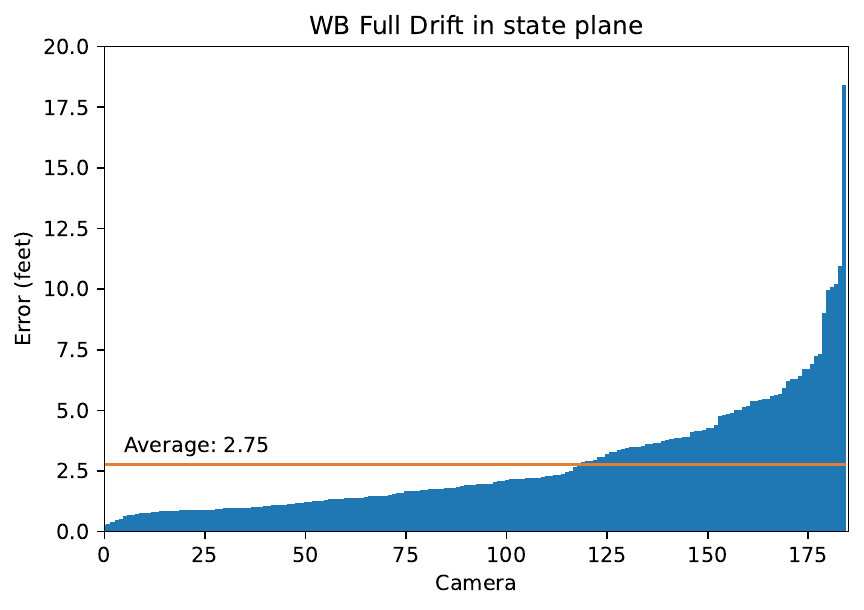}
    \end{subfigure}
    \caption{Mean averaged SubDrift and FullDrift calculated for all westbound cameras in the system, sorted by error magnitude.}
    \label{fig:homo_compare_error_wb}
    %\newline
\end{figure*}

\begin{figure*}[h]
    \centering   
    \begin{subfigure}[b]{0.49\textwidth}
        \centering
        \includegraphics[width=\textwidth]{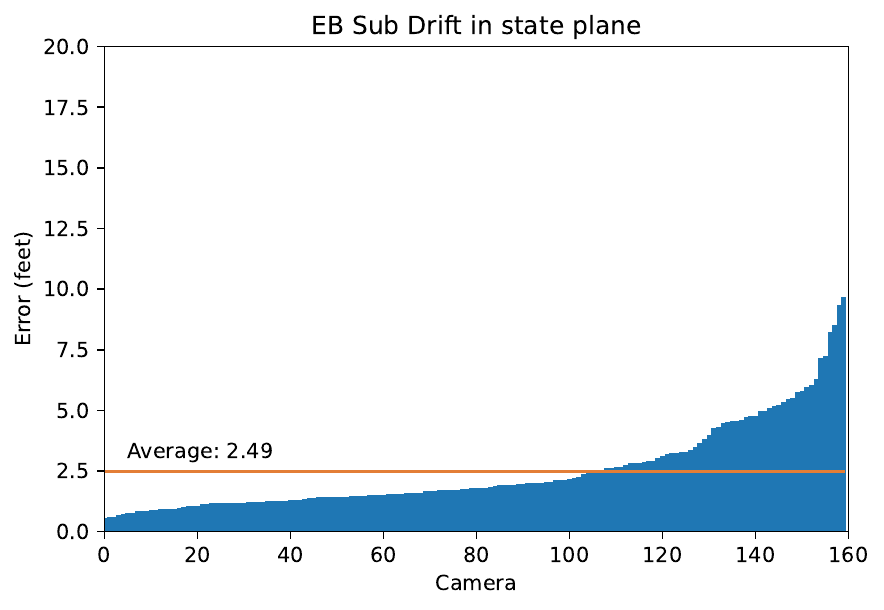}
        \caption{}
    \end{subfigure}
    \hfill 
    \begin{subfigure}[b]{0.49\textwidth}
        \centering
        \includegraphics[width=\textwidth]{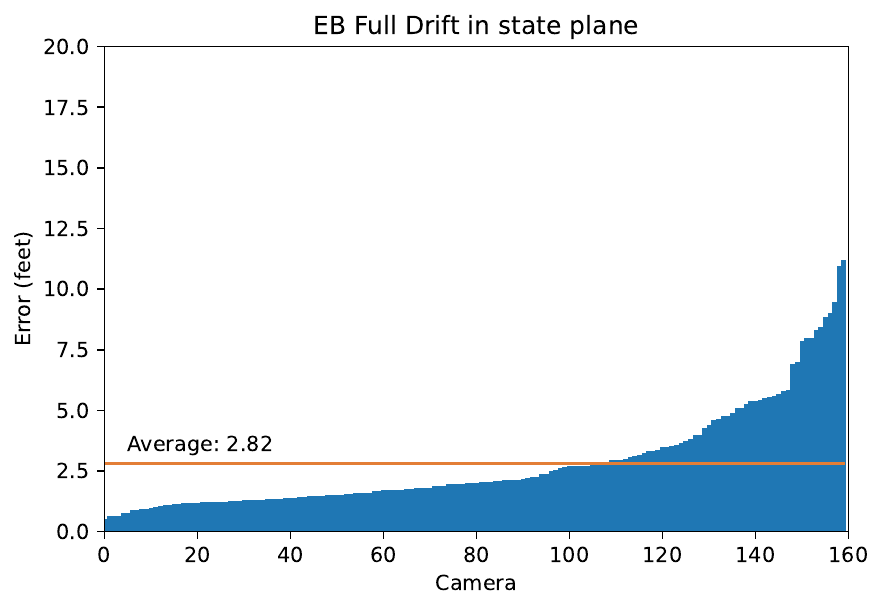}
        \caption{}
    \end{subfigure}
    \caption{Mean averaged SubDrift and FullDrift calculated for all eastbound cameras in the system, sorted by error magnitude.}
    \label{fig:homo_compare_error_eb}
\end{figure*}

Figure \ref{fig:homo_compare_error_wb}, \ref{fig:homo_compare_error_eb} shows the mean FullDrift and SubDrift calculated and averaged over the time of the recording, for all cameras in the system, sorted by magnitude, for each direction of roadway travel. Note that not all cameras have fields of view defined for both sides of the roadway (hence there are more total westbound fields of view than eastbound fields of view). The mean SubDrift is Lower than the mean FullDrift over all cameras, and the maximum mean SubDrift (over all correspondence points and times) is lower than for FullDrift. However as seen in Figure \ref{fig:error_fitness_P05C06}, for some time instances the maximum SubDrift can be higher than the maximum FullDrift.

\begin{figure}[h]
    \centering
    \begin{subfigure}[b]{0.49\textwidth}
        \includegraphics[width=\textwidth]{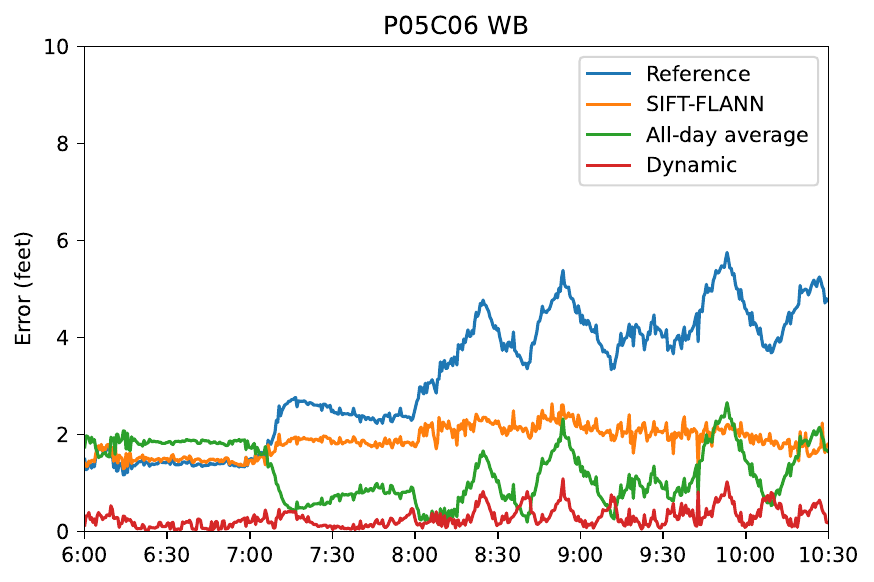}
        \caption{}
        \label{fig:homo_compare_cam_P05C06}
    \end{subfigure}
    \hfill
    \begin{subfigure}[b]{0.49\textwidth}
        \includegraphics[width=\textwidth]{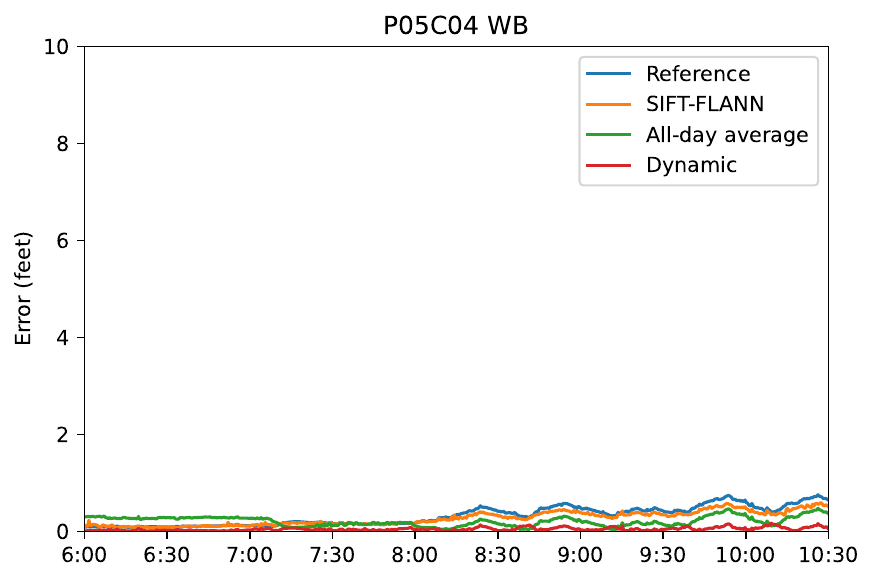}
        \caption{}
        \label{fig:homo_compare_cam_P05C04}
    \end{subfigure}

    \caption{Error dynamics for two cameras over time with each homography re-estimation methods. Long field of view cameras e.g. \ref{fig:homo_compare_cam_P05C06}} have more pronounced errors over down-looking cameras, e.g. \ref{fig:homo_compare_cam_P05C04}.
    \label{fig:homo_compare_cam}

\end{figure}

The effectiveness of the homography re-estimation methods is illustrated on Figure \ref{fig:homo_compare_cam}, which shows the additional error above the error floor, utilizing the FullDrift metric, for two cameras. The reference (blue) indicates the resulting error without any mitigation, showing both long term (high mean) and short term (high variance) error. The SIFT-FLANN method (orange) illustrates the performance of an optical flow based "camera stabilization" solution. The static, all-day average method (green) removes the long term error, although it is mostly unable to remove the error caused by the sunflower effect. Finally since the dynamic homography (red) utilizes nearby (temporal) homography estimations for a given time instance, it can cope with short-term fluctuations. Note that in most cases the all-day average is superior to the SIFT-FLANN method, although there can be time instances where the former can produce better results. Figure \ref{fig:homo_compare_cam} also illustrates, that long field of view cameras (e.g. P05C06) are more sensitive to camera movement, compared to mainly downwards-facing cameras (e.g. P05C04), thus the error caused by drift is more pronounced for them. 

\begin{figure*}[h]
    \centering
    \begin{subfigure}[b]{0.45\textwidth}
        \centering
        \includegraphics[width=\textwidth]{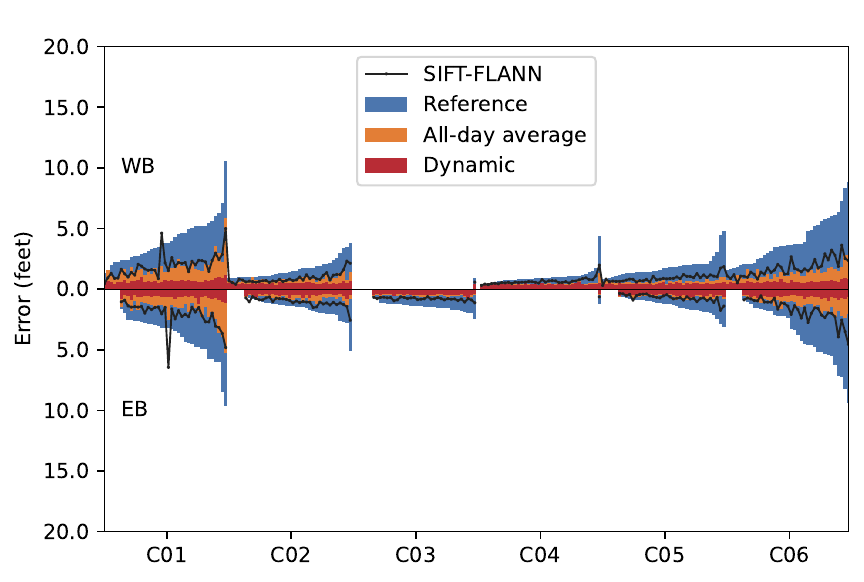}
        \caption{SubDrift}
        \label{fig:homo_compare_fittness}
    \end{subfigure}
    \hfill 
    \begin{subfigure}[b]{0.45\textwidth}
        \centering
        \includegraphics[width=\textwidth]{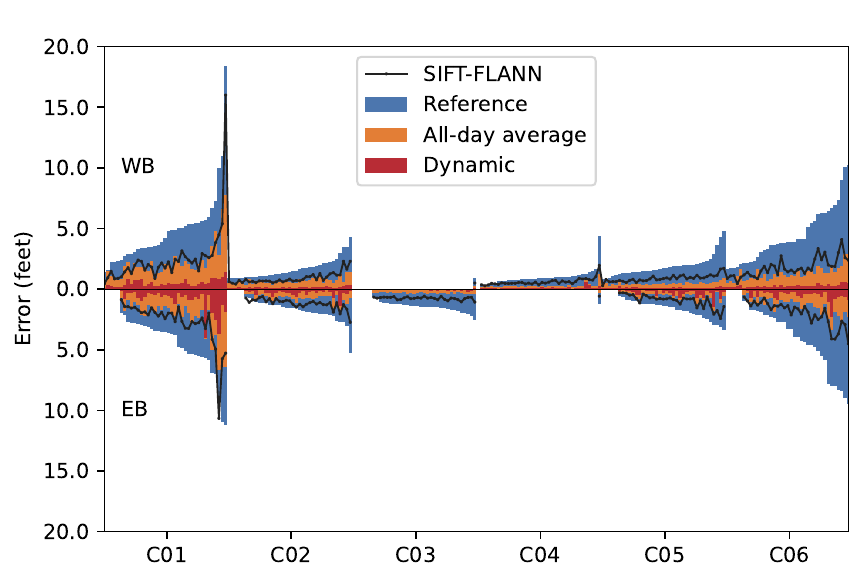}
        \caption{FullDrift}
        \label{fig:homo_compare_all}
    \end{subfigure}
    \caption{Remaining SubDrift and FullDrift errors for each camera after (black) SIFT-FLANN feature-matching, (orange) one-day best fit homography re-estimation, and (red) dynamic homography re-estimation methods relative to orignal reference homography baseline (blue). Cameras are grouped by position on pole (see Figure 2 in main text.) and by side of roadway (westbound homographies on top, eastbound on bottom).)}
    \label{fig:homo_compare}
\end{figure*}

Lastly, figure \ref{fig:homo_compare} compares the remaining mean average SubDrift and FullDrift errors for each camera after (black) SIFT-FLANN feature-matching, (orange) one-day best fit homography re-estimation, and (red) dynamic homography re-estimation methods relative to original reference homography baseline (blue). The overall trends and relative performance amongst methods is unchanged; in most cases the SIFT-FLANN baseline outperforms the uncorrected reference homography, the all-day average outperforms SIFT-FLANN, and the dynamic method outperforms the all-day average.

\FloatBarrier
\pagebreak

\section{Additional GPS Trajectory Plots}
\begin{figure*}[hb]
    \centering
    \begin{subfigure}[b]{\textwidth}
        \centering
        \includegraphics[width=\textwidth]{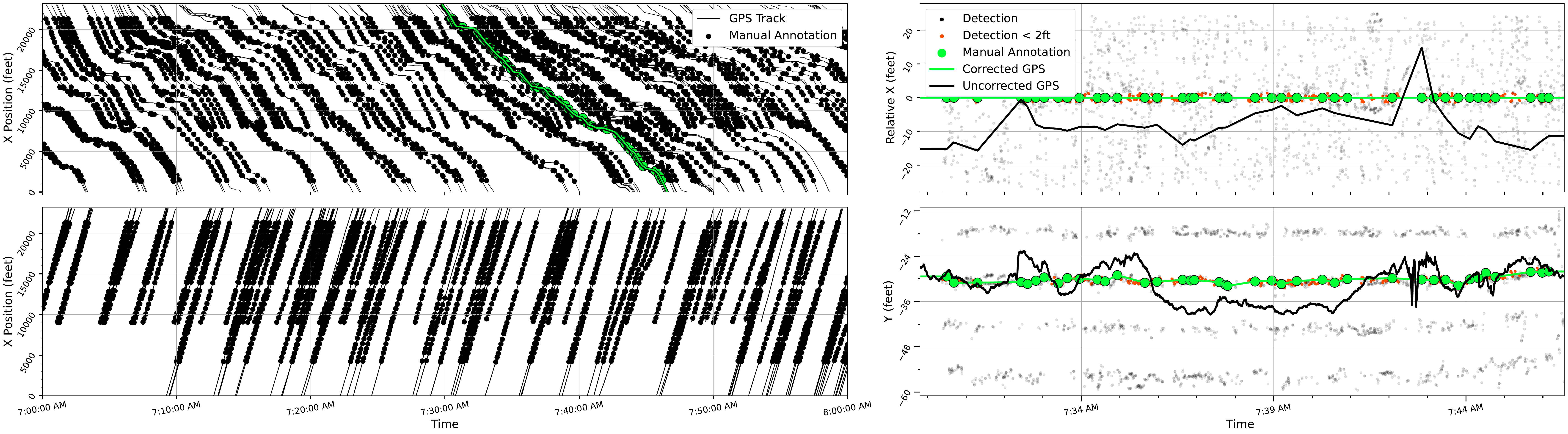}
    \end{subfigure} 
    \hrule 
        \centering
    \begin{subfigure}[b]{\textwidth}
        \centering
        \includegraphics[width=\textwidth]{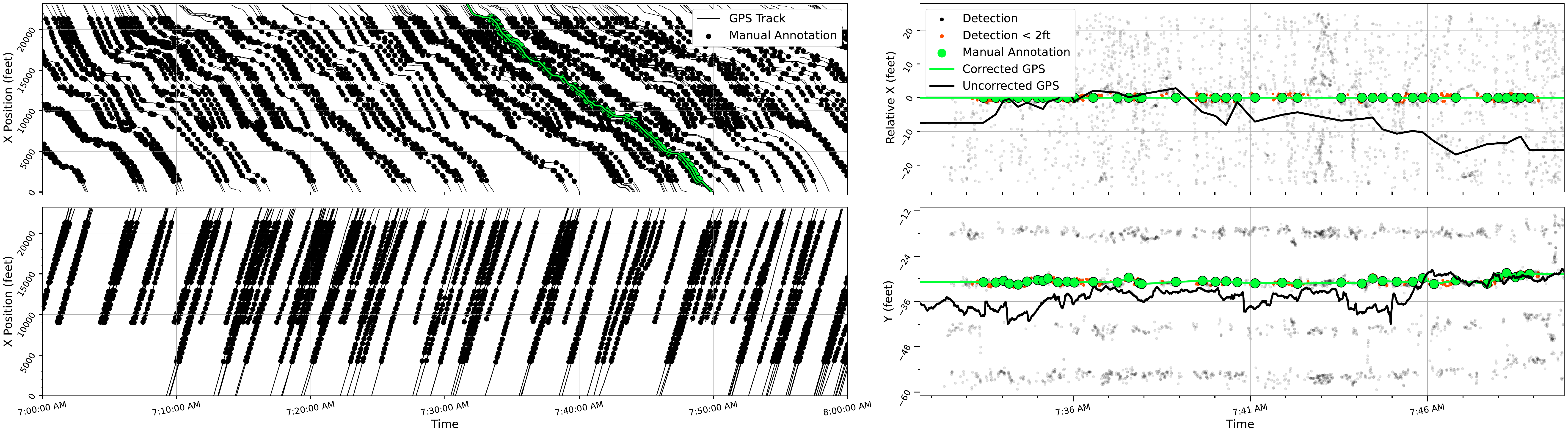}
    \end{subfigure}
    \hrule
        \centering
    \begin{subfigure}[b]{\textwidth}
        \centering
        \includegraphics[width=\textwidth]{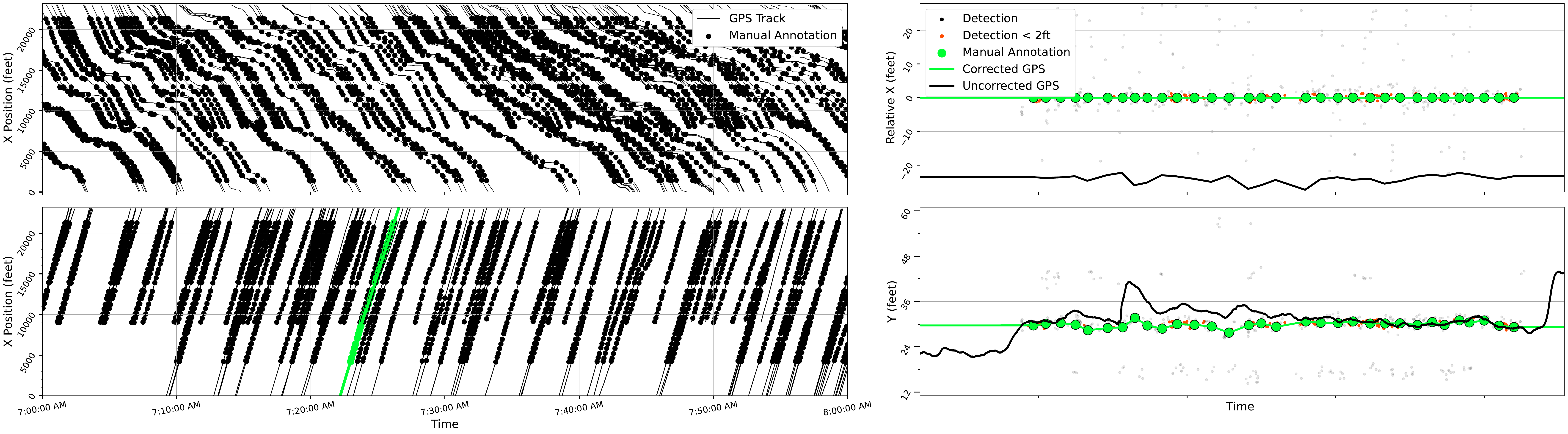}
    \end{subfigure}
    \hrule 
        \centering
    \begin{subfigure}[b]{\textwidth}
        \centering
        \includegraphics[width=\textwidth]{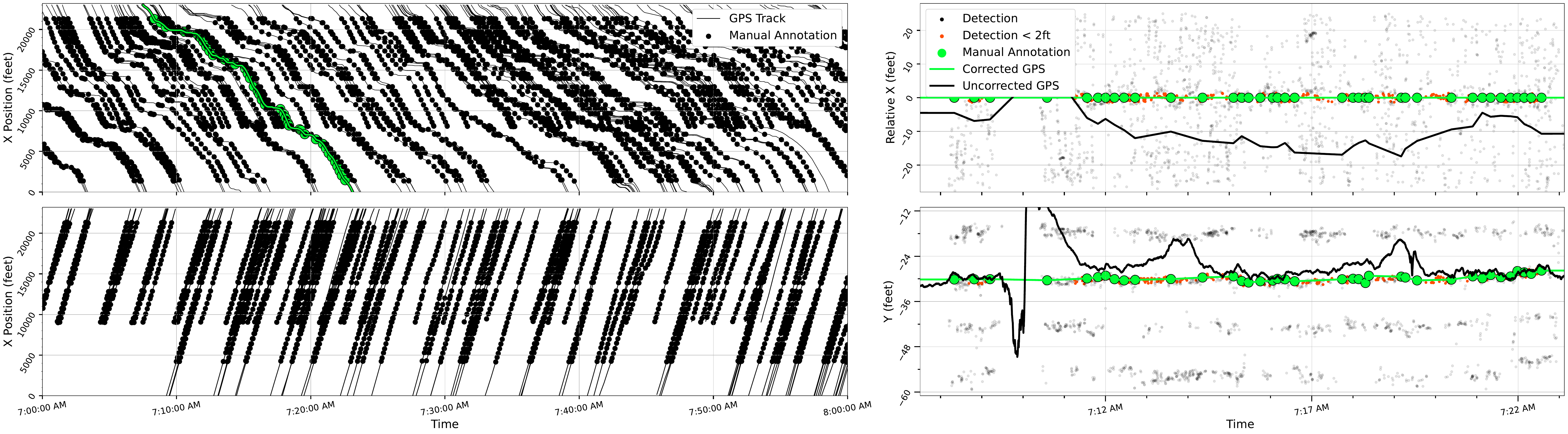}
    \end{subfigure}
    
    \caption{Plots for 4 individual GPS trajectories. In each sub-figure, \textbf{(top left)} shows westbound X-position and \textbf{(bottom left)} shows eastbound X-position for the whole scene duration. \textbf{(top right)} shows nearby (in relative X-coordinates) and \textbf{(bottom right)} shows nearby (in Y-coordinates) detections, manual annotations, and uncorrected GPS trajectory, relative to the corrected GPS trajectory. }
    \label{fig:more-gps}
\end{figure*}

Figure \ref{fig:more-gps} shows plots for additional GPS tracks through the video scene. Manually annotated object positions (circles) and GPS positions (lines) for are shown for the overall scene, divided by westbound (top left) and eastbound (bottom left) direction of travel. (right) x-position relative to the corrected GPS track (top) and absolute y-position (bottom) are plotted for the highlighted GPS track. It can be seen that a.) the corrected GPS trajectories align more closely with object detections (black dots) and b.) there is significant deviation between the corrected and uncorrected GPS trajectories. Note that especially the Y-coordinate error in the uncorrected trajectories varies in character across different GPS trajectories.

\FloatBarrier
\pagebreak

\section{Object Detector}
\label{sec:detection}

\subsection{3D Bounding Box Parameterization}
Object detections are produced by a Retinanet object detector with Resnet-50 FPN backbone \cite{lin2017focal}. We extend the first convolutional layer of this network to allow two input frames (the current and a rolling average frame). This network has two output heads: a classification head and a regression head, each of which outputs a set of outputs per anchor box. The shape of the classification head output $C$ is $[n,c]$ where $n$ is the number of anchor boxes in the network and $c$ is the number of classes. For an anchor box $i$, the final output object class and object confidence are taken as:

\begin{equation}
    \text{class}_i = \argmax_c (C[i,:])
\end{equation}
\begin{equation}
    \text{confidence}_i = \max (C[i,:])
\end{equation}

The shape of the regression head output $R$ is $[n,8]$. Let $A_i$ parameterize a single anchor box $i$ according to:
\begin{equation}
    A[i,:] = [x_a,y_a,w_a,h_a]
\end{equation}
denoting the x-coordinate,y-coordinate,width and height of the anchor box, respectively. The regression output corresponding to that anchor box $R[i,:]$ parameterizes a rectangular prism (ignoring the effects of perspective foreshortening) according to:
\begin{equation}
    R[i,:] = [x_c,y_c,x_l,y_l,x_w,y_w,x_h,y_h]
\end{equation}

\begin{figure}
    \centering
    \includegraphics[width = \textwidth]{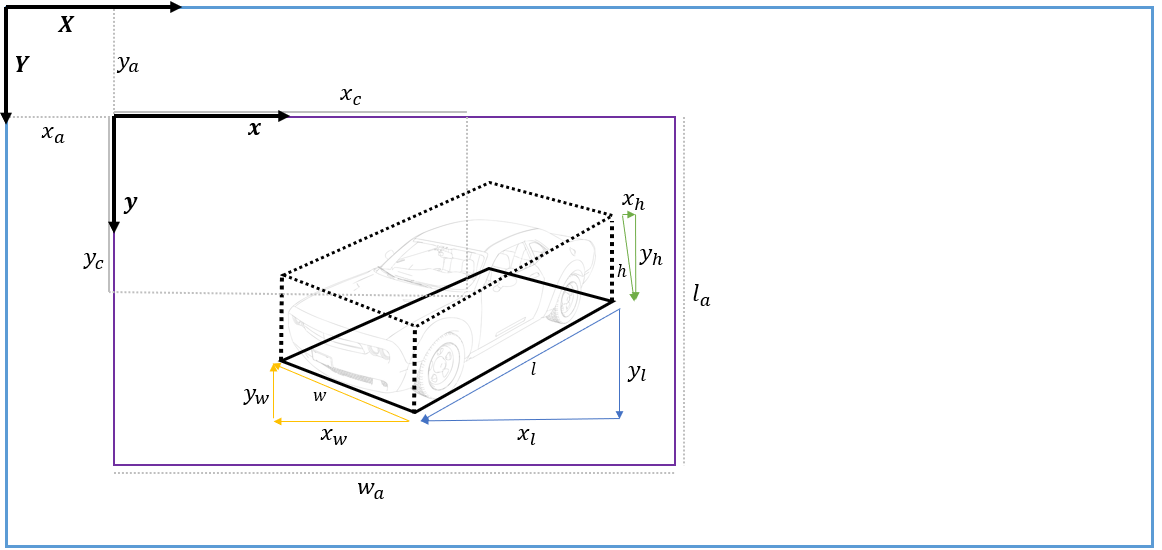}
    \caption{Bounding box \textit{rectangular prism} parameterization. Anchor box (purple) reference coordinate system denoted as (\textbf{x},\textbf{y}) and global frame (blue) reference coordinate system denoted as (\textbf{X},\textbf{Y}). Angle of $h$ is exaggerated to show sub-components $x_h$ and $y_h$.}
    \label{fig:bbox-param}
\end{figure}

where ($x_c$,$y_c$) indicates the center of the rectangular prism relative to the anchor box top left corner, $x_l$ and $y_l$ indicate the x-pixel and y-pixel components of the rectangular prism's length, $x_w$ and $y_w$ indicate the x-pixel and y-pixel components of the rectangular prism's width, and $x_h$ and $y_h$ indicate the x-pixel and y-pixel components of the rectangular prism's height. All components (e.g. $x_l$,$y_l$) are relative to the anchor box dimensions (in this case $w_a$ and $h_a$). By convention, each dimension is directional and is measured relative to the rear bottom left corner. Figure \ref{fig:bbox-param} provides a visual overview of each component.

The (\textbf{X},\textbf{Y}) pixel coordinates for back bottom left coordinate ($x_{rbl},y_{rbl}$ within the overall image can then be written as:
\begin{align}
    & x_{bbl} = x_a + (x_c - x_l/2 - x_w/2 + x_h/2) * w_a \\ \nonumber 
    & y_{bbl} = y_a + (y_c - y_l/2 - y_w/2 + y_h/2) * h_a 
\end{align}

The 7 other corner coordinates can be similarly written. The resulting rectangular prism can then be converted into state plane and roadway coordinates via the methods discussed in Appendix \ref{sec:coordinates}. Readers familiar with 3D detection in the autonomous vehicle context will likely question the use of 3D bounding boxes produced explicitly in state space, as in e.g. \cite{hu2019joint,roddick2018orthographic}. In these contexts, a single camera is statically mapped to the world coordinate plane (i.e. the same pixel within an image always corresponds to the same point on the ground plane in world space). This is not the case in our application, in which a single CNN is desired to produce object detections for frames from all 234 cameras. Thus, a \textit{viewpoint agnostic} anchor box / bounding box formulation such as the one used here is required. We also consider the direct regression of 3D bounding box corner coordinates as in \cite{li2020rtm3d} but empirically this results in poor performance.

\subsection{Training}

We train the above object detector on the I24-3D dataset \cite{gloudemans2023dataset}. We use the following loss formulation:

\begin{equation}
    Loss = Loss_{2D} + \beta \times Loss_{3D} + Loss_{cls} + \gamma Loss_{vp}
\end{equation}

where: 
\begin{itemize}
    \item  $Loss_{2D}$ is intersection-over-union loss \cite{rezatofighi2019generalized} for the minimum enclosing 2D bounding boxes for each of the predicted and target bounding box
    \item $Loss_{3D}$ is MSE loss computed between the target and predicted bounding box corners
    \item $Loss_{cls}$ is classification focal loss \cite{lin2017focal}\
    \item $\beta$ is a weighting coefficient, here set to 2.
    \item $\gamma$ is a weighting coefficient, here set to 1/3
\end{itemize}

and $Loss_{vp}$ is a term designed to enforce the length, width, and height components of the predicted rectangular prism to align closely to the corresponding vanishing points by penalizing the angle between these vectors and the vanishing point directions (relative to the center of the predicted bounding box). \textbf{Succinctly, $Loss_{vp}$  is 0 when each bounding box is perfectly axis-aligned relative to the vanishing points, and is 1 when the axes of the bounding box are perfectly orthogonal to their respective vanishing points. }In this way, the training enforces the CNN to utilize visual cues from the image and align bounding boxes along these visual axes.

\begin{align}
    Loss_{vp} =  & 1/2 -  \alpha_l((y_l  (y_{vpl} - y_c)) + x_l(x_{vpl} - x_c))/(2(x_{vpl} - x_c)(y_{vpl} - y_c) y_l x_l) \:+ \\ \nonumber 
                 & 1/2 -  \alpha_w((y_w  (y_{vpw} - y_c)) + x_w(x_{vpw} - x_c))/(2(x_{vpw} - x_c)(y_{vpw} - y_c) y_w x_w) \: + \\ \nonumber 
                 & 1/2 -  \alpha_h((y_h  (y_{vph} - y_c)) + x_h(x_{vph} - x_c))/(2(x_{vph} - x_c)(y_{vph} - y_c) y_h x_h)
\end{align}

where ($x_{vpl}$,$y_{vpl}$) is the vanishing point, in pixel coordinates, aligned with the roadway direction of travel, ($x_{vpw}$,$y_{vpw}$) is the vanishing point, in pixel coordinates, perpendicular with the roadway direction of travel, and ($x_{vph}$,$y_{vph}$) is the vanishing point, in pixel coordinates for vertically aligned lines within the frame, $\alpha_l$ is -1 if the rear of the bounding box is closer to ($x_{vpl}$,$y_{vpl}$) than the front of the bounding box and 1 otherwise, $\alpha_w$ is -1 if the left side of bounding box is closer to ($x_{vpw}$,$y_{vpw}$) than the right side of the bounding box and 1 otherwise, and $\alpha_h$ is -1 if the bottom of the bounding box is closer to ($x_{vph}$,$y_{vph}$) than the front of the bounding box and 1 otherwise.

\FloatBarrier
\pagebreak
\section{Experimental Details}
\subsection{Evaluation Protocol}
Each object tracker is run using the detection set from Appendix \ref{sec:detection}. GPS trajectories and detections from each camera are obtained at slightly different times. To account for this, tracking evaluation is performed at fixed 0.1 second intervals, and each GPS trajectory and object tracklet position is linearly interpolated at each evaluation time. GPS trajectories extend somewhat outside of the temporal duration for which detections are available. Evaluation is performed only over the temporal range for which both GPS trajectories and detections exist. Moreover, each GPS trajectory is clipped in the X-range $[0,23000]$ such that the trajectory is always visible within the field of view of the overall camera network. Evaluation is performed as in \cite{bernardin2008evaluating}. For all metrics other than HOTA metrics, a lax IOU of 0.1 is required for an object tracklet and GPS trajectory to be matched, and matching is performed with the Hungarian Algorithm for bipartite matching \cite{kuhn1955hungarian}. For each object tracklet, we use the median reported dimension ($l$,$w$, and $h$, in feet) over all reported tracklet dimension measurements.

\subsection{Parameter Settings}
Table \ref{tab:parameters} lists relevant parameter settings for each implemented algorithm. Kalman filter parameters were empirically fine-tuned using the I24-3D dataset \cite{gloudemans2023dataset}.

\begin{table}[h]
    \centering
    \begin{tabular}{l|l|l|l} 
    \toprule
         \textbf{Algorithm} & \textbf{Parameter} & \textbf{Value} & \textbf{Description} \\ \midrule \midrule

        \textbf{ALL}                                       & $t_{max}$       & 2     & Maximum time (sec) between detections before track is terminated \\
                                                           & $t_{min}$       & 2     & Minimum track length (sec) \\ 
                                                           & $f_{eval}$      & 10    & Evaluation time step (Hz) \\ \midrule \midrule
                                               
        \textbf{SORT} \cite{bewley2016simple}              & $\sigma_{high}$ & 0.5   & Required object confidence to be included in detection set \\
                                                           & $\phi_{nms}$    & 0.1   & Non-maximal suppression IOU threshold \\ 
                                                           & $d_{max}$       & 10    & Maximum allowable distance for a match (ft) \\ 
                                                           & $f_{track}$     & 10    & Tracking time step (Hz) \\ \midrule
                                                           
        \textbf{IOUT} \cite{bochinski2017high}             & $\sigma_{high}$ & 0.5   & Required object confidence to be included in detection set \\
                                                           & $\phi_{nms}$    & 0.1   & Non-maximal suppression IOU threshold \\ 
                                                           & $\phi_{min}$    & 0.1   & Minimum IOU for a match \\ 
                                                           & $f_{track}$     & 15    & Tracking time step (Hz) \\ \midrule
        
        \textbf{KIOU} \cite{kiout}                         & $\sigma_{high}$ & 0.5   & Required object confidence to be included in detection set \\
                                                           & $\phi_{nms}$    & 0.1   & Non-maximal suppression IOU threshold \\ 
                                                           & $\phi_{min}$    & 0.1   & Minimum IOU for a match \\ 
                                                           & $f_{track}$     & 10    & Tracking time step (Hz) \\ \midrule
                                                           
        \textbf{ByteTrack} (L2) \cite{zhang2022bytetrack}  & $\sigma_{high}$ & 0.01  & Required object confidence to be included in detection set \\
                                                           & $\phi_{nms}$    & 0.1   & Non-maximal suppression IOU threshold \\ 
                                                           & $d_{max}$       & 10    & Maximum allowable distance for a match (ft) \\ 
                                                           & $\tau_{high}$   & 0.4   & Required confidence to be included in first matching step \\
                                                           & $f_{track}$     & 10    & Tracking time step (Hz) \\ \midrule  
        
        \textbf{ByteTrack} (IOU) \cite{zhang2022bytetrack} & $\sigma_{high}$ & 0.01  & Required object confidence to be included in detection set \\
                                                           & $\phi_{nms}$    & 0.1   & Non-maximal suppression IOU threshold \\ 
                                                           & $\phi_{min}$    & 0.1   & Minimum IOU for a match \\ 
                                                           & $\tau_{high}$   & 0.4   & Required confidence to be included in first matching step \\
                                                           & $f_{track}$     & 10    & Tracking time step (Hz) \\ \midrule 
                                                           
        \textbf{Oracle}                                    & $\phi_{min}$    & 0.1   & Minimum IOU for a match \\
                                                           & $f_{track}$     & 10    & Tracking time step (Hz) \\ \bottomrule
        
    \end{tabular}
    \caption{Caption}
    \label{tab:parameters}
\end{table}

\FloatBarrier
\pagebreak
\section{Treatment of Personally Identifiable Information}

As stated in \cite{gloudemans202324}, the primary purpose of the I-24 system is not to produce video data, but rather is to produce vehicle trajectory data. The data management plan for the system states that in general, video data is not released to the public or stored for long periods of time, but occasionally video data may be released to enable work on training, testing, and validation of trajectory generation algorithms. Thus, the video released in this work represents an edge case for the system,

Nevertheless, the release of video data (or any data with individuals represented) carries with it the risk of releasing \textit{personally identifiable information} (PII) on the included individuals. In this work we make every effort to prevent the leakage of PII to the public, and furthermore make it infeasible to automatically extract PII such that doing so becomes extremely onerous to potential bad actors. A three-tiered approach is used: 1.) We release, more or less, a random hour of data, such that the potential for capturing a discrete event of interest to a third party is near zero. 2.) We automatically redact all license plate information from all visible vehicles. 3.) We manually redact any regions containing private property or visible people.

\begin{figure}[hb]
    \centering
    \includegraphics[width = \textwidth]{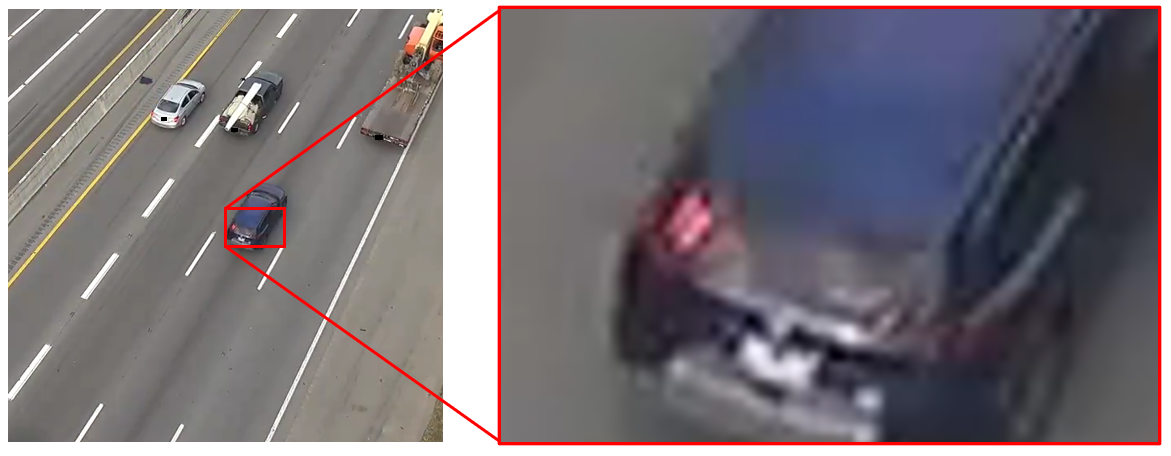}
    \caption{Example license plate from this dataset before redaction. License plate information is unrecoverable; for reference, the orange numbers on the rear window of the vehicle are about 5 times as large as license plate text and are barely discernible. (The pictured vehicle is part of the GPS instrumented vehicle fleet and does not belong to an individual).}
    \label{fig:license}
\end{figure}

Even in un-redacted video data, license plates are in almost all cases impossible to read. Figure \ref{fig:license} provides a typical camera field of view, with the license plate only about 15 pixels wide and subject to significant blurring from vehicle motion. (The vehicle used in this image is part of the GPS instrumented vehicle fleet and does not belong to an individual). Nevertheless, we run an off-the-shelf license plate redaction algorithm \cite{silva2021flexible} on all frames and cover all detected license plates with a black rectangle.

\begin{figure}
    \centering
    \includegraphics[width = \textwidth]{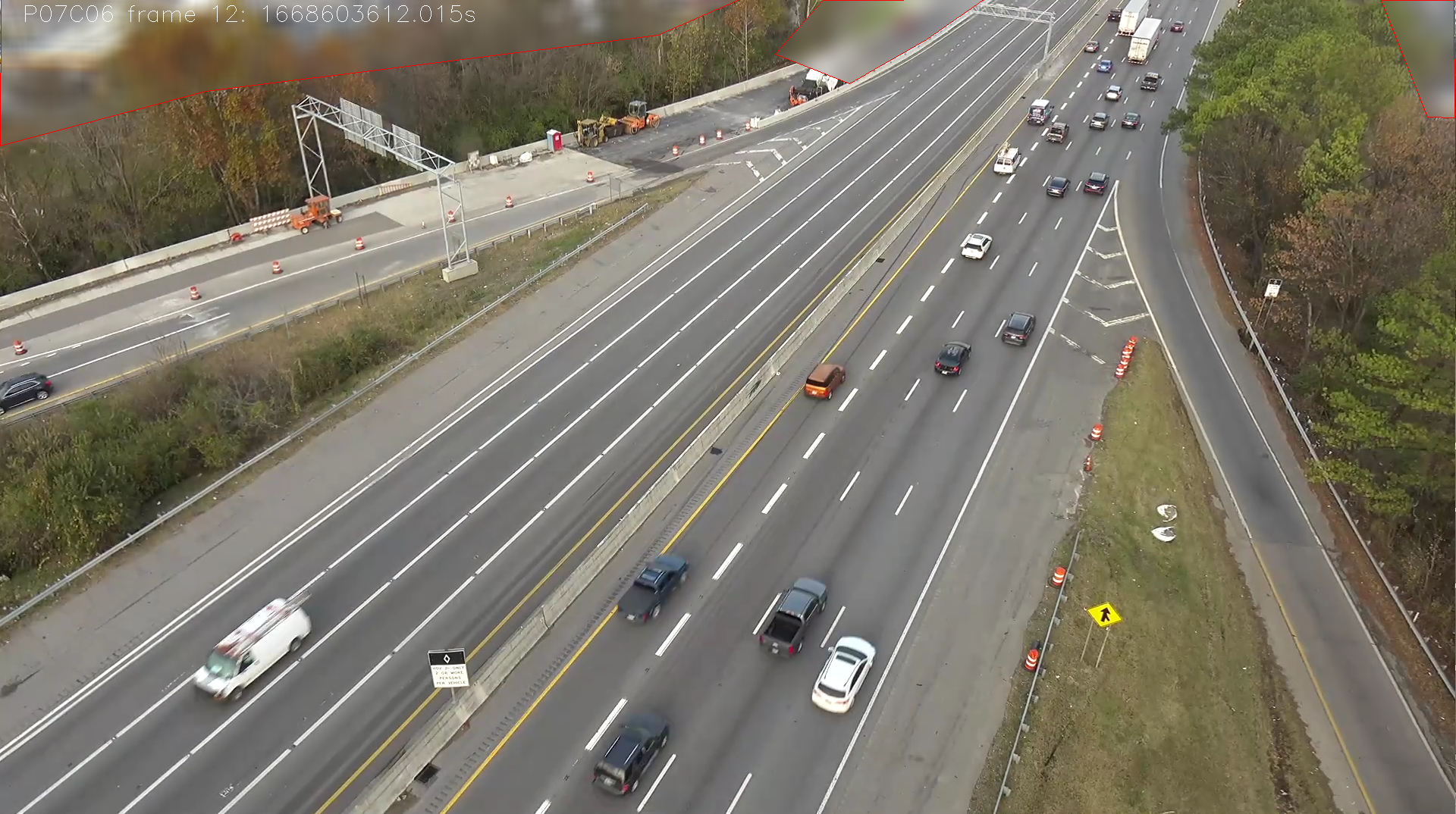}
    \caption{Example redacted regions (red outline, blurred) for one camera field of view .}
    \label{fig:redact}
\end{figure}

We then manually inspect each video and redact the following sets of information. These areas are blurred in the released videos. Polygons defining the redacted regions are also released such that data users can replace these pixel values as desired for computer vision applications. Figure \ref{fig:redact} shows an example of redacted regions.
\begin{itemize}
    \item All visible people
    \item All private property
    \item Any stopped law enforcement vehicle (we leave $\sim1$ sec of the stopped vehicle visible to allow for graceful handling of these vehicles by tracking algorithms.
\end{itemize}

\FloatBarrier
\pagebreak 
\section{Known Data Artifacts and Anomalies}
\label{sec:errata}

\begin{figure}
    \centering
    \includegraphics[width=0.5\textwidth]{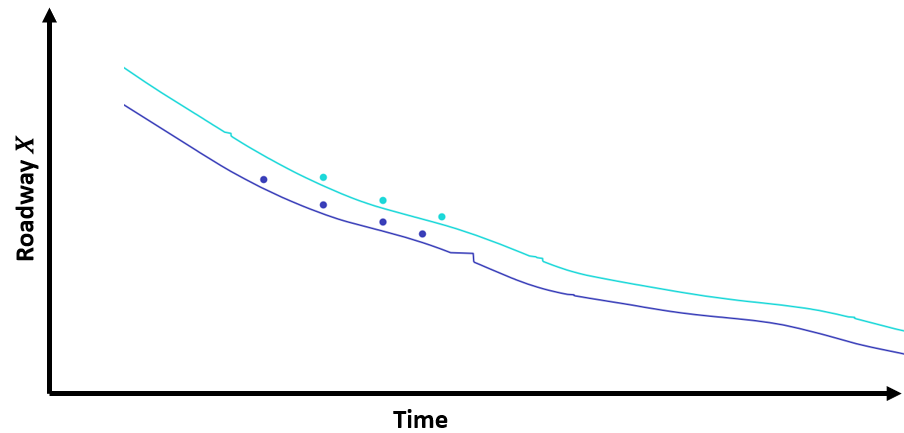}
    \caption{``Sawtooth'' artifact in uncorrected GPS trajectories (lines). Circles depict corresponding manually annotated vehicle positions.}
    \label{fig:sawtooth}
\end{figure}

Here, we report a list of known anomalies and artifacts in the data as of submission time: 
\begin{itemize}
    \item Visible in some cameras on poles 3 and 4, there is a region that is blacked out in all frames and camera fields of view. This region corresponds to a private residence. Different from other redaction areas in this dataset, a virtual mask is applied to this region in all cameras in which it is visible, such that no visual informatiom from this region is ever recorded. Other regions are redacted after recording.
    \item All cameras from Pole 25 are missing; this pole was struck by a vehicle a few days prior to the recording day and could not be restored in time.
    \item Homographies on the eastbound roadway side are not defined for cameras on poles 1-7. Construction work required that temporary solid lane markings be painted, for which i.) corner points were not uniquely distinguishable and ii.) no matching aerial imagery exists due to the short term nature of the construction work.
    \item Reference homographies for cameras on poles 1 and 2 are defined, but homographies are not re-estimated for these cameras. The lane markings were altered between the reference homography day and the recording day meaning that SIFT-FLANN matching and lane marking re-detection fail.
    \item Camera P22C04 has a black ring visible on the left portion of the frame due to a mechanical misalignment of the camera lens and body.
    \item Occasionally, GPS trajectories have missing recordings for time periods on the order of $\sim1$ sec. The onboard sensor filtering attempts to compensate for this missing data, producing ``sawtooth'' artifacts in the trajectory. Figure \ref{fig:sawtooth} shows an example. The majority of these artifacts were removed during data refinement, but due some artifacts may still remain.
\end{itemize}

\FloatBarrier
\pagebreak

%%%%%%%%% REFERENCES
{\small
\bibliographystyle{ieee_fullname}
\bibliography{sources}
}